\documentclass[journal]{IEEEtran}
\usepackage{amssymb}
\usepackage{amsmath}
\usepackage{xcolor}
\usepackage{mathrsfs}
\usepackage{booktabs}
\usepackage{multirow}
\usepackage[flushleft]{threeparttable}
\usepackage{graphicx}
\usepackage{caption}
\usepackage{subfig}
\usepackage{tikz,pgfplots}
\usepackage{tikz}
\usepackage{cleveref}
\usepackage{balance}
\usepackage{fancyhdr}
\usepackage{hyperref}

\usepackage[square,sort,comma,numbers]{natbib}

\begin{document}

\title{Weakly Supervised Estimation of Shadow Confidence Maps in Fetal Ultrasound Imaging}

% \author{Michael~Shell,~\IEEEmembership{Member,~IEEE,}
%         John~Doe,~\IEEEmembership{Fellow,~OSA,}
%         and~Jane~Doe,~\IEEEmembership{Life~Fellow,~IEEE}% <-this % stops a space
\author{Qingjie Meng,
        Matthew Sinclair,
        Veronika Zimmer,
        Benjamin Hou,
        Martin Rajchl,
        Nicolas Toussaint,
        Ozan Oktay,\\
        Jo Schlemper,
        Alberto Gomez,
        James Housden, 
        Jacqueline Matthew,
        Daniel Rueckert,~\IEEEmembership{Fellow, ~IEEE},\\
        Julia A. Schnabel,~\IEEEmembership{Senior member, ~IEEE},
        and Bernhard Kainz,~\IEEEmembership{Senior member, ~IEEE}% <-this % stops a space
% Qingjie, Matt, Veronika, Martin, Nico, ALberto, James, Daniel, Julia, Bernhard
% \thanks{M. Shell was with the Department
% of Electrical and Computer Engineering, Georgia Institute of Technology, Atlanta,
% GA, 30332 USA e-mail: (see http://www.michaelshell.org/contact.html).}% <-this % stops a space
% \thanks{J. Doe and J. Doe are with Anonymous University.}% <-this % stops a space
% \thanks{Manuscript received April 19, 2005; revised August 26, 2015.}}
\thanks{Q. Meng, M. Sinclair, B. Hou, M. Rajchl, O. Otkay, J. Schlemper, D. Rueckert and B. Kainz are with the Biomedical Image Analysis Group, Department of Computing, Imperial College London, London SW7 2AZ, UK, (e-mail:
q.meng16@imperial.ac.uk).}% <-this % stops a space
\thanks{V. Zimmer, N. Toussaint, A. Gomez, J. Housden, J. Matthew and J. A. Schnabel are with School of Biomedical Engineering and Imaging Sciences, King's College London, London WC2R 2LS, UK.}% <-this % stops a space
\thanks{To appear in IEEE TRANSACTIONS ON MEDICAL IMAGING \url{https://ieeexplore.ieee.org/document/8698843} DOI: 10.1109/TMI.2019.2913311. \textcopyright \textcopyright 2019 IEEE. Personal use of this material is permitted. Permission from IEEE must be obtained for all other uses, in any current or future media, including reprinting/republishing this material for advertising or promotional purposes, creating new collective works, for resale or redistribution to servers or lists, or reuse of any copyrighted component of this work in other works.}}%

% The paper headers
%\markboth{Journal of \LaTeX\ Class Files,~Vol.~14, No.~8, August~2015}%
\markboth{TO APPEAR in IEEE TRANSACTIONS ON MEDICAL IMAGING DOI: 10.1109/TMI.2019.2913311}%
{Shell \MakeLowercase{\textit{et al.}}: Bare Demo of IEEEtran.cls for IEEE Journals}
% {Shell \MakeLowercase{\textit{et al.}}: Bare Demo of IEEEtran.cls for IEEE Journals}

\maketitle

\thispagestyle{fancy}
\fancyhead{} 
\lhead{} 
\lfoot{} 
\cfoot{\small{Copyright (c) 2019 IEEE. Personal use of this material is permitted. However, permission to use this material for any other purposes must be obtained from the IEEE by sending a request to pubs-permissions@ieee.org.}} 
\rfoot{}

% As a general rule, do not put math, special symbols or citations
% in the abstract or keywords.
\begin{abstract}
%Detecting acoustic shadows in ultrasound images is challenging in clinical and engineering practice. 
Detecting acoustic shadows in ultrasound images is important in many clinical and engineering applications. 
Real-time feedback of acoustic shadows
% in ultrasound imaging
can guide sonographers to a standardized diagnostic viewing plane with minimal artifacts and can provide additional information for other automatic image analysis algorithms. 
%Deep learning models have been used successfully for a range of cutting-edge applications in clinical ultrasound. 
However, automatically detecting shadow regions using learning-based algorithms is challenging because
% Providing ground truth for automatic learning-based detection algorithms is challenging because 
pixel-wise ground truth annotation of acoustic shadows is subjective and time consuming. In this paper we propose a weakly supervised method for automatic confidence estimation of acoustic shadow regions. Our method is able to generate a dense shadow-focused confidence map. 
% Our architecture consists of a multi-net module, a transfer function and a shadow confidence estimation network. 
In our method, a shadow-seg module is built to learn general shadow features for shadow segmentation, based on global image-level annotations as well as a small number of coarse pixel-wise shadow annotations. 
A transfer function is introduced to extend the obtained binary shadow segmentation to a reference confidence map.
% , which is used by a confidence estimation network that learns the mapping between input images and the reference confidence maps. 
Additionally, a confidence estimation network is proposed to learn the mapping between input images and the reference confidence maps. 
This network is able to predict shadow confidence maps directly from input images during inference. We use evaluation metrics such as DICE, inter-class correlation and etc. to verify the effectiveness of our method. Our method is more consistent than human annotation, and outperforms the state-of-the-art quantitatively in shadow segmentation and qualitatively in confidence estimation of shadow regions. We further demonstrate the applicability of our method by integrating shadow confidence maps into tasks such as ultrasound image classification, multi-view image fusion and automated biometric measurements.
% \footnote{This work has been submitted to the IEEE for possible publication. Copyright may be transferred without notice, after which this version may no longer be accessible.}
\end{abstract}

% Note that keywords are not normally used for peerreview papers.
\begin{IEEEkeywords}
Ultrasound imaging, deep learning, weakly supervised, shadow detection, confidence estimation.
\end{IEEEkeywords}

\IEEEpeerreviewmaketitle

% \vspace{-0.3cm}
\section{Introduction}
\IEEEPARstart{U}{ltrasound} (US) imaging is a medical imaging technique based on reflection and scattering of high-frequency sound in tissues. Compared with other imaging techniques (e.g. Magnetic Resonance Imaging (MRI) and Computed Tomography (CT)), US imaging has various advantages including portability, low cost, high temporal resolution and real-time imaging capability. With these advantages, US is an important medical imaging modality that is utilized to examine a range of anatomical structures in both adults and fetuses. In most countries, US imaging is an essential part of clinical routine for pregnancy health screening between 11 and 22 weeks of gestation~\cite{salomon2011}. 

Although US imaging is capable of providing real-time images of anatomy, diagnostic accuracy is limited by the relatively low image quality. Artifacts such as noise~\cite{abbott1979}, distortions~\cite{steel2005} and acoustic shadows~\cite{feldman2005} make interpretation challenging and highly dependent on experienced operators. These artifacts are unavoidable in clinical practice due to the low energies used and the physical nature of sound wave propagation in human tissues. Better hardware and advanced image reconstruction algorithms have been developed to reduce speckle noise~\cite{Coupe2009, choi2015}. Prior anatomical expertise~\cite{Lange2009} and extensive sonographer training are the only way to handle distortions and shadows to date. %Compared with noise and distortions, acoustic shadows pose a greater challenge.

\begin{figure}[ht]
 \centering
 \setcounter{subfigure}{0}
 \subfloat[Images (image-level labels)]{
 \begin{tabular}{@{\hspace{0.5\tabcolsep}}c@{\hspace{0.3\tabcolsep}}c@{\hspace{0.3\tabcolsep}}}
  \includegraphics[height=1.2cm]{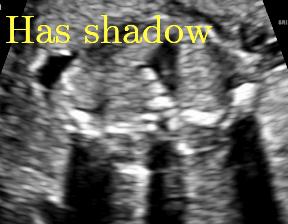} &
  \includegraphics[height=1.2cm]{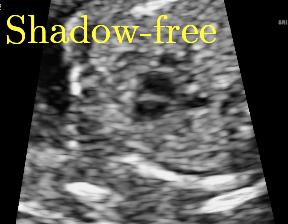} \\
  \includegraphics[height=1.2cm]{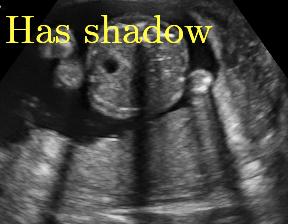} & 
  \includegraphics[height=1.2cm]{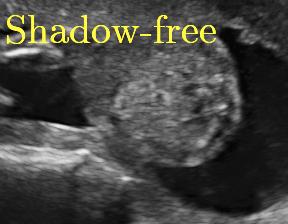}
  \end{tabular}
  }
  \hfill
  \hspace{-5ex}
  \setcounter{subfigure}{1}
  \subfloat[Images with pixel-wise annotations]{
   \begin{tabular}{@{\hspace{0.5\tabcolsep}}c@{\hspace{0.3\tabcolsep}}c@{\hspace{0.3\tabcolsep}}c@{\hspace{0.5\tabcolsep}}}
  \includegraphics[height=1.2cm]{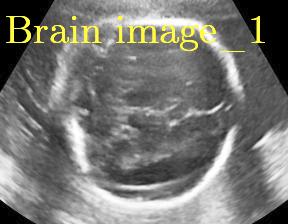} & 
  \includegraphics[height=1.2cm, trim=3cm 1.75cm 2.5cm 1.95cm, clip]{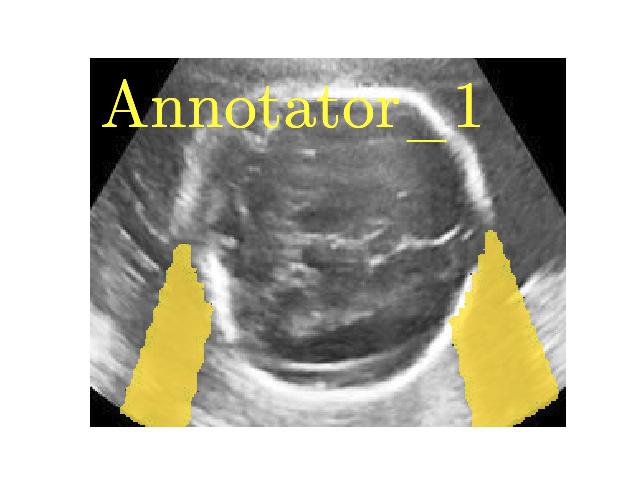} & 
  \includegraphics[height=1.2cm, trim=3cm 1.75cm 2.5cm 1.95cm, clip]{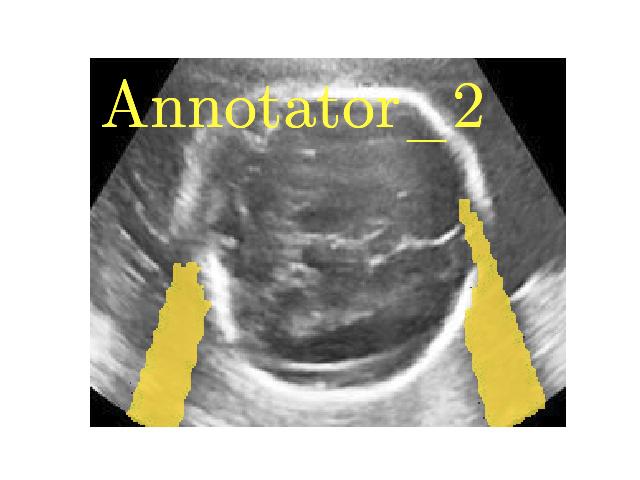} \\
  \includegraphics[height=1.2cm]{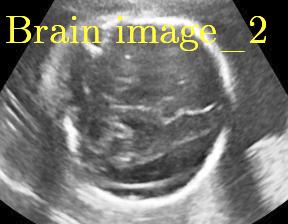} & 
  \includegraphics[height=1.2cm, trim=3cm 1.75cm 2.5cm 1.95cm, clip]{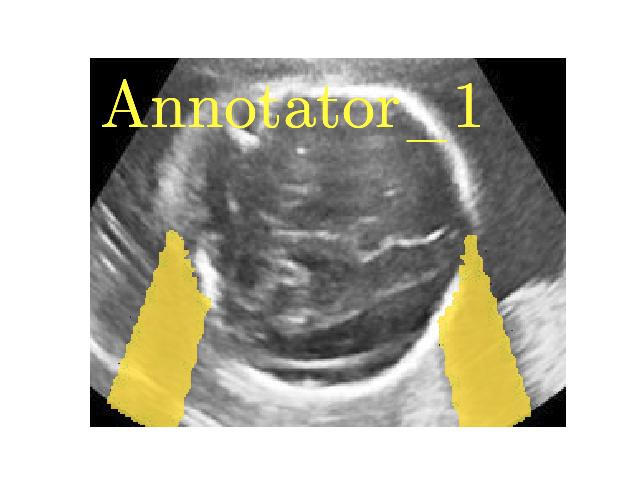} &
  \includegraphics[height=1.2cm, trim=3cm 1.75cm 2.5cm 1.95cm, clip]{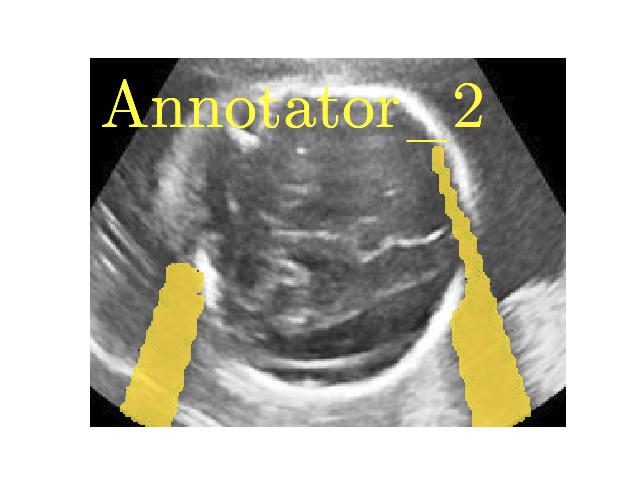}
  \end{tabular}
 }
  \caption{Examples of data sets. (a) Images with global image-level labels (``has shadow" and ``shadow-free"), and (b) Images with coarse pixel-wise annotations from two annotators.}
  \label{DataPresentation}
\end{figure}

Sound-opaque occluders, including bones and calcified tissues, block the propagation of sound waves by strongly absorbing or reflecting sound waves during scanning. The regions behind these sound-opaque occluders return little to no reflections to the US transducer. Thus these areas have low intensity but very high acoustic impedance gradients at their boundaries (e.g. Fig.~\ref{DataPresentation}(a) left column). 
% [CITE artifacts in ultrasound imaging] 
Reducing acoustic shadows and correct interpretation of images containing shadows rely heavily on sonographer experience. Experienced sonographers avoid shadows by moving the probe to a more preferable viewing direction during scanning or, if no shadow-free viewing direction can be found, a mental map is compounded with iterative acquisitions from different orientations.

With less anatomical information in shadow regions, especially when shadows cut through the anatomy of interest, images containing strong shadows can be problematic for automatic real-time image analysis methods such as biometric measurements~\cite{Sinclair2018}, anatomy segmentation~\cite{Berton2016} and US image classification~\cite{christian2017}. Moreover, the shortage of experienced sonographers~\cite{swor2017} exacerbates the challenges of accurate US image-based screening and diagnostics. Therefore, shadow-aware US image analysis is greatly needed and would be beneficial, both for engineers who work on medical image analysis, as well as for sonographers in clinical practice.

%Providing knowledge of shadow regions is useful for automatic medical image analysis algorithms to determine the anatomical properties, while real-time detection of shadow regions is capable of helping training sonographers, as well as guiding non-experienced sonographers to easily find satisfied viewing direction during scanning.

% visual representation of data set

% \subsection{Contribution}
\subsubsection*{\textbf{Contribution}} 
% Motivated by the above engineering and clinical needs, 
We propose a novel method based on convolutional neural networks (CNNs) to automatically estimate pixel-wise confidence maps of acoustic shadows in 2D US images. Our method learns an initial latent space of shadow regions from images consisting of multiple anatomies and with global image-level labels (``has shadow'' and ``shadow-free''), e.g. Fig.~\ref{DataPresentation}(a). The basic latent space is then estimated by learning from fewer images of a single anatomy (fetal brain) with coarse pixel-wise shadow annotations (approximately $10\%$ of the images with global image-level labels), e.g. Fig.~\ref{DataPresentation}(b). The resulting latent space is then refined by learning shadow intensity distributions using fetal brain images so that the latent space is suitable for confidence estimation of shadow regions.
By using shadow intensity information, our method can detect more shadow regions than the coarse manual segmentation, especially relatively weak shadow regions. 

The proposed training process is able to build a direct mapping between input images and the corresponding shadow confidence maps in any given anatomy, which allows real-time application through direct inference. 

%We build a shadow-seg module to learn a generalized shadow representation from different weak annotations, including image-level labels (``has shadow" and ``shadow-free") of multiple classes of anatomy and a small number of coarse pixel-wise shadow annotations from a single class of anatomy. With these generalized shadow representations, our method can predict confidence maps for shadow regions produced by various anatomical structures, instead of being constrained to certain types of shadows when only a limited set of annotations can be used as reference. 

In contrast to our preliminary work~\cite{meng2018}, which uses separate, heuristically linked components, here we establish a pipeline to make full use of existing data sets and annotations. During inference %(shown in Fig.~\ref{inference}), 
our method can predict both a binary shadow segmentation and a dense shadow-focused confidence map. The shadow segmentation is not limited by hyperparameters such as thresholds in \cite{meng2018}, and the segmentation accuracy as well as shadow confidence maps are greatly improved compared to the state-of-the-art.
% (see Fig.~\ref{compareConf}). 

We have demonstrated in~\cite{meng2018} that shadow confidence maps can improve the performance of an automatic biometric measurement task. In this study, we further evaluate the usefulness of the shadow confidence estimation for other automatic image analysis algorithms such as an US image classification task and a multi-view image fusion task.

\subsection*{Related work}
% \subsection{Related work}
\subsubsection*{Automatic US shadow detection} Acoustic shadows have a significant impact on US image quality, and thus a serious effect on robustness and accuracy of image processing methods. In clinical literature, US artifacts including shadows have been well studied and reviewed~\cite{Kremkau1986,Bouhemad2007,Noble2010}. However, the shadow problem is not well covered in automated US image analysis literature. Automatic estimation of acoustic shadows has rarely been the focus within the medical image analysis community.

Identifying shadow regions in US images has been utilized as a preprocessing step for extracting relevant image content and improving image analysis accuracy in some applications. Penney et al.~\cite{penney2004} have identified shadow regions by thresholding the accumulated intensity along each scanning beam line. Afterwards, these shadow regions have been masked out from US images for US to MRI hepatic image registration. Instead of excluding shadow regions, Kim et al.~\cite{Kim2008} focused on accurate attenuation estimation, and aimed to use attenuation properties for determination of the anatomical properties which can help diagnose diseases. They proposed a hybrid attenuation estimation method that combines spectral difference and spectral shift methods to reduce the influence of local spectral noise and backscatter variations in Radio Frequency (RF) US data. To detect shadow regions in B-Mode scans directly and automatically, Hellier et al.~\cite{Hellier2010} used the probe's geometric properties and statistically modelled the US B-Mode cone. Compared with previous statistical shadow detection methods such as~\cite{penney2004}, their method can automatically estimate the probe's geometry as well as other hyperparameters, and has shown improvements in 3D reconstruction, registration and tracking. However, the method can only detect a subset of `deep' acoustic shadows because of the probe geometry-dependent sampling strategy. 

To improve the accuracy of US attenuation estimation and shadow detection, Karamalis et al.~\cite{Karamalis2012} proposed a more general solution using the Random Walks (RW) algorithm to predict a per-pixel confidence of US images. In \cite{Karamalis2012}, confidence maps represent the uncertainty of US images resulting from shadows, and thus, show the acoustic shadow regions. The confidence maps obtained by this work can improve the accuracy of US image processing tasks, such as intensity-based US image reconstruction and multi-modal registration. However, such confidence maps are sensitive to US transducer settings and limited by the US formation process. Klein et al.~\cite{Klein2015} have further extended the RW method to generate distribution-based confidence maps and applied it to RF US data. This method is more robust since the confidence prediction is no longer intensity-based. 

Some studies have utilized acoustic shadow detection as additional information in their pipeline for other US image processing tasks. Broersen et al.~\cite{Broersen2015} combined acoustic shadow detection for the characterization of dense calcium tissue in intravascular US virtual histology, and Berton et al.~\cite{Berton2016} automatically and simultaneously segment vertebrae, spinous process and acoustic shadow in US images for a better assessment of scoliosis progression. In these applications, acoustic shadow detection is task-specific, and is mainly based on heuristic image intensity features as well as special anatomical constraints.

The aforementioned literature relies heavily on manually selected relevant features, intensity information or a probe-specific US formation process. With the advances in deep learning, US image analysis algorithms have gained better semantic image interpretation abilities.
However, current deep learning segmentation methods require a large amount of pixel-wise, manually labelled ground truth images. This is challenging in the US imaging domain because of (a) a lack of experienced annotators and (b) weakly defined structural features that cause a high inter-observer variability. %Unsupervised and weakly supervised deep learning methods are thus a potentially promising avenue for US image interpolation tasks such as shadow detection. 

\subsubsection*{Weakly supervised image segmentation} Weakly supervised automatic detection of class differences has been explored in other imaging domains (e.g. MRI). For example, Baumgartner et al.~\cite{Baumgartner2017} proposed to use a generative adversarial network (GAN) to highlight class differences only from global image-level labels (Alzheimer's disease or healthy). We used a similar idea in~\cite{meng2018} and initialized potential shadow areas based on saliency maps~\cite{Springeberg2014} from a classification task between images containing shadows and those without. %\cite{meng2018} is able to predict confidence maps for shadow regions but this work requires several heuristic pre- and post-processing steps and consists of four separate components in the framework. 
Inspired by recent weakly supervised deep learning methods that have drastically improved semantic image analysis~\cite{Krizhevsky2012,Zhou2016,Rajchl2017} and to overcome the limitations of~\cite{meng2018}, we develop a confidence estimation algorithm that takes advantages of both types of weak labels, including global image-level labels and a sparse set of coarse pixel-wise labels. 
% The final prediction component  
Our method is able to predict dense, shadow-focused confidence maps directly from input US images in effectively real-time.

\iffalse
\begin{figure}[t]
 \centering
 \includegraphics[width=0.5\textwidth, trim=5.5cm 10cm 8.5cm 7.5cm, clip]{Inference.pdf}
 \caption{Inference framework of the proposed weakly supervised shadow confidence estimation.}
 \label{inference}
\end{figure}
\fi

\section{Method}
\begin{figure*}[htb]
 \centering
 \includegraphics[width=\textwidth, trim=5cm 6cm 4cm 4cm, clip]{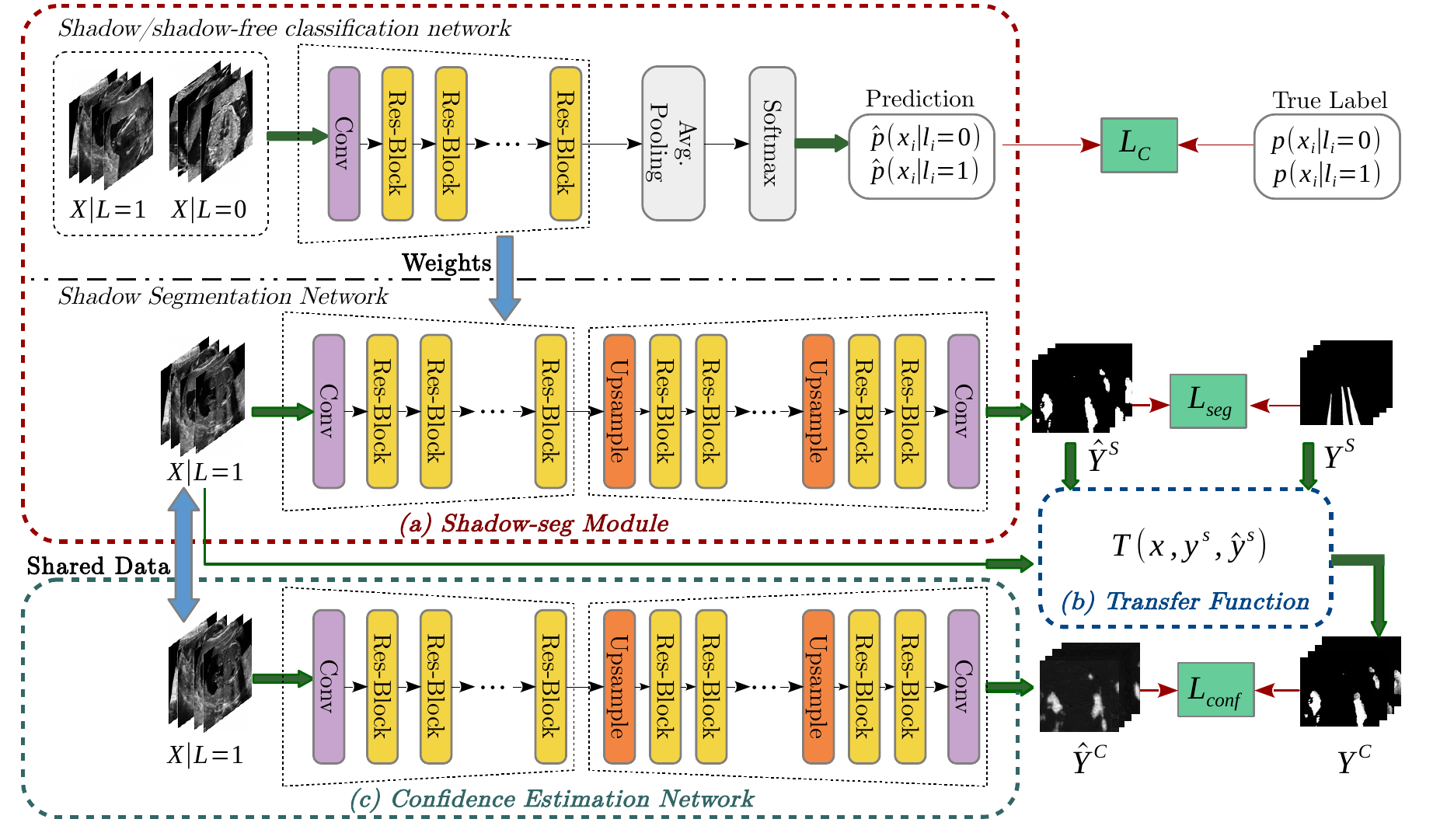}
 \caption{Training framework of the proposed method. (a) The shadow-seg module containing a shadow/shadow-free classification network and a shadow segmentation network. (b) The transfer function that expands a binary mask to a reference confidence map. (c) The confidence estimation network which establishes direct mapping between input images and confidence maps.}
 \label{overview}
\end{figure*}

%The proposed method is to generate a dense confidence map for shadow regions in 2D ultrasound from very weak annotations. 
In our proposed method, a shadow-seg module is first trained to produce a semantic segmentation of shadow regions. In this module, shadow features are initialized by training a shadow/shadow-free classification network and generalized by training a shadow segmentation network. After obtaining the shadow segmentation, a transfer function is used to extend the predicted binary shadow segmentation to a confidence map based on the intensity distribution within suspected shadow regions. This confidence map is regarded as a reference confidence map for the next confidence estimation network. Lastly, a confidence estimation network is trained to learn the mapping between the input shadow-containing US images and the corresponding reference confidence maps. The outline for the training process is shown in Fig.~\ref{overview}. 
During inference,  %(shown in Fig.~\ref{inference}), 
we use the confidence estimation network to predict a dense, shadow confidence map directly from the input image. Additionally, we integrate attention mechanisms~\cite{shen2018} into our method to enhance the shadow features extracted by the networks.  

%\subsection{Shadow-seg Module}
\subsubsection*{\textbf{Shadow-seg Module}}
We propose a shadow-seg module to extract generalized shadow features for a large range of shadow types in fetal US images under limited weak manual annotations. %, and thus, achieve higher shadow segmentation accuracy. 
Since shadow regions have different shapes, various intensity distributions and uncertain edges, the pixel-wise annotation of shadow regions is time consuming and relies heavily on annotator's experience (e.g. various annotations in Fig.~\ref{DataPresentation}(b)). This generally results in manual annotations of limited quantity and quality. 
%in a limited number of collected pixel-wise shadow annotations, which in addition have a coarse quality. 
Compared with pixel-wise shadow annotations, global image-level labels (``has shadow" and ``shadow-free" in our case) are easier to obtain, and shadow images with global image-level labels can contain a larger variety of shadow types. Therefore, we use a shadow-seg module that combines unreliable pixel-wise annotations and global image-level labels as weak annotations.%to extract refined shadow representations to obtain accurate shadow segmentation. 
The proposed shadow-seg module contains two tasks, (1) shadow/shadow-free classification using image-level labels, and (2) shadow segmentation that uses few coarse pixel-wise manual annotations ($10\%$ of the global image-level labels). 
% Shadow features are generalized in this module by sharing weights between feature encoders of both tasks. By sharing weights, 
Shadow features can be extracted during simple shadow/shadow-free classification and subsequently optimized for the more challenging shadow segmentation task. In our case, shadow features extracted by the classification network cover various shadow types in a range of anatomical structures. These shadow features become suitable for the shadow segmentation after being optimized by a shadow segmentation network.  

\begin{figure}[htb]
 \centering
 \includegraphics[width=0.5\textwidth, trim=7cm 10.15cm 7cm 6.8cm, clip]{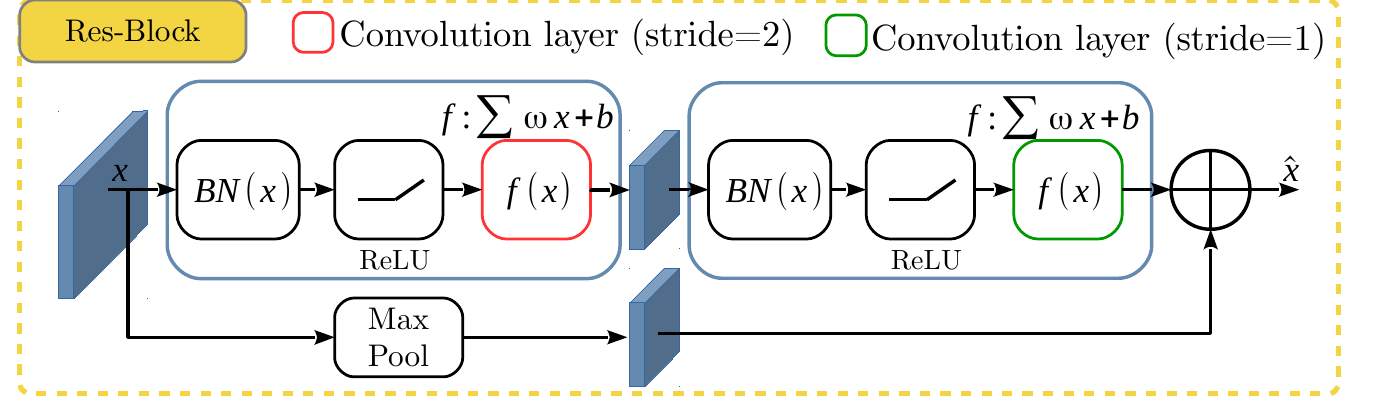}
 \caption{The architecture of the residual-block. $BN(x)$ refers to a batch normalization layer and $f(x)$ is a convolutional layer.}
 \label{resB}
  % \vspace{-0.4cm}
\end{figure}

\subsubsection*{\textbf{Network Architecture}} We build two sub-networks from residual-blocks~\cite{he2016identity} as shown in Fig.~\ref{resB}. Residual-blocks can reduce the training error when using deeper networks and support better network optimization~\cite{he2016identity}. They have been widely used for various image processing algorithms~\cite{Zhu2017,he2016,zhang2018}. 
The first and initially trained network is a shadow/shadow-free classification network that learns to distinguish images containing shadows from shadow-free images, and thus learns the defining features of acoustic shadow. This classification network consists of a feature encoder followed by a global average pooling layer. The feature encoder uses six residual-blocks (Fig.~\ref{resB}) to extract shadow features that define shadow-containing images in the classifier. We refer to $l=1$ as the label of the shadow-containing class and $l=0$ as the label of the shadow-free class. Image set $X^C=\{x_1^C,x_2^C,...,x_K^C\}$ and their corresponding labels $L=\{l_1,l_2,...,l_K\} \text{ s.t.} l_i\in\{0,1\}$ are used to train the feature encoder as well as the global average pooling layer. We use softmax cross-entropy loss as the cost function $L_C$ between the predicted labels and the true labels. 
% The detailed architecture of the classification network is shown in \cref{appdclassN}. 

Representative shadow features extracted by the feature encoder of the shadow/shadow-free classification network are then optimized by the shadow segmentation network with a limited number of densely segmented US images. 
The feature encoder of the segmentation network has the same architecture as the classification network. The weights of the feature encoder in the segmentation network are initialized by that of the classification network and are further fine-tuned for the segmentation task. Therefore, the extracted shadow features are suitable for the segmentation in addition to classification.
The decoder of the segmentation network is symmetrical to the feature encoder. Feature layers from the feature encoder are concatenated to the corresponding layers in the decoder by skip connections. Here, we denote the image set used to train the shadow segmentation with $X^S=\{x_1^S,x_2^S,...,x_M^S\}$ and the corresponding pixel-wise manual segmentation with $Y^S=\{y_1^S,y_2^S,...,y_M^S\}$. The shadow segmentation provides a pixel-wise binary prediction $\hat{Y}^S=\{\hat{y}_1^S,\hat{y}_2^S,...,\hat{y}_M^S\}$ for shadow regions and the cost function $L_{seg}$ is the softmax cross-entropy between $\hat{Y}^S$ and $Y^S$.
% (Eq.~\ref{Loss_seg}). 
% The detailed architecture of the segmentation network is shown in \cref{appdsegN}.
% \begin{equation}\label{Loss_seg}
% \mathcal{L}_{seg} = 
% \lVert \hat{Y}^S - Y^S \rVert_2.
% \end{equation}

%\subsection{Transfer Function}
\subsubsection*{\textbf{Transfer Function}}
Binary masks lack information about inherent uncertainties at the boundaries of shadow regions. Therefore, we use a transfer function to extend the binary segmentation prediction to a confidence map, which is more appropriate to describe shadow regions. The main task of the transfer function is to learn the intensity distribution of shadow regions so as to estimate confidence of pixels in false positive (FP) regions of the predicted binary shadow segmentation. 
This transfer function is built and only used during training to provide reference confidence maps for the confidence estimation network.
% In detail, we build a matrix $T(\cdot)$ to describe the intensity distribution of shadow regions. A transfer network is then proposed to learn the continuous regression of the matrix using CNNs, which enables our method end-to-end trainable.

% \subsubsection*{Matrix} 
When comparing the manual segmentation $y^S$ and the predicted segmentation $\hat{y}^S$ of shadow regions in image $x$, we define the true positive (TP) regions $x_{TP}$ as shadow regions with the full confidence, $C_{x_{ij}} = 1, x_{ij}\in x_{TP}$. Here, $C_{x_{ij}}$ refers to the confidence of pixel $x_{ij}$ being shadow.
% \begin{equation}\label{conf_TP}
% C_{x_{ij}} = 1,\quad x_{ij}\in x_{TP}.
% \end{equation}

For each pixel $x_{ij}$ in the FP regions ($x_{FP}$), the confidence of belonging to a shadow region is computed by a transfer function $T(x_{ij} \mid x_{ij}\in x_{FP})$ based on the intensity of the pixel ($I_{x_{ij}}$) and the mean intensity of $x_{TP}$ ($I_{mean}$). $I_{mean}$ is defined in Eq.~\ref{Imean}. With weak signals in the shadow regions, the average intensity of shadow pixels is lower than the maximum intensity ($I_{max}=max(x)$) but not lower than the minimum intensity ($I_{min}=min(x)$), that is $I_{mean}\in[I_{min}, I_{max})$. 
%$I_{max}$ and $I_{min}$ are shown in Eq.~\ref{min_max}, and 
% \begin{equation}\label{min_max}
% I_{min} = min(x), \quad I_{max} = max(x).
% \end{equation}

\begin{equation}\label{Imean}
I_{mean} = 
\begin{cases}
    mean(y^S \cap \hat{y}^S) &  y^S \cap \hat{y}^S \neq \emptyset,\\
    mean(y^S)                & y^S \cap \hat{y}^S = \emptyset,
\end{cases}
\end{equation}

The transfer function $T(\cdot)$ computing $C_{x_{ij}}$ for pixels in $x_{FP}$ is defined according to the range of $I_{mean}$. For $I_{mean}\in(I_{min}, I_{max})$, $T(\cdot)$ is shown in Eq.~\ref{T_1}. For $I_{mean} = I_{min}$, $T(\cdot)$ is shown in Eq.~\ref{T_2}. 
% $T(x_{ij} \mid x_{ij}\in x_{FP}) = 1$ if $I_{x_{ij}} = I_{mean}$, otherwise $T(x_{ij} \mid x_{ij}\in x_{FP}) = \frac{I_{x_{ij}} - I_{mean}}{I_{max} - I_{mean}}$. 
% T(x_{ij}\in x_{FP} \mid x,y^S,\hat{y}^S)
\begin{equation}\label{T_1}
T(x_{ij} \mid x_{ij}\in x_{FP}) = 
\begin{cases}
    \frac{I_{x_{ij}} - I_{min}}{I_{mean} - I_{min}}, &  I_{min} \leq I_{x_{ij}} < I_{mean},\\
    \frac{I_{max} - I_{x_{ij}}}{I_{max} - I_{mean}}, &  I_{mean} < I_{x_{ij}} \leq I_{max},\\
    1,                                                &  I_{x_{ij}} = I_{mean},
\end{cases}
\end{equation}
\begin{equation}\label{T_2}
T(x_{ij} \mid x_{ij}\in x_{FP}) = 
\begin{cases}
    \frac{I_{x_{ij}} - I_{mean}}{I_{max} - I_{mean}}, &  I_{mean} < I_{x_{ij}},\\
    1,                                                &  I_{x_{ij}} = I_{mean},
\end{cases}
\end{equation}

After using the transfer function, the binary map of the predicted segmentation $y^S$ is extended to a confidence map $y^C$. $y^C$ acts as a reference ("ground truth") for the training of the next confidence estimation network.

% After using the matrix, the binary map of the predicted segmentation $y^S$ is extended to a confidence map $y^T$. $y^T$ acts as a ground truth for the training of the transfer network.

%\subsubsection*{Transfer Matrix Generalization} 

%the explicit way to define the integration of shadow intensity information. Continuous functions such as quadratic or Gaussian functions are able to approximate $T(\cdot)$.
%A continuous approximation allows to refine network (c) and (a) in a final step an enables gradient error information to flow backwards through the networks. 
%An alternative way is to utilize CNNs to learn intensity distribution modelled by $T(\cdot)$. 
%By using continuous functions or CNNs to integrating intensity information, the segmentation network and the confidence estimation network can be further jointly fine-tuned.

% \subsubsection*{Network} The transfer network contains a feature encoder with four residual-blocks and a symmetric decoder. Feature layers are concatenated between encoder and decoder. The pixel-wise manual segmentation $Y^S$ is used to train the transfer network according to the ground truth $Y^T$. The cost function $\mathcal{L}_{trans} = 
% \lVert Y^C - Y^T \rVert_2$ is the mean squared error between the prediction $Y^C$ and the ground truth $Y^T$. $Y^C$ is regarded as the reference for training the next confidence estimation network. 

% \begin{equation}\label{Loss_trans}
% \mathcal{L}_{trans} = 
% \lVert Y^C - Y^T \rVert_2.
% \end{equation}

%\subsection{Confidence Estimation Network}
\subsubsection*{\textbf{Confidence Estimation Network}}
After obtaining reference confidence maps from the predicted binary segmentation, a confidence estimation network is trained to map an image with shadows ($x$) to the corresponding reference confidence map ($y^C$). This confidence estimation network can be independently used to directly predict a dense shadow confidence map for an input image during inference. 

The confidence estimation network consists of a down-sampling encoder, a symmetric up-sampling decoder, and skip connections between feature layers from the encoder and the decoder at different resolution levels. Both the encoder and the decoder are composed of six residual-blocks.
% , which reduces the training error of using a deeper network and results in easier network optimization \cite{he2016identity}. 
The cost function of the confidence estimation network is defined as the mean squared error between the predicted confidence map $\hat{Y}^C$ and the reference confidence map $Y^C$ ($\mathcal{L}_{conf} = 
\lVert \hat{Y}^C - Y^C \rVert_2$).
% (shown in Eq.~\ref{Loss_conf}). 
% The detailed architecture of the confidence estimation network is shown in \cref{appdconfN}.
% \begin{equation}\label{Loss_conf}
% \mathcal{L}_{conf} = 
% \lVert \hat{y}^C - y^C \rVert_2.
% \end{equation}

%\subsection{Attention Gates}
\subsubsection*{\textbf{Attention Gates}}
\begin{figure}[t]
 \centering
 \includegraphics[width=0.5\textwidth, trim=7.5cm 9.5cm 7cm 8cm, clip]{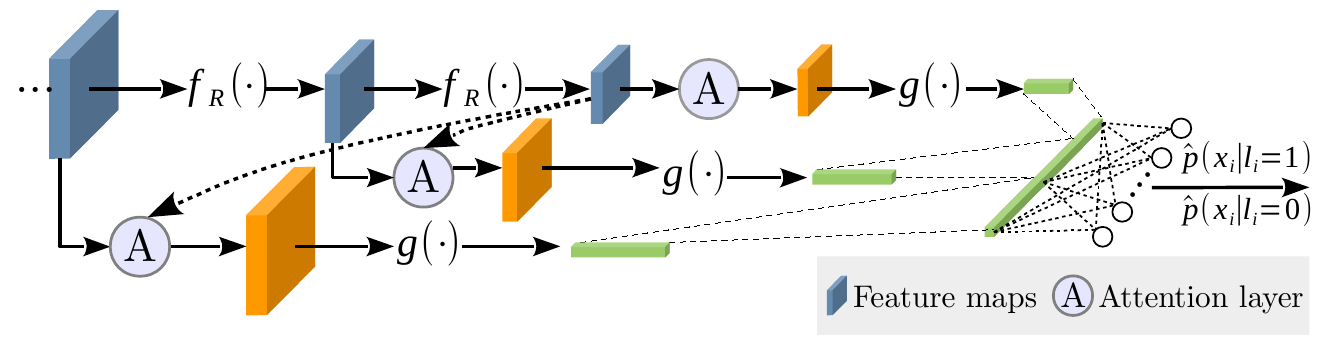}
 \caption{The architecture of the shadow/shadow-free classification network with attention mechanism. $f_R(\cdot)$ refers to residual-blocks. $g(\cdot)$ refers to a global average pooling layer.}
 \label{AGs}
  %\vspace{-0.4cm}
\end{figure}

Attention gates are believed to generally highlight relevant features according to image context and thus improve network performance for medical image analysis~\cite{ozan2018}. We integrate attention gates~\cite{shen2018} into our approach to explore if attention mechanisms can further improve the confidence estimation of shadow regions in 2D ultrasound. 
% Since attention mechanisms are able to highlight relevant features according to image context, these mechanisms can be used to guide the networks to focus on shadow features. 
% In our case, we integrate a self-attention gating module as proposed in~\cite{ozan2018} into all three networks. 
In our case, we connect the self-attention gating modules proposed in~\cite{ozan2018} to the feature maps before the last two down-sampling operations in the encoders of all three networks. For the shadow/shadow-free classification network, the global average pooling layer is modified when adding this self-attention gating module. In detail, as shown in Fig.~\ref{AGs}, 
% after extracting feature maps from all resolution levels and applying the self-attention gating module to the last two resolution levels, 
the global average pooling layers are operated separately on the two attention-gated feature maps as well as the original last feature map to obtain three average feature maps. These three average feature maps are then concatenated, followed by a fully connected layer to compute the final classification prediction.

\section{Implementation}

All the residual-blocks used in the proposed method are implemented as proposed in~\cite{pawlowski2017state}, which provides a convenient interface to realize residual-blocks.% and can be utilized conveniently.

We optimize the different modules separately and consecutively in three steps. First we train $\sim 70$ epochs for the parameters of the shadow/shadow-free classification network, and then $\sim700$ epochs for the pixel-wise shadow segmentation network.
After obtaining a well-trained shadow segmentation network, we train the confidence estimation network for another 700 epochs. 
% After obtaining a well-trained shadow segmentation network, we train the transfer network for 250 epochs, followed by another 700-epoch updates of the confidence estimation network. 

For all networks, we use Stochastic Gradient Descent (SGD) with momentum optimizer to update the parameters since SGD has better generalization capability than adaptive optimizer~\cite{wilson2018}. The parameters of the optimizer are $momentum=0.9$, with a learning rate of $10^{-3}$. We apply L2 regularization to all weights during training to help prevent network over-fitting. The scale of the regularizer is set as $10^{-5}$. The training batch size is 25 and our networks are trained on a Nvidia Titan X GPU with 12 GB of memory.

\section{Evaluation}
The proposed method is trained and evaluated using two data sets, (1) a multi-class data set consisting of 13 classes of 2D US fetal anatomy with global image-level label (``has shadow" or ``shadow-free") and including 48 non-brain images with manual shadow segmentations, and (2) a single-class data set containing 2D US fetal brain with coarse pixel-wise manual shadow segmentations. To reduce the variance in parameter estimation during training, we split relatively bigger training data sets. In the multi-class data set, we use $88\%$ of the data for training, $11\%$ for validation and the 48 non-brain images for testing, while in the single-class data set we use $78\%$ of the data for training, $8\%$ for validation and $14\%$ for testing. 
% Fig.~\ref{Dataset} shows an overview over our data split strategy.

% \begin{figure}[htb]
%  \centering
%  \includegraphics[width=0.5\textwidth, trim=11.2cm 9cm 7.8cm 8.3cm, clip]{dataset.pdf}
%  \caption{Visual representation of data sets used in evaluations.}
%  \label{Dataset}
% \end{figure}

To verify the effectiveness of the proposed method and the importance of the shadow/shadow-free classification network in the shadow-seg module, we compare the variants of our method to a baseline which only contains a shadow segmentation network and a confidence estimation network. %The baseline is described in detail in Part.$B$ of this section.

We use standard measurements such as Dice coefficient (DICE)~\cite{DiceLR1945}, recall, precision and Mean Squared Error (MSE) for shadow segmentation evaluation, and use the Interclass Correlation (ICC)~\cite{shrout1979} as well as soft DICE~\cite{anbeek2005} for confidence estimation evaluation. In order to verify the performance of our method, we also compute quantitative measurements between the chosen manual annotation (weak ground truth) and another manual annotation from a different annotator to show the human performance for the shadow detection task. Lastly, we show the practical benefits of shadow confidence maps for different applications such as a standard plane classification task, an image fusion task from multiple views and a segmentation task for automatic biometric measurements.

% In this section, all proposed methods are build with $T(\cdot)$ as defined in Sec.\Romannum{2} Part.$B$ by default. 

%\subsection{Data Sets}
%\subsubsection*{\textbf{Data Sets}}
\subsubsection*{\textbf{Multi-class Data Set}} This data set consists of $\sim 8.5k$ 2D fetal US images sampled from 13 different anatomical standard plane locations as defined in the UK FASP handbook~\cite{screening2015}. These images have been sampled from 2694 2D ultrasound examinations from volunteers with gestational ages between $18-22$ weeks (iFIND Project~\footnote{http://www.ifindproject.com/ \label{ifindfoot}}). Eight different ultrasound systems of identical make and model (GE Voluson E8) were used for the acquisitions. Various image settings based on different sonographers' personal preference for scanning are included in this data set. The images have been classified by expert observers as containing strong shadow, being shadow-free, or being corrupted, e.g. poor tissue contact caused by lacking acoustic impedance gel. Corrupted images ($<3\%$) have been excluded as discussed in Section VI with Fig.~\ref{InsufficientShadow}.

\subsubsection*{\textbf{Single-class Data Set}} This data set comprises 643 fetal brain images and has no overlap with the multi-class data set. Shadow regions in this data set have been coarsely segmented by two bio-engineering students using  trapezoid-shaped segmentation masks for individual shadow regions.

\subsubsection*{\textbf{Training Data}} 3448 shadow images and 3842 clear images have been randomly selected from  
the multi-class data set to train the shadow/shadow-free classification network. 500 fetal brain images have been randomly chosen from the single-class data set to train the shadow segmentation network, and the confidence estimation network. These 500 fetal images have been flipped as data augmentation during training.

\subsubsection*{\textbf{Validation and Test Data}}
The remaining 491 shadow images and 502 clear images in the multi-class data set are used for testing and validation. Here, a subset ($M_{test}$) comprising 48 randomly selected images from the 491 shadow images is used for testing. These 48 images contain various fetal anatomies (except fetal brain), such as abdominal, kidney, cardiac and etc. Shadow regions in these images have been manually segmented to provide ground truth. The remaining 443 shadow images and 502 clear images are used for the validation of the shadow/shadow-free classification. Similarly, the remaining 143 fetal brain images of the single-class data set are split into two subsets, where $S_{val}$ contains 50 images for validation of the shadow segmentation, binary-to-confidence transformation and the confidence estimation, and $S_{test}$ with 93 images for testing. For all images from the single-class data set, we randomly choose one group of annotations from two different existing groups of annotations as ground truth for training, validation and testing.

%Fig.~\ref{Dataset} shows an overview over the data split.% for visual understanding of the relationships between aforementioned data sets.

\subsection{Baseline}
%\subsubsection*{\textbf{Baseline}}
The baseline method is used to demonstrate that the shadow-seg module is of importance for capturing generalized shadow features and obtaining accurate confidence estimation of shadow regions.  
It comprises a shadow segmentation network and a confidence estimation network, which have the same architectures as shown in Fig.~\ref{overview}. 
% Similar to the training of the proposed method, 
We firstly train the shadow segmentation network in the baseline method using the 500 fetal brain images from the single-class data set. 
% After applying the transfer function on the predicted binary masks of the shadow segmentation, 
After applying the transfer function on the binary segmentation prediction,
we train the confidence estimation network for a direct mapping between shadow images and reference confidence maps.

\subsection{Evaluation Metrics}
%\subsubsection*{\textbf{Evaluation Metrics}}
In this section, we define the aforementioned statistical metrics
% including DICE, recall, precision, MSE, ICC and soft DICE, 
and the computation of the inter-observer variability between two pixel-wise manual annotations of shadow regions. 

\subsubsection*{DICE, Recall, Precision and MSE} We refer to the binary prediction of shadow segmentation as $P$ and the binary manual segmentation as $G$. $\text{DICE} = 2|P \cap G|/(|P|+|G|)$, $\text{Recall}=|P \cap G|/|G|$, $\text{Precision}=|P \cap G|/|P|$ and $\text{MSE} = |P - G|$.
% DICE coefficient is shown in Eq.~\ref{dice}. Recall and precision is defined in Eq.~\ref{RP} and $\text{MSE} = |P - G|$. Here, both $P$ and $G$ are binary masks.
% \begin{equation}\label{dice}
% \text{DICE} = \frac{2|P \cap G|}{|P|+|G|}.
% \end{equation}

% \begin{equation}\label{RP}
% \text{Recall} = \frac{|P \cap G|}{|G|}, ~\text{Precision} = \frac{|P \cap G|}{|P|}.
% \end{equation}

\subsubsection*{ICC} We use $ICC$ as proposed by~\cite{shrout1979} (Eq.~\ref{ICC21}) to measure the agreement between two annotations. Each pixel in an image is regarded as a target. $R_{MS}$, $C_{MS}$ and ${M}_{MS}$ are respectively mean squared value of rows, columns and interaction. $N$ is the number of targets.
\begin{equation}\label{ICC21}
ICC = \frac{R_{MS} - {M}_{MS}}{R_{MS} + {M}_{MS} + 2 \times (C_{MS} - {M}_{MS}) / N}.
\end{equation}

\subsubsection*{Soft DICE} Soft DICE can be used to tackle probability maps. We use real-value in the DICE definition to compute soft DICE between the predicted shadow confidence maps $\hat{Y}^C$ and reference confidence maps $Y^C$.
% \subsubsection*{Soft DICE} Since soft DICE can be used to tackle probability maps, we use float value in DICE definition for soft DICE computation to evaluate the performance of the transfer network (soft DICE between $Y^C$ and $Y^T$) and the performance of the confidnece estimation network (soft DICE between $\hat{Y}^C$ and $Y^C$).

\subsubsection*{Human Performance} We consider another binary segmentation of shadow regions from a different annotator as $Y^S_{new}$. The computed metrics between $Y^S_{new}$ and the chosen manual segmentation $Y^S$ reflects the human inter-observer variability.

\subsection{Shadow Segmentation Analysis}
%\subsubsection*{\textbf{Shadow Segmentation Analysis}}
We compare the segmentation performance of the state-of-the-art  (\cite{Karamalis2012} and \cite{meng2018}), the proposed methods and the human performance. This comparison is used to examine the importance of the shadow-seg module for the shadow segmentation, and further, for the confidence estimation of shadow regions.

\begin{table}[t]
\centering
\caption{Shadow segmentation performance ($\mu \pm \sigma$) of different methods on test data $S_{test}$. RW and $\text{RW}^\ast$ are Random Walk algorithm~\cite{Karamalis2012} with different set of parameters. Pilot~\cite{meng2018} is our previous work. Baseline, the proposed method (abbreviated as ``Proposed") and the proposed method with attention gates (abbreviated as ``Proposed$+$AG" in the rest of the paper) are our proposed methods. $\text{Anno}^\ast$ refers to the human inter-observer variability, thus expected human performance on the shadow segmentation task. Best results are shown in bold.}
\label{B_data_comp}
\begin{tabular}{*6c}
\toprule
Methods                                               & 
~~~~~                                               &
DICE                                                & 
Recall                                              & 
Precision                                           & 
MSE                                                 \\ 
\midrule
\multirow{2}{*}{RW~\cite{Karamalis2012}}                           &
$\mu$                                        &
0.2096                                    &
0.6535                                   &     
0.1288                                   &
194.8618                                 \\
~~~~~~~                                  &
($\sigma$)                                        &
(0.099)                                    &
(0.2047)                                   &     
(0.0675)                                   &
(7.6734)                                 \\
\cmidrule{2-6}
\multirow{2}{*}{$\text{RW}^\ast$~\cite{Karamalis2012}}                      & 
$\mu$                                        &
0.231                                   & 
0.6921                                   & 
0.1432                                   & 
189.0828                                  \\
~~~~~~~                                  &
($\sigma$)                                        &
(0.1123)                                   & 
(0.2196)                                   & 
(0.0771)                                   & 
(8.3484)                                 \\
\cmidrule{2-6}
\multirow{2}{*}{Pilot~\cite{meng2018}}                      & 
$\mu$                                        &
0.3227                                   & 
0.4275                                   & 
0.2863                                   & 
110.2959                                  \\
~~~~~~~                                  &
($\sigma$)                                        &
(0.1398)                                   & 
(0.201)                                   & 
(0.1352)                                   & 
(14.837)                                 \\
\midrule
\multirow{2}{*}{Baseline}                           &
$\mu$                                        &
0.6933                                    &
0.6884                                   &     
0.7246                                   &
60.3680                                 \\
~~~~~~~                                  &
($\sigma$)                                        &
(0.212)                                    &
(0.2255)                                   &     
(0.2326)                                   &
(12.2885)                                 \\ 
\cmidrule{2-6}
\multirow{2}{*}{Proposed}                          &
$\mu$                                        &
\textbf{0.7167}                                   & 
\textbf{0.7217}                                   & 
\textbf{0.7382}                                   & 
\textbf{58.6974}                                  \\
~~~~~~~                                  &
($\sigma$)                                        &
(0.1988)                                   & 
(0.2131)                                   & 
(0.2255)                                   & 
(11.867)                                  \\
\cmidrule{2-6}
\multirow{2}{*}{Proposed$+$AG}                      & 
$\mu$                                        &
0.7027                                   & 
0.7199                                   & 
0.7132                                   & 
61.241                                  \\
~~~~~~~                                  &
($\sigma$)                                        &
(0.2014)                                   & 
(0.2169)                                   & 
(0.2247)                                   & 
(12.6317)                                 \\
\midrule
\multirow{2}{*}{$\text{Anno}^\ast$}                &
$\mu$                                        &
0.5443                                   & 
0.6126                                   & 
0.567                                    & 
65.7286                                 \\
~~~~~~~                                  &
($\sigma$)                                        &
(0.2635)                                   & 
(0.3196)                                   & 
(0.3124)                                    & 
(23.0339)                                 \\
\bottomrule
\end{tabular}
\end{table}

Table~\ref{B_data_comp} shows DICE, recall, precision and MSE of different methods on $S_{test}$. RW and $\text{RW}^\ast$ are results of \cite{Karamalis2012} with various parameters. For fair comparison, we run 24 tests on both test sets using the RW algorithm with different parameter combinations ($\alpha\in\{1,2,6\}$; $\beta\in\{90,120\}$; $\gamma\in\{0.05,0.1,0.2,0.3\}$). With a negative relationship between the likelihood of shadows and the confidence in \cite{Karamalis2012} and to consistently compare all methods, we use $1-S$ instead $S$ to display the results of RW and $\text{RW}^\ast$ in all comparison experiments. Here $S$ is a confidence map obtained by \cite{Karamalis2012}. To generate shadow segmentation, we threshold the obtained confidence maps by $T\in\{0.25,0.3\}$ so that pixels with confidence higher than $T$ are shadows. We chose the parameters and the threshold which achieve the highest average DICE on all samples in both test sets. The chosen RW parameters and the threshold are $\alpha=1$; $\beta=90$; $\gamma=0.3$; $T=0.3$. We also applied the parameters and the threshold in \cite{Karamalis2012} ($\alpha=2$; $\beta=90$; $\gamma=0.05$; $T=0.25$) in our experiments, which is denoted as $\text{RW}^\ast$.  Note that we use the public Matlab code~\footnote{http://campar.in.tum.de/Main/AthanasiosKaramalisCode} of~\cite{Karamalis2012} to test RW and $\text{RW}^\ast$.
% The baseline, the proposed method and the proposed method with attention gates (abbreviated as ``the proposed$+$AG" in the rest of the paper) are our proposed methods. Pilot refers to our previous work in \cite{meng2018} and $\text{Anno}^\ast$ represents the human performance on the shadow segmentation task.

As shown in Table~\ref{B_data_comp}, the baseline, the proposed method and the proposed$+$AG greatly outperform the state-of-the-art. Among all methods, the proposed method achieves highest DICE.
% between the binary masks of the predicted shadow segmentation and the manual segmentation.
Recall and precision of the proposed method are respectively $3.33\%$ and $1.16\%$ higher than that of the baseline while MSE of the proposed method is $1.67$ lower than that of the baseline. 
After adding attention gates to the proposed method (the proposed$+$AG), the shadow segmentation performance is nearly the same to the proposed method without attention gates, but better than the baseline. Additionally, the relatively low scores of $\text{Anno}^\ast$ indicate high inter-observer variability and how ambiguous human annotation can be for this task. A mean DICE of $0.7167$ shows that the proposed method performs better and more consistently than human annotation.

% \begin{figure}[t]
%  \centering
%  \includegraphics[width=0.5\textwidth, trim=1.7cm 0cm 1.5cm 0.4cm, clip]{C48SegAnalysis.eps}
%  \caption{Comparison of segmentation performance of different methods on test data $M_{test}$. The bar charts show the mean value of DICE, recall, precision and MSE of the state-of-the-art methods as well as the proposed methods.}
%  \label{C48SegAna}
% \end{figure}

\begin{table}[t]
\centering
\begin{threeparttable}
\caption{Comparison of shadow segmentation performance ($\mu \pm \sigma$) of different methods on test data $M_{test}$. Best results are shown in bold.}
\label{C48_data_comp}
\begin{tabular}{*6c}
\toprule
Methods                                               & 
~~~~~                                               &
DICE                                                & 
Recall                                              & 
Precision                                           & 
MSE                                                 \\ 
\midrule
\multirow{2}{*}{RW~\cite{Karamalis2012}}                           &
$\mu$                                        &
0.1795                                    &
\textbf{0.8456}                                   &     
0.1038                                   &
193.2229                                 \\
~~~~~~~                                  &
($\sigma$)                                        &
(0.0855)                                    &
(0.1241)                                   &     
(0.0592)                                   &
(7.866)                                 \\
\cmidrule{2-6}
\multirow{2}{*}{$\text{RW}^\ast$~\cite{Karamalis2012}}                      & 
$\mu$                                        &
0.1766                                   & 
0.8038                                   & 
0.1025                                   & 
190.5627                                  \\
~~~~~~~                                  &
($\sigma$)                                        &
(0.0871)                                   & 
(0.1528)                                   & 
(0.0602)                                   & 
(7.5643)                                 \\
\cmidrule{2-6}
\multirow{2}{*}{Pilot~\cite{meng2018}}                      & 
$\mu$                                        &
0.467                                   & 
0.728                                   & 
0.371                                   & 
86.9005                                  \\
~~~~~~~                                  &
($\sigma$)                                        &
(0.1079)                                   & 
(0.137)                                   & 
(0.1308)                                   & 
(17.0491)                                 \\
\midrule
\multirow{2}{*}{Baseline}                           &
$\mu$                                        &
0.4765                                    &
0.5026                                   &     
0.5108                                   &
68.5054                                 \\
~~~~~~~                                  &
($\sigma$)                                        &
(0.1798)                                    &
(0.2233)                                   &     
(0.1712)                                   &
(18.3773)                                 \\ 
\cmidrule{2-6}
\multirow{2}{*}{Proposed}                          &
$\mu$                                        &
\textbf{0.5463}                                   & 
0.5968                                   & 
\textbf{0.565}                                   & 
\textbf{64.6912}                                  \\
~~~~~~~                                  &
($\sigma$)                                        &
(0.155)                                   & 
(0.2335)                                   & 
(0.1357)                                   & 
(17.2147)                                  \\
\cmidrule{2-6}
\multirow{2}{*}{Proposed$+$AG}                      & 
$\mu$                                        &
0.5302                                   & 
0.5741                                   & 
0.5454                                   & 
66.4474                                  \\
~~~~~~~                                  &
($\sigma$)                                        &
(0.1544)                                   & 
(0.2035)                                   & 
(0.1562)                                   & 
(17.6628)                                 \\
\bottomrule
\end{tabular}
\begin{tablenotes}
\item The symbols of the methods are the same to Table~\ref{B_data_comp}.
\end{tablenotes}
\end{threeparttable}
\end{table}

We further conduct the same experiments on another non-brain test data set $M_{test}$ to verify the feature generalization ability of the shadow-seg module.
% $M_{test}$ contains various fetal anatomies (except fetal brain), such as abdomen, kidney, cardiac and etc. 
% Results of the same evaluation metrics are shown in Table.~\ref{C48_data_comp}. 
Results are shown in Table~\ref{C48_data_comp}.
Similarly, the proposed weakly supervised methods and the baseline outperform all state-of-the-art methods.
%The proposed method as well as the proposed$+$AG are quantitatively better than the baseline on all evaluation metrics, which indicates that the shadow/shadow-free classification network contributes greatly to obtaining more general shadow features.

To statistically evaluate the difference among various methods, we use the paired sample t-test on two test data sets $S_{test}$ and $M_{test}$. Here, we compare the evaluation metrics (Dice, Recall, Precision and MSE) of the proposed method and the Pilot~\cite{meng2018} because the Pilot~\cite{meng2018} outperforms other state-of-the-art in Table~\ref{B_data_comp} and Table~\ref{C48_data_comp}. We also compare the evaluation metrics of the proposed method and the baseline. The obtained corresponding p-values are shown in Table~\ref{P_value_seg}, using $0.01$ as the threshold for statistical significance, Table~\ref{P_value_seg} shows that the proposed method greatly improves the shadow segmentation performance compared with the Pilot~\cite{meng2018} and the baseline. 

\begin{table}[t]

\caption{The p-value of the Proposed method vs. Pilot~\cite{meng2018} and of the Proposed method vs. Baseline. Statistically significant results ($p<0.01$) are shown in bold.}
\label{P_value_seg}
\centering
\begin{threeparttable}
\begin{tabular}{*5c}
\toprule
~~~~~~  &
\multicolumn{4}{c}{$S_{test}$}                           \\
\cmidrule{2-5}
~~~~~~~  &
DICE     &
Recall   &
Precision &
MSE    \\
Pilot~\cite{meng2018}     &
\textbf{0.0001}   &
\textbf{0.0001} &
\textbf{0.0001}    &
\textbf{0.0001}    \\
Baseline     &
\textbf{0.0015}   &
\textbf{0.001} &
0.0694    &
0.012   \\
\hline
\hline
~~~~~~  &
\multicolumn{4}{c}{$M_{test}$}          \\
\cmidrule{2-5}
~~~~~~~  &
DICE     &
Recall   &
Precision &
MSE    \\
Pilot~\cite{meng2018}     &
\textbf{0.0032}   &
\textbf{$\text{0.0013}^\dag$} &
\textbf{0.0001}    &
\textbf{0.0001}    \\
Baseline     &
\textbf{0.0001}  &
\textbf{0.0001} &
\textbf{0.0037}    &
\textbf{0.0014}   \\
\bottomrule
\end{tabular}
\begin{tablenotes}
% \item [$\ast$] The two-tailed P value with the symbol $\dag$ refers to the proposed method performs worse and otherwise the proposed method is better.
\item [$\dag$] refers to the proposed method performs worse and otherwise the proposed method is better.
\end{tablenotes}
\end{threeparttable}
\end{table}

\subsection{Shadow Confidence Estimation}
%\subsubsection*{\textbf{Shadow Confidence Estimation}}
In this part, we evaluate the performance of the confidence estimation by comparing the shadow confidence maps of different methods.

\begin{figure}[t]
 \centering
 \includegraphics[width=0.5\textwidth, trim=1cm 10.8cm 1cm 10.5cm, clip]{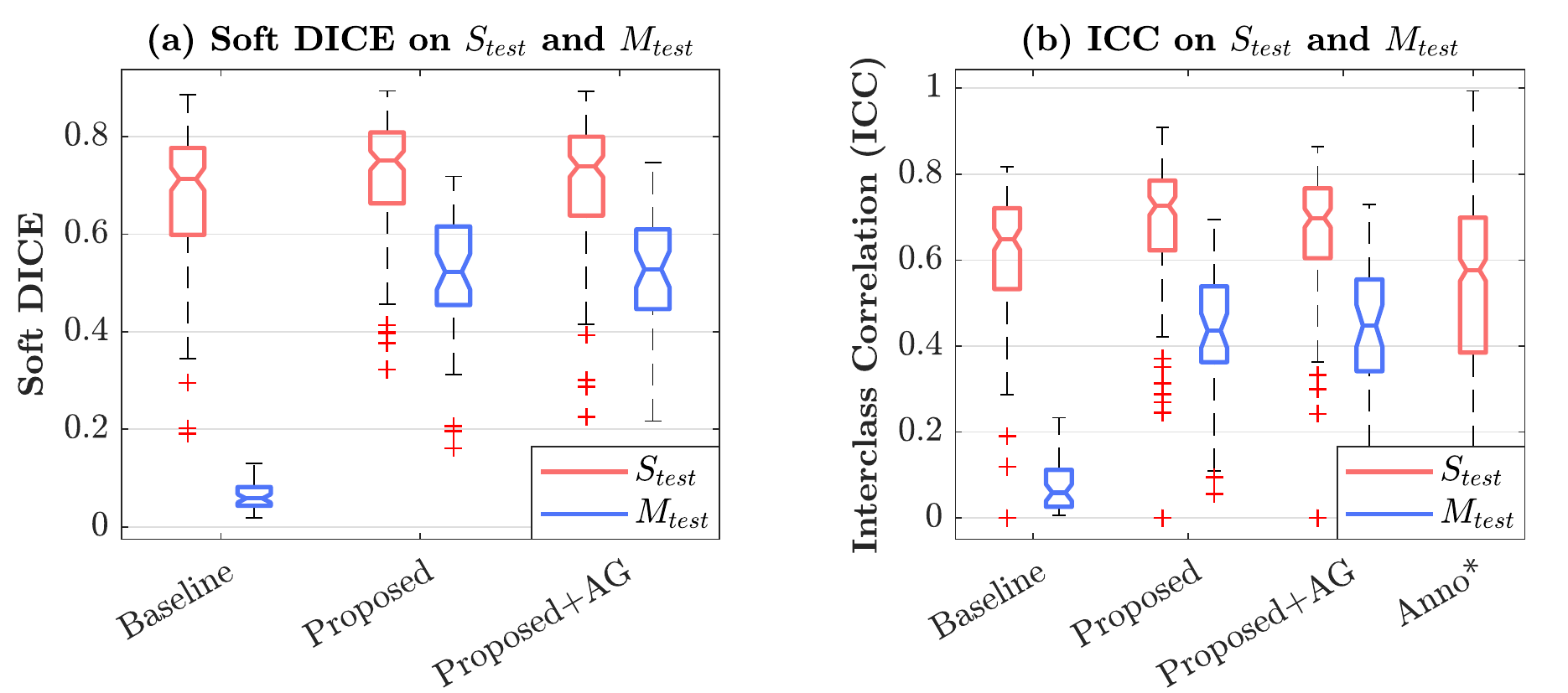}
 \caption{Results of shadow confidence estimation. (a) Soft DICE of the baseline, the proposed method and the proposed method with attention gates (proposed$+$AG) on $S_{test}$ and $M_{test}$. (b) Interclass correlation (ICC) of the baseline, the proposed method and the proposed$+$AG on $S_{test}$ and $M_{test}$. Additionally, ICC of the human performance is shown as $\text{Anno}^\ast$ for $S_{test}$.}
 \label{softDICE_ICC}
\end{figure}

% \begin{figure}[t]
%  \centering
%  \includegraphics[width=0.5\textwidth, trim=1.5cm 0cm 1.5cm 0cm, clip]{SoftDICE.eps}
%  \caption{Soft DICE of the baseline, the proposed method and the proposed method with attention gates (proposed$+$AG) on $S_{test}$ and $M_{test}$.}
%  \label{softDICE}
% \end{figure}

% \begin{figure}[b]
%  \centering
%  \includegraphics[width=0.5\textwidth, trim=1.7cm 0.3cm 1.5cm 0cm, clip]{ICC.eps}
%  \caption{Interclass correlation (ICC) of the baseline, the proposed method and the proposed method with attention gates (proposed$+$AG) on $S_{test}$ and $M_{test}$. Additionally, ICC of the human performance is shown as $\text{Anno}^\ast$ for $S_{test}$.}
%  \label{ICC}
% \end{figure}

% For the evaluation of the transfer network, the average soft DICE scores between $Y^C$ and $Y^T$ on $S_{test}$ are $0.9507$, $0.9451$, $0.9357$ for the baseline, the proposed and the proposed$+$AG methods respectively, while the corresponding scores on $M_{test}$ are $0.8893$, $0.8943$, $0.8580$.
%We also show the learning ability of the confidence estimation network through soft DICE measurements in Fig.~\ref{softDICE}. 
% The soft DICE is computed by comparing the predicted shadow confidence estimation $\hat{y}^C$ and the reference confidence map ${y}^C$. 
Fig.~\ref{softDICE_ICC} (a) shows the soft DICE evaluation on $S_{test}$ and $M_{test}$. The proposed method and the proposed$+$AG method achieve higher soft DICE on both test sets than the baseline, and are more robust than the baseline on $M_{test}$. 
The baseline fails in this experiment on $M_{test}$ because it is unable to obtain accurate shadow segmentation in the previous step (shown in Table~\ref{C48_data_comp}). With less accurate shadow segmentation, the shadow confidence estimation can hardly establish a valid mapping between input images and reference confidence maps. This demonstrates that the shadow-seg module is beneficial for shadow segmentation and confidence estimation.

\begin{figure*}[t]
  \centering
%   \multicolumn{8}{c}{
  \begin{tikzpicture}
  \begin{axis}[
     hide axis,
     scale only axis,
    height=0cm,
    width=17.5cm,
    colormap/viridis,
      colorbar horizontal,
    point meta min=0,
    point meta max=1,
    colorbar style={
        height=5,                 % Höhe der Colorbar
      xtick={0,0.2,0.4,0.6,0.8,1},
      tick label style={font=\tiny},
      xticklabel pos=upper
    }]
    % \addplot [] {};
  \end{axis}
  \end{tikzpicture}
 \\
  \vspace{-5pt}
  \subfloat{\includegraphics[height=1.7cm]{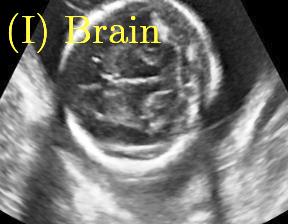}} \hfill 
  \subfloat{\includegraphics[height=1.7cm,  trim=2.29cm 1.29cm 1.88cm 1.48cm, clip]{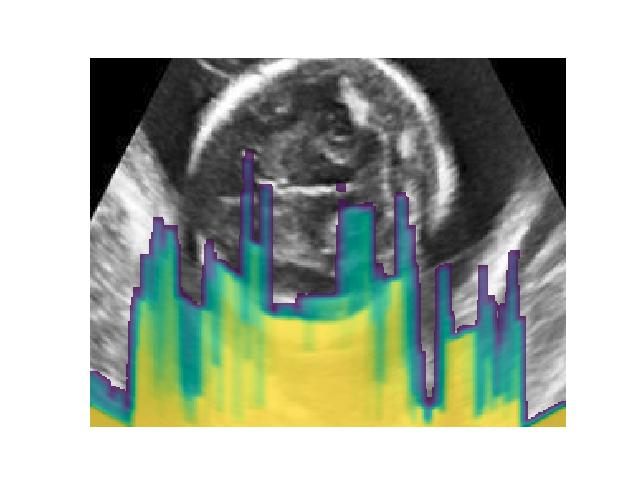}} \hfill 
  \subfloat{\includegraphics[height=1.7cm,  trim=2.29cm 1.29cm 1.88cm 1.48cm, clip]{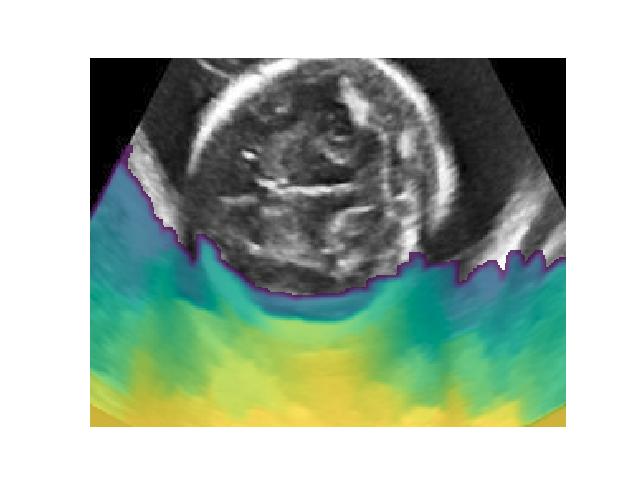}} \hfill 
  \subfloat{\includegraphics[height=1.7cm,  trim=2.29cm 1.29cm 1.88cm 1.48cm, clip]{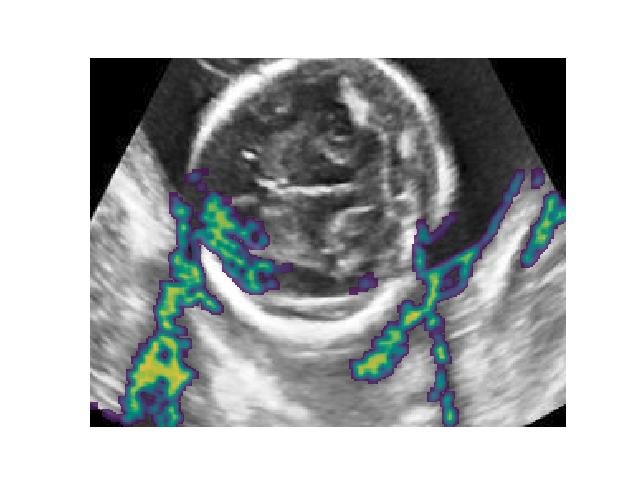}} \hfill 
  \subfloat{\includegraphics[height=1.7cm,  trim=2.29cm 1.29cm 1.88cm 1.48cm, clip]{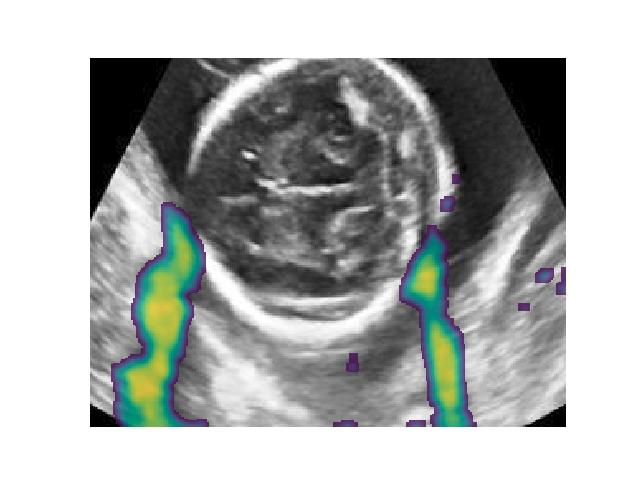}} \hfill 
  \subfloat{\includegraphics[height=1.7cm,  trim=2.29cm 1.29cm 1.88cm 1.48cm, clip]{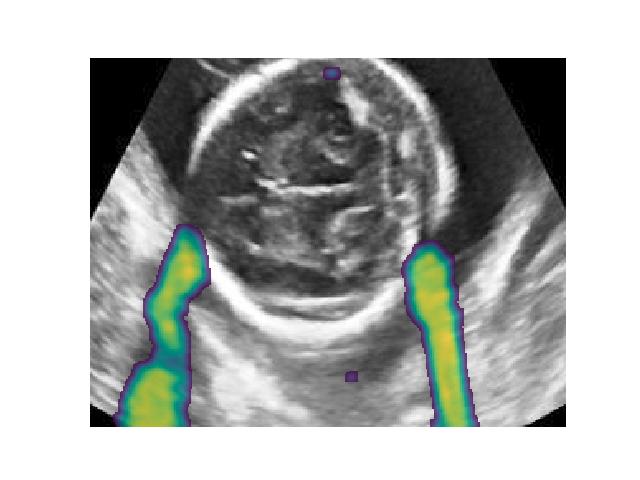}} \hfill 
  \subfloat{\includegraphics[height=1.7cm,  trim=2.29cm 1.29cm 1.88cm 1.48cm, clip]{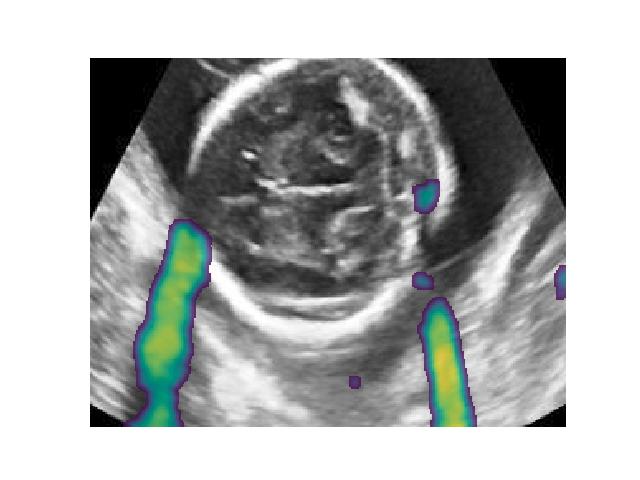}} \hfill 
  \subfloat{\includegraphics[height=1.7cm,  trim=2.29cm 1.29cm 1.88cm 1.48cm, clip]{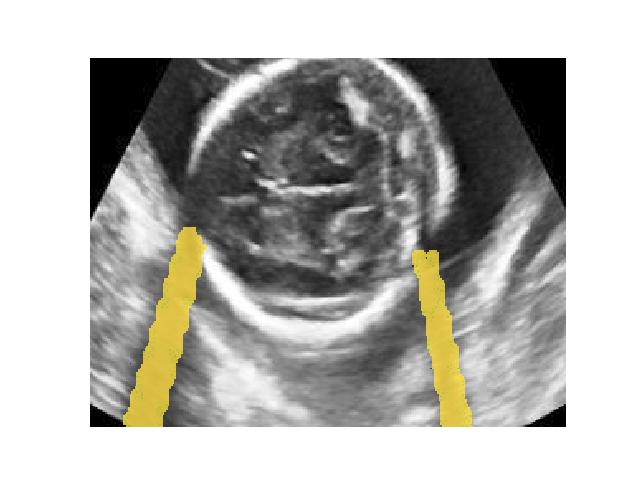}} \hfill  \\

  \subfloat{\includegraphics[height=1.7cm]{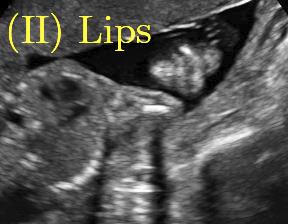}} \hfill 
  \subfloat{\includegraphics[height=1.7cm,  trim=2.29cm 1.29cm 1.88cm 1.48cm, clip]{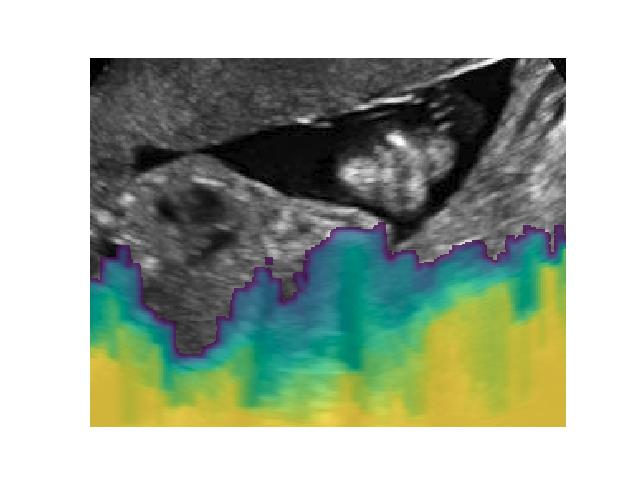}} \hfill 
  \subfloat{\includegraphics[height=1.7cm,  trim=2.29cm 1.29cm 1.88cm 1.48cm, clip]{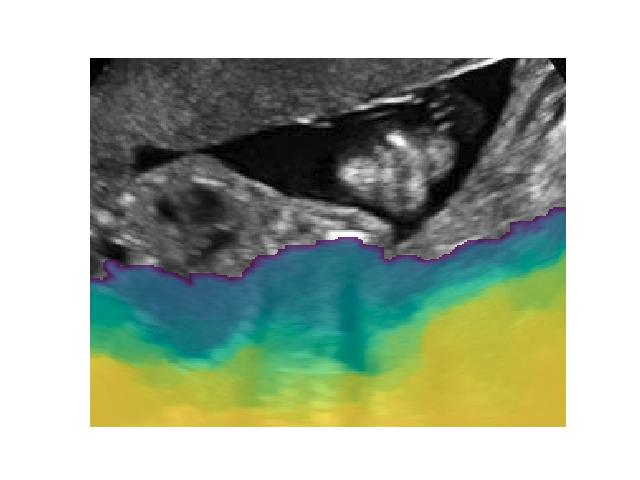}} \hfill 
  \subfloat{\includegraphics[height=1.7cm,  trim=2.29cm 1.29cm 1.88cm 1.48cm, clip]{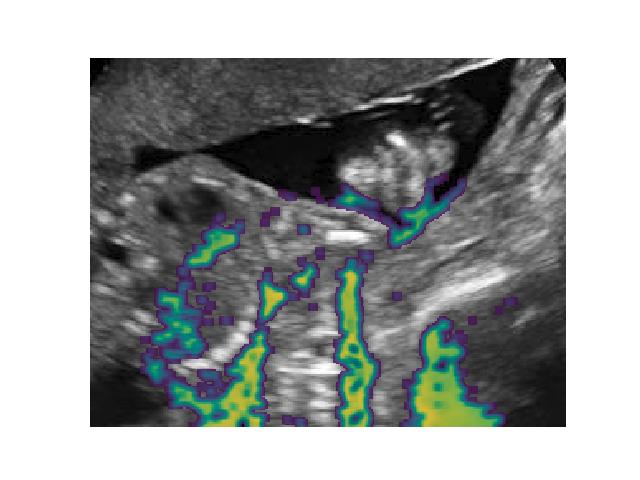}} \hfill 
  \subfloat{\includegraphics[height=1.7cm,  trim=2.29cm 1.29cm 1.88cm 1.48cm, clip]{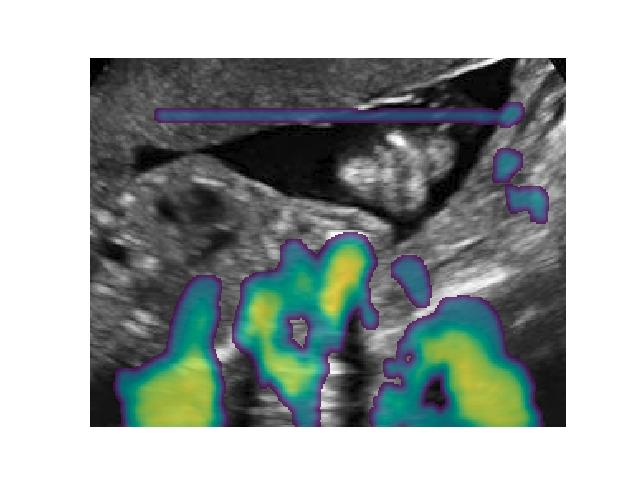}} \hfill 
  \subfloat{\includegraphics[height=1.7cm,  trim=2.29cm 1.29cm 1.88cm 1.48cm, clip]{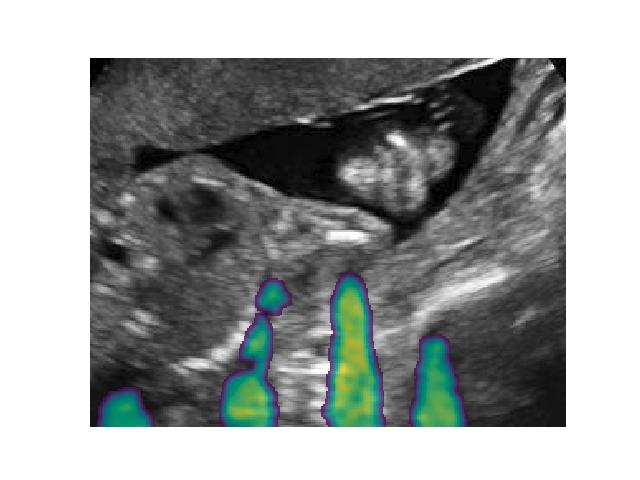}} \hfill 
  \subfloat{\includegraphics[height=1.7cm,  trim=2.29cm 1.29cm 1.88cm 1.48cm, clip]{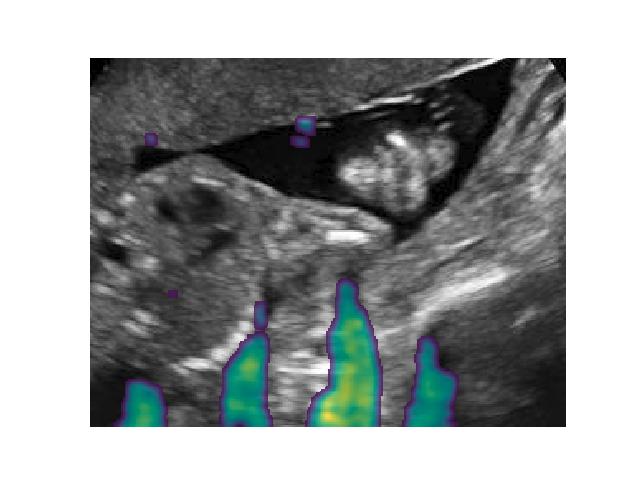}} \hfill 
  \subfloat{\includegraphics[height=1.7cm,  trim=2.29cm 1.29cm 1.88cm 1.48cm, clip]{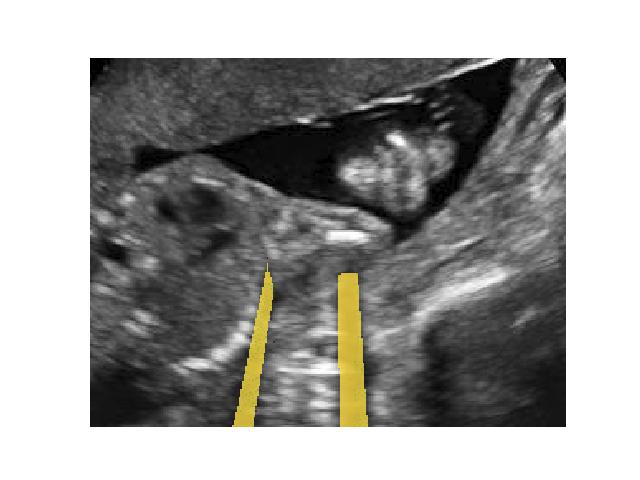}} \hfill  \\
  
  \subfloat{\includegraphics[height=1.7cm]{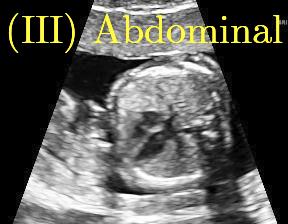}} \hfill 
  \subfloat{\includegraphics[height=1.7cm,  trim=2.29cm 1.29cm 1.88cm 1.48cm, clip]{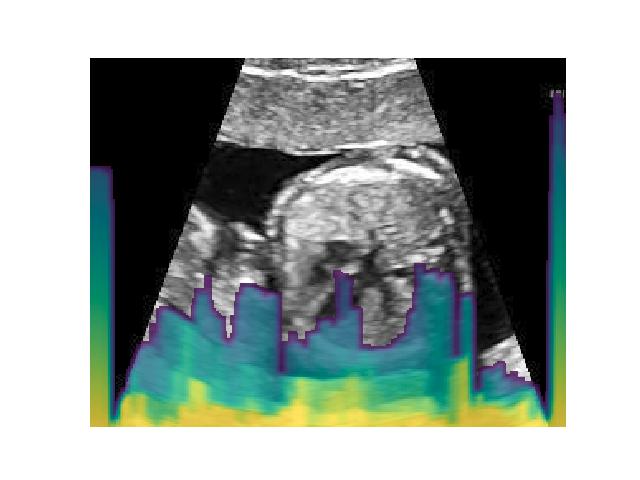}} \hfill 
  \subfloat{\includegraphics[height=1.7cm,  trim=2.29cm 1.29cm 1.88cm 1.48cm, clip]{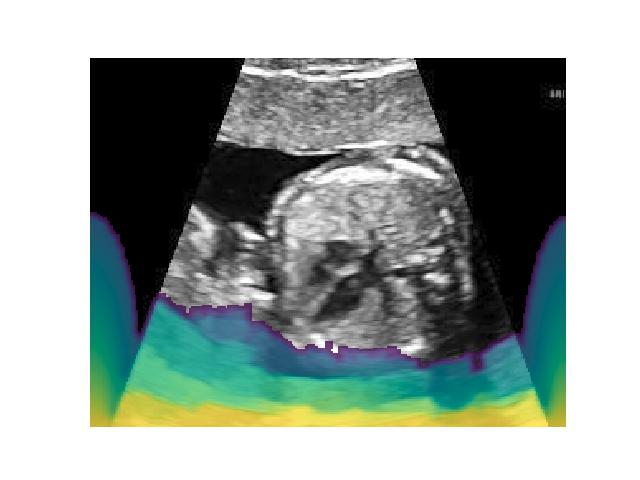}} \hfill 
  \subfloat{\includegraphics[height=1.7cm,  trim=2.29cm 1.29cm 1.88cm 1.48cm, clip]{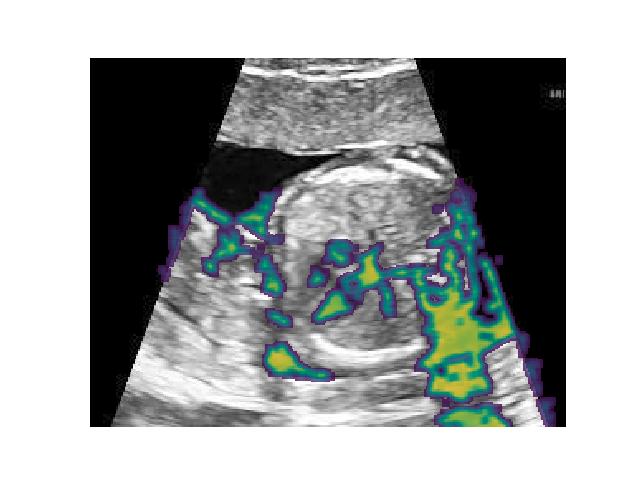}} \hfill 
  \subfloat{\includegraphics[height=1.7cm,  trim=2.29cm 1.29cm 1.88cm 1.48cm, clip]{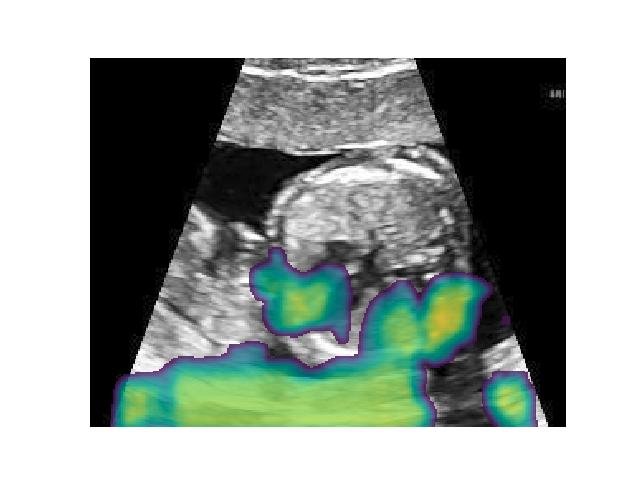}} \hfill 
  \subfloat{\includegraphics[height=1.7cm,  trim=2.29cm 1.29cm 1.88cm 1.48cm, clip]{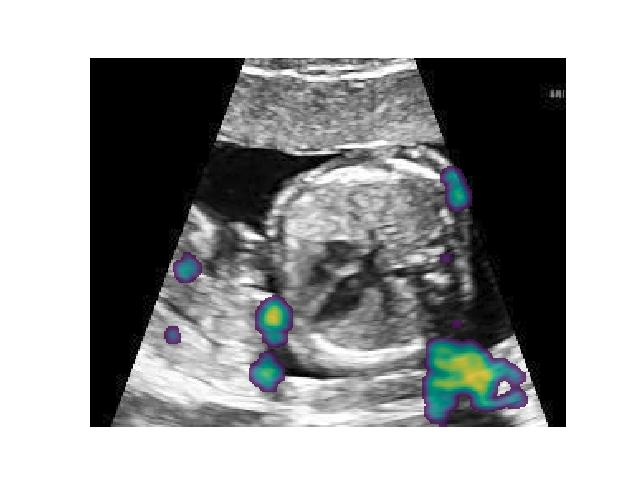}} \hfill 
  \subfloat{\includegraphics[height=1.7cm,  trim=2.29cm 1.29cm 1.88cm 1.48cm, clip]{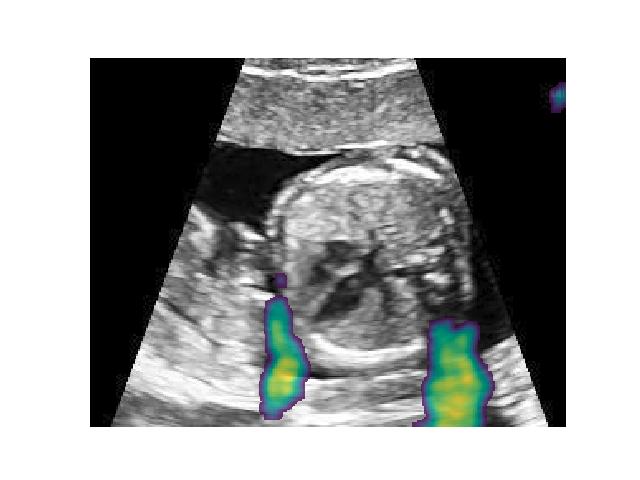}} \hfill 
  \subfloat{\includegraphics[height=1.7cm,  trim=2.29cm 1.29cm 1.88cm 1.48cm, clip]{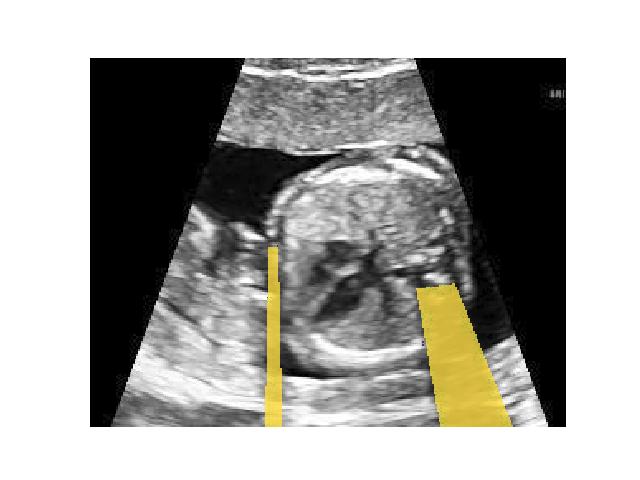}} \hfill  \\
  
 \setcounter{subfigure}{0}
   \subfloat[Image]{\includegraphics[height=1.7cm]{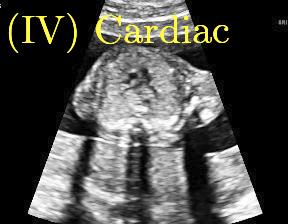}} \hfill 
  \subfloat[RW~\cite{Karamalis2012}]{\includegraphics[height=1.7cm,  trim=2.29cm 1.29cm 1.88cm 1.48cm, clip]{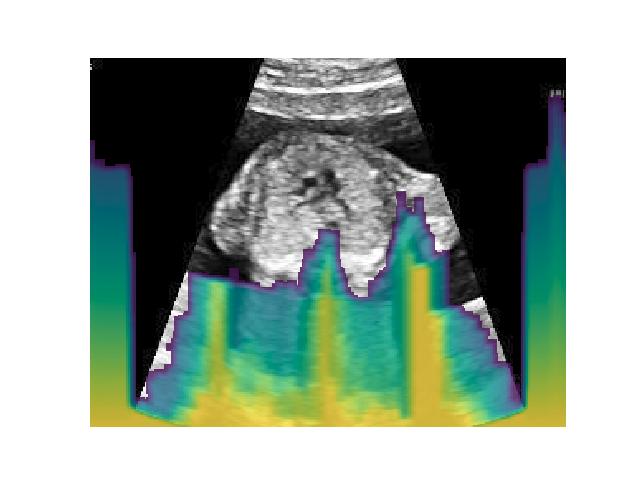}} \hfill
  \subfloat[$\text{RW}^\ast$~\cite{Karamalis2012}]{\includegraphics[height=1.7cm,  trim=2.29cm 1.29cm 1.88cm 1.48cm, clip]{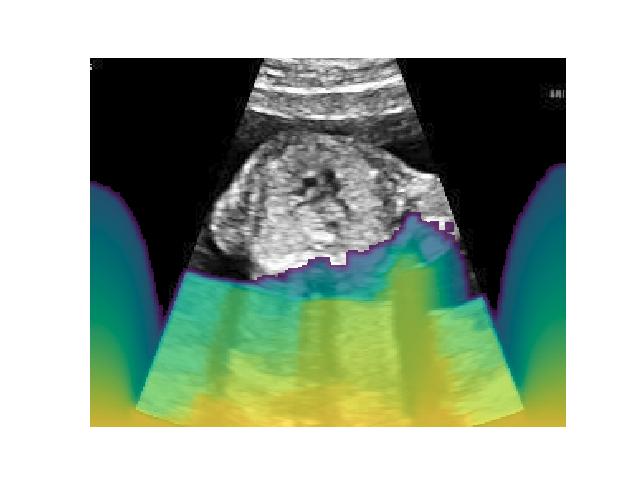}} \hfill 
  \subfloat[Pilot~\cite{meng2018}]{\includegraphics[height=1.7cm,  trim=2.29cm 1.29cm 1.88cm 1.48cm, clip]{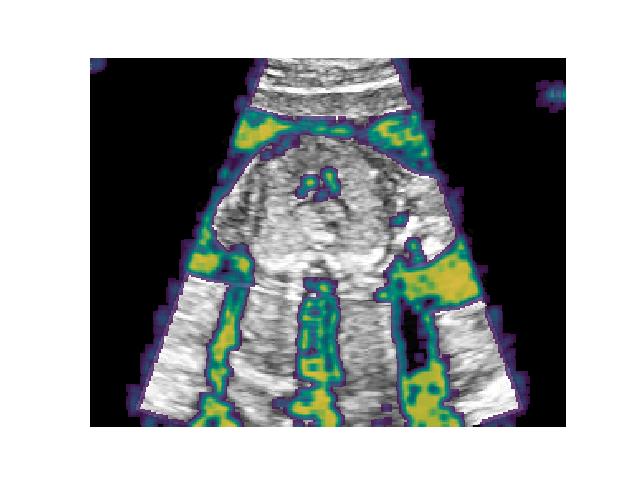}} \hfill 
  \subfloat[Baseline]{\includegraphics[height=1.7cm,  trim=2.29cm 1.29cm 1.88cm 1.48cm, clip]{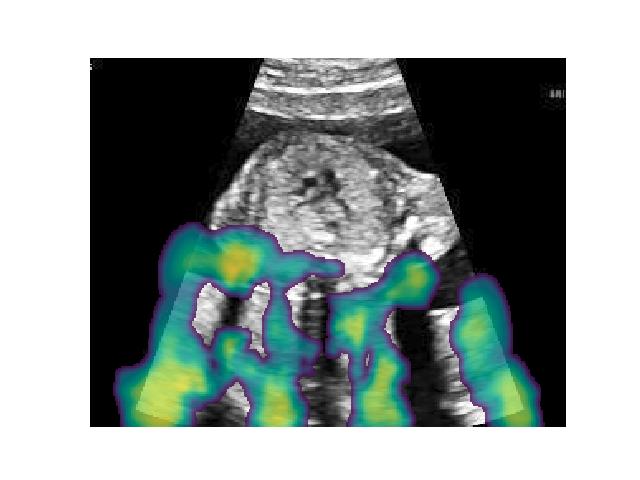}} \hfill 
  \subfloat[Proposed]{\includegraphics[height=1.7cm,  trim=2.29cm 1.29cm 1.88cm 1.48cm, clip]{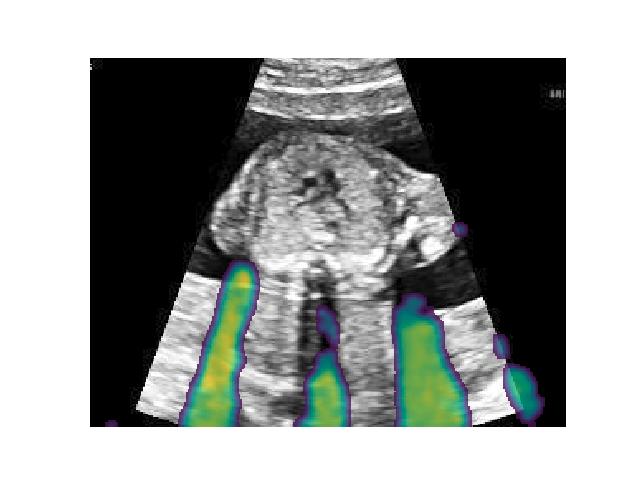}} \hfill 
  \subfloat[Proposed+AG]{\includegraphics[height=1.7cm,  trim=2.29cm 1.29cm 1.88cm 1.48cm, clip]{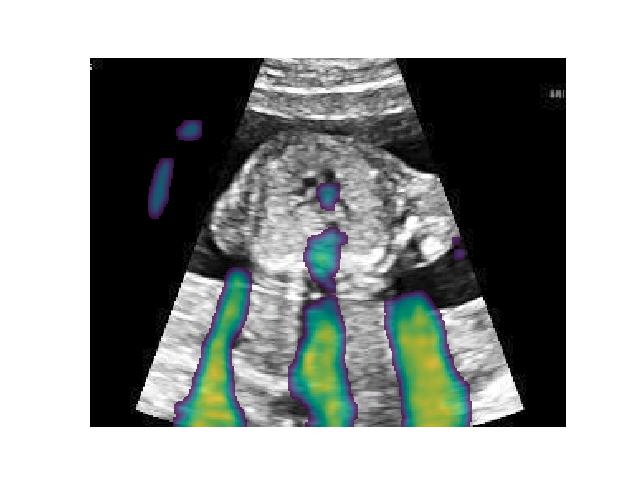}} \hfill
  \subfloat[Weak GT]{\includegraphics[height=1.7cm,  trim=2.29cm 1.29cm 1.88cm 1.48cm, clip]{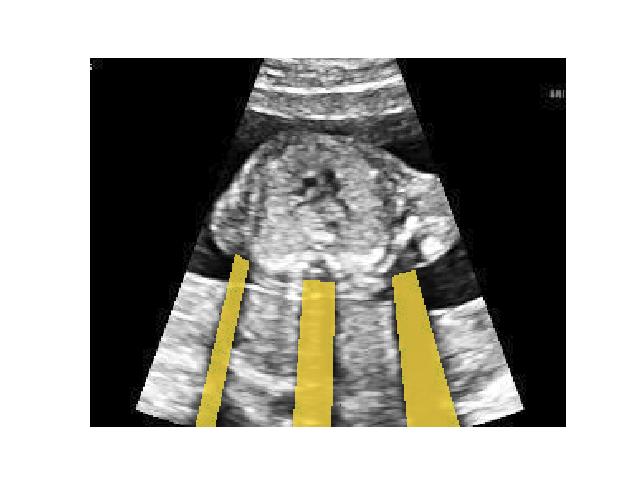}} \hfill  \\
  
  \caption{Confidence estimation of shadow regions using the state-of-the-art methods and our methods. Rows I-IV show four examples: Brain (top), Lip (second), Abdominal (third) and Cardiac (bottom). Column (a) is the original US image. Columns (b-d) are shadow confidence maps from the RW algorithm~\cite{Karamalis2012} and our previous work~\cite{meng2018}. Columns (e-g) show the shadow confidence maps of the baseline, the proposed method and the Proposed$+$AG method. Column (h) is the binary map of the manual shadow segmentation. The color bar on the top of this figure shows that the more yellow/brighter (closer to 1), the higher the confidence of being shadow regions.}
  \label{compareConf}
\end{figure*}

We additionally evaluate the reliability of the shadow confidence estimation by measuring the agreement between the decision of each method and the manual segmentation. Regarding the baseline, the proposed and the proposed$+$AG as different judges and the manual segmentation of shadow regions as a contrasting judge, we use the ICC to measure the agreement between each different judge and the contrasting judge. Fig.~\ref{softDICE_ICC} (b) shows the ICC evaluation on two test data sets, which indicate that the proposed method and the proposed$+$AG are more consistent on estimating shadow confidence maps compared with the baseline. When considering another manual segmentation of shadow regions as an extra judge, we can evaluate the agreement of human annotations. Fig.~\ref{softDICE_ICC} (b) shows that the ICC of two human annotations (shown as $\text{Anno}$) is normally $0.51$. The proposed method with an ICC of $0.66$ is more consistent than annotations from two human annotators.

Fig.~\ref{compareConf} compares the shadow confidence maps of the state-of-the-art methods and the proposed methods. RW and $\text{RW}^\ast$ have the same parameters as used for Table~\ref{B_data_comp}. The shadow confidence maps of the baseline, the proposed method and the proposed$+$AG method are generated directly from input shadow images by confidence estimation networks. Overall, the proposed method and the proposed$+$AG method achieve more visually reasonable shadow confidence estimation than the baseline and the state-of-the-art on different anatomical structures shown in Fig.~\ref{compareConf}. The proposed method and the proposed$+$AG method are able to highlight multiple shadow regions while the RW algorithm shows limitations for most cases, especially for disjoint shadow regions.
 
Row I in Fig.~\ref{compareConf} shows a fetal brain image from $S_{test}$. The confidence estimation of shadow regions from the baseline, the proposed method and the proposed$+$AG method are similarly accurate since we use fetal brain images to train the confidence estimation networks in these three methods. These outperform \cite{Karamalis2012} and \cite{meng2018}. 
Rows (II-IV) in Fig.~\ref{compareConf} show shadow confidence maps of non-brain anatomy from $M_{test}$, including lips, abdominal and cardiac. The baseline failed on unseen data during inference. However, the proposed methods are able to generate accurate shadow confidence maps because of the generalized shadow features obtained by the shadow-seg module. Furthermore, the ``Lips" example shows that our method is capable of detecting weaker shadow regions that have not been annotated in manual segmentation. This indicates that the confidence estimation network has learned general properties of shadow regions.
% , thus the network appears to have learned general properties of shadow regions. 

\subsection{Transfer Function Performance}
\begin{figure}[ht]
 \centering
\vspace{-0.25cm}
  \subfloat{\includegraphics[height=1.7cm]{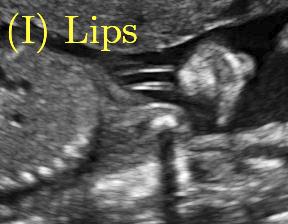}} \hfill 
  \subfloat{\includegraphics[height=1.7cm, trim=2.29cm 1.29cm 1.88cm 1.48cm, clip]{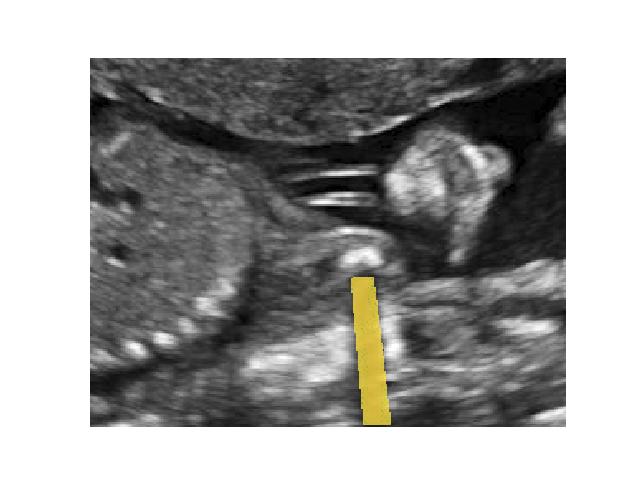}} \hfill 
  \subfloat{\includegraphics[height=1.7cm, trim=2.29cm 1.29cm 1.88cm 1.48cm, clip]{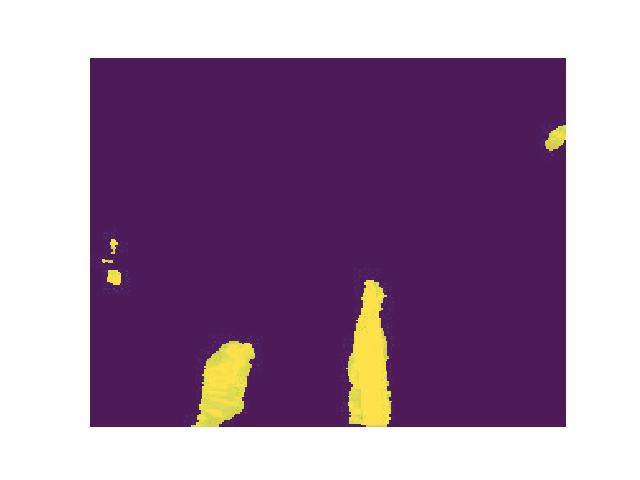}} \hfill 
  \subfloat{\includegraphics[height=1.7cm, trim=2.29cm 1.29cm 1.88cm 1.48cm, clip]{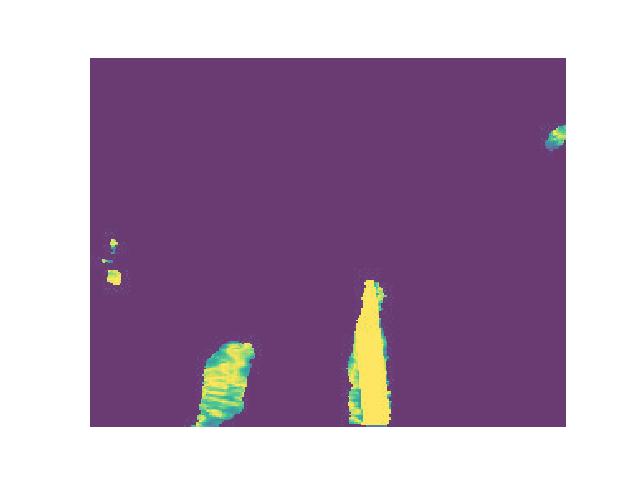}} \hfill 
  \\
 \setcounter{subfigure}{0}
   \subfloat[Image]{\includegraphics[height=1.7cm]{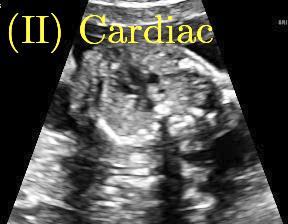}} \hfill 
  \subfloat[Weak GT]{\includegraphics[height=1.7cm, trim=2.29cm 1.29cm 1.88cm 1.48cm, clip]{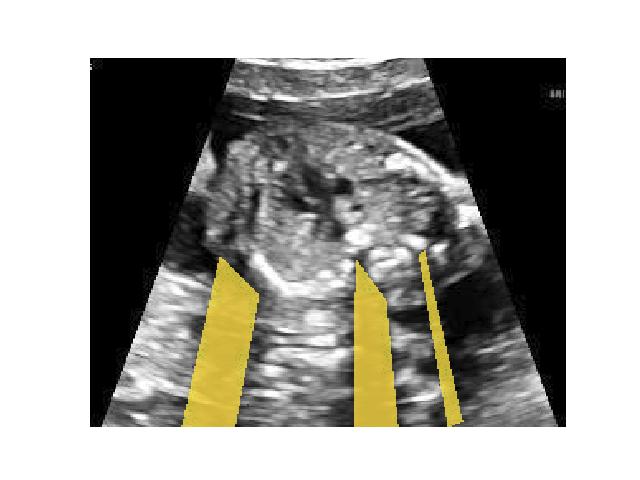}} \hfill
  \subfloat[$\hat{Y}^S$]{\includegraphics[height=1.7cm, trim=2.29cm 1.29cm 1.88cm 1.48cm, clip]{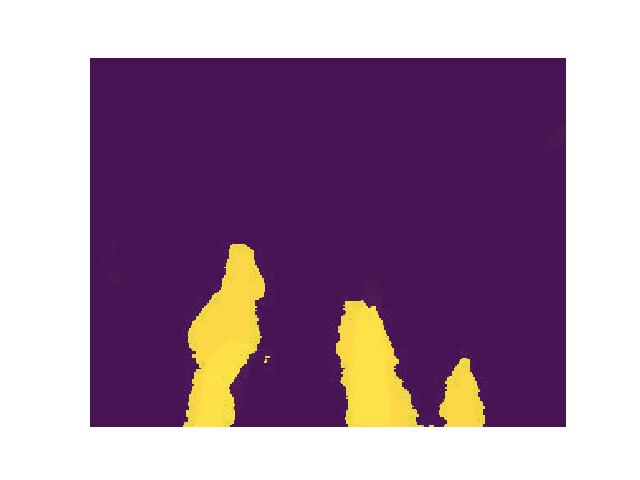}} \hfill 
  \subfloat[$Y^C$]{\includegraphics[height=1.7cm, trim=2.29cm 1.29cm 1.88cm 1.48cm, clip]{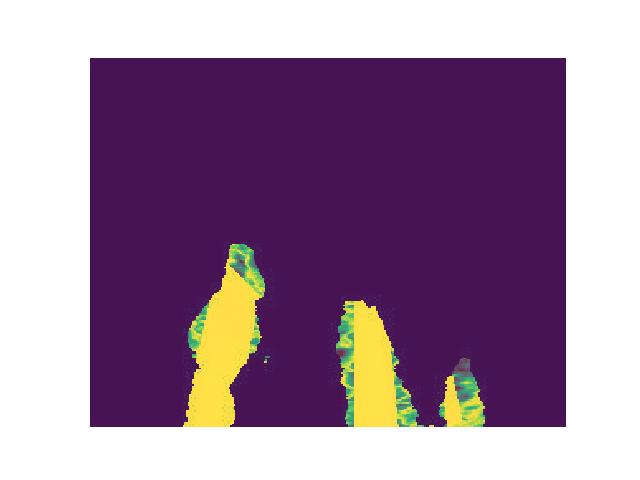}} \hfill 
  \\
  \caption{Two examples showing the performance of the transfer function. (a) is the input image and (b) is the binary manual segmentation. (c) is the predicted segmentation before applying the transfer function while (d) is the corresponding reference confidence map after the transfer function.}
  \label{TFunc}
\end{figure}

%\subsubsection*{\textbf{Transfer Function Performance}}
We show two illustrative examples in Fig.~\ref{TFunc} to demonstrate the performance of the transfer function. Fig.~\ref{TFunc} (c) and (d) show that the transfer function computes the confidence of each pixel in the false positive areas of the predicted segmentation, so that to extend a binary segmentation to a reference confidence map.

\subsection{Runtime}
%\subsubsection*{\textbf{Runtime}}
%We show the average confidence estimation inference times of the state-of-the-art methods and our proposed methods. 
The RW algorithm \cite{Karamalis2012} is implemented in Matlab (CPU Xeon E5-2643) while the previous work \cite{meng2018} and the proposed methods use Tensorflow and run on a Nvidia Titan X GPU.
For the RW algorithm \cite{Karamalis2012} and the previous work \cite{meng2018}, the inference time are $0.4758s$ and $11.35s$ respectively. Since the baseline, the proposed method and the proposed$+$AG method have the same confidence estimation networks, they have the same inference time, which is $0.0353s$. 
%These inference times show that, under the aforementioned implementation setting for each method, 
%Our implementation is about $10 \times$ faster than \cite{Karamalis2012} and much faster than \cite{meng2018} on systems as described above, which indicates that our proposed method is able to offer low-overhead, effective real-time feedback in clinical practice. 
A system-independent evaluation can be performed by estimating the required Giga-floating point operations (GFlops,  fused multiply-adds) during inference. Our method requires $\sim 2.5 - 3$ GFlops (estimated from conv-layers including ReLU activation, Appendix~\ref{flopsEq}, supplementary materials are available in the supplementary files /multimedia tab.) while the RW algorithm~\cite{Karamalis2012} requires $\sim 4 - 4.5$ GFlops (according to the built-in Matlab profiler) and the previous work~\cite{meng2018} requires $\sim19$ GFlops (estimated from conv-layers including ReLU activation, Appendix~\ref{flopsEq}).
%The evaluation results of our method  indicate that our proposed method can offer low-overhead, effective real-time feedback in clinical practice.

% The average confidence estimation inference times of the state-of-the-art methods (\cite{Karamalis2012, meng2018}) and our proposed methods are shown in Table.~\ref{runtime}. Since the baseline, the proposed method and the proposed$+$AG have the same confidence estimation network, these three methods have the same inference time. Table.~\ref{runtime} shows that our proposed methods is about $10$ times faster than the RW algorithm and much faster than the previous worl, which indicates that our proposed method is capable of offering real-time feedback in clinical practice.

% \begin{table}[h]
% \centering
% \caption{Comparison of Average Inference Time.}
% \label{runtime}
% \begin{tabular}{*4c}
% \toprule
% ~~~~~~~~~~~~~~~                                 & 
% RW~\cite{Karamalis2012}                                   & 
% PreWork~\cite{meng2018}                                 & 
% the Proposed methods                                       \\
% \midrule
% Runtime                                               &     
% 0.4758~s                                               &
% 11.35~s                                                 &
% 0.0353~s                                               \\ 
% \bottomrule
% \end{tabular}
% \end{table}

\section{Applications}
To verify the practical benefits of our method, we integrate the shadow confidence maps into different applications such as 2D US standard plane classification, multi-view image fusion and automated biometric measurements.

\subsection{Ultrasound Standard Plane Classification}
%\subsubsection*{\textbf{Ultrasound Standard Plane Classification}}
Classifying 2D fetal standard planes is of great importance for early detection of abnormalities during mid-pregnancy~\cite{salomon2011}. However, distinguishing different standard planes is a challenging task and requires intense operator training and experience. %Selecting the correct scan planes also relies heavily on the experience of sonographers during scanning. %Yaqub et al.~\cite{yaqub2015} have proposed a guided random forest to categorise seven standard scan planes from 2D ultrasound images at 18 to 22 weeks of gestation. This method is a robust classification method that utilise invariant features to categorise images containing one or more fetal and non-fetal structures. Instead of categorising standard scan planes, 
Baumgartner et al.~\cite{christian2017} have proposed a deep learning method for the detection of various fetal standard planes. 
%With the real time detection, the authors further localise the region of interest (ROI) in the standard scan plane.
We extend~\cite{christian2017} and utilize shadow confidence maps to provide extra information for standard plane classification.
% and further, to improve the classification accuracy of the standard planes. 
% We use the shadow confidence maps as an extra channel in the standard scan planes classification network.

The data is the same as used in \cite{christian2017}, which is a set of $2694$ 2D ultrasound examinations between 18-22 weeks of gestation (iFIND Project~\ref{ifindfoot}). We select nine classes of standard planes including Three Vessel View (3VV), Four Chamber View (4CH), Abdominal, Brain View at the level of the cerebellum (Brain (cb.)), Brain view at posterior horn of the ventricle (Brain (tv.)), Femur, Lips, Left Ventricular Outflow Tract (LVOT) and Right Ventricular Outflow Tract (RVOT). 
The data set is split into training ($16089$), validation ($450$), and testing ($4368$) images,   similar to~\cite{christian2017} (see \cref{appddata} for individual class split numbers). We use image whitening (subtracting the mean intensity and
divide by the variance) on each image to preprocess the whole data set.

Four networks based on  SonoNet-32~\cite{christian2017} are trained and tested in order to verify the utility of shadow confidence maps. The first network is trained with the standard plane images from the training data. The next three networks are separately trained with standard plane images and their corresponding shadow confidence maps obtained by the baseline, the proposed method and the proposed$+$AG method. Thus, the training data in the first network has one channel while the remaining networks have two input channels.
%the data in the three comparison networks have two channels. 
% the standard plane images and the shadow confidence maps obtained by the baseline, the proposed method and the proposed$+$AG. 
We train these networks for $75$ epochs with a learning rate of $0.001$. 
% on a Nvidia Titan X GPU with 12 GB of memory.

Table~\ref{classification} shows the standard plane classification performance of the four networks. Networks with shadow confidence maps achieve higher classification accuracy on almost all classes (except Abdominal, LVOT and RVOT), as well as on average classification accuracy.
$CM_{PAG}$ achieves highest classification accuracies for five classes (3VV, 4CH, Brain(Cb.), Brain(Tv.) and Femur). Of particular note, the accuracies of the 3VV and 4CH classes increase over the baseline by $11.75\%$ and $5.5\%$ respectively. Five other classes (Abdominal, Brain(Cb.), Brain(Tv.), Femur and Lips) achieve near $100\%$ accuracy in both the baseline and $CM_{PAG}$, while LVOT and RVOT classes see modest decreases in $CM_{PAG}$ compared with the baseline, $2.1\%$ and $1.0\%$ respectively. Therefore, when compared $CM_{PAG}$ with the baseline, the increase in average classification accuracy across all classes ($97.37\%$ to $98.74\%$) is primarily driven by the large improvements in 3VV and 4CH. 
These results indicate that shadow confidence maps are able to provide extra information and improve the performance of another automatic medical image analysis algorithm.

\begin{table}[t]
\centering
\caption{Classification accuracy ($\%$) with vs. without shadow confidence maps. w/o $CM$ is the network
without shadow confidence maps while ${CM}_B$, ${CM}_P$ ,
${CM}_{PAG}$ are networks with shadow confidence maps from the baseline, the proposed method and the proposed+AG method. Best results are shown in bold.}
\label{classification}
\begin{tabular}{*5c}
\toprule
Class                                               & 
w/o $CM$                                              &
${CM}_B$                                            & 
${CM}_P$                                            & 
${CM}_{PAG}$                                          \\ 
\midrule                                            
3VV                                                 &
80.87                                               &
89.93                                               &     
88.93                                               &
\textbf{92.62}                                               \\
4CH                                                 &
94.50                                               &
100.00                                              &
98.38                                              &     
\textbf{100.00}                                              \\ 
Abdominal                                           &
\textbf{100.00}                                              & 
99.82                                               & 
99.28                                              & 
99.82                                               \\
Brain(Cb.)                                          &
100.00                                              &
99.84                                              & 
100.00                                              & 
\textbf{100.00}                                             \\
Brain(Tv.)                                          &
99.11                                               & 
99.78                                               & 
99.78                                              & 
\textbf{99.89}                                               \\
Femur                                               &
99.04                                               & 
99.81                                               & 
99.81                                               & 
\textbf{99.81}                                               \\
Lips                                                &
98.29                                               & 
99.81                                               & 
\textbf{100.00}                                              & 
99.81                                              \\
LVOT                                                &
\textbf{97.90}                                               &
93.69                                              & 
94.29                                               & 
95.80                                               \\
RVOT                                                &
\textbf{95.95}                                               &
93.24                                               & 
92.57                                               & 
94.93                                               \\
\midrule 
\textbf{Avg.}                                    &
97.37                                               &
98.24                                               & 
98.03                                               & 
\textbf{98.74}                                               \\ 
\bottomrule
\end{tabular}
\end{table}

\begin{figure}[t]
 \centering
 \includegraphics[width=0.5\textwidth, trim=0cm 11cm 0cm 10.5cm, clip]{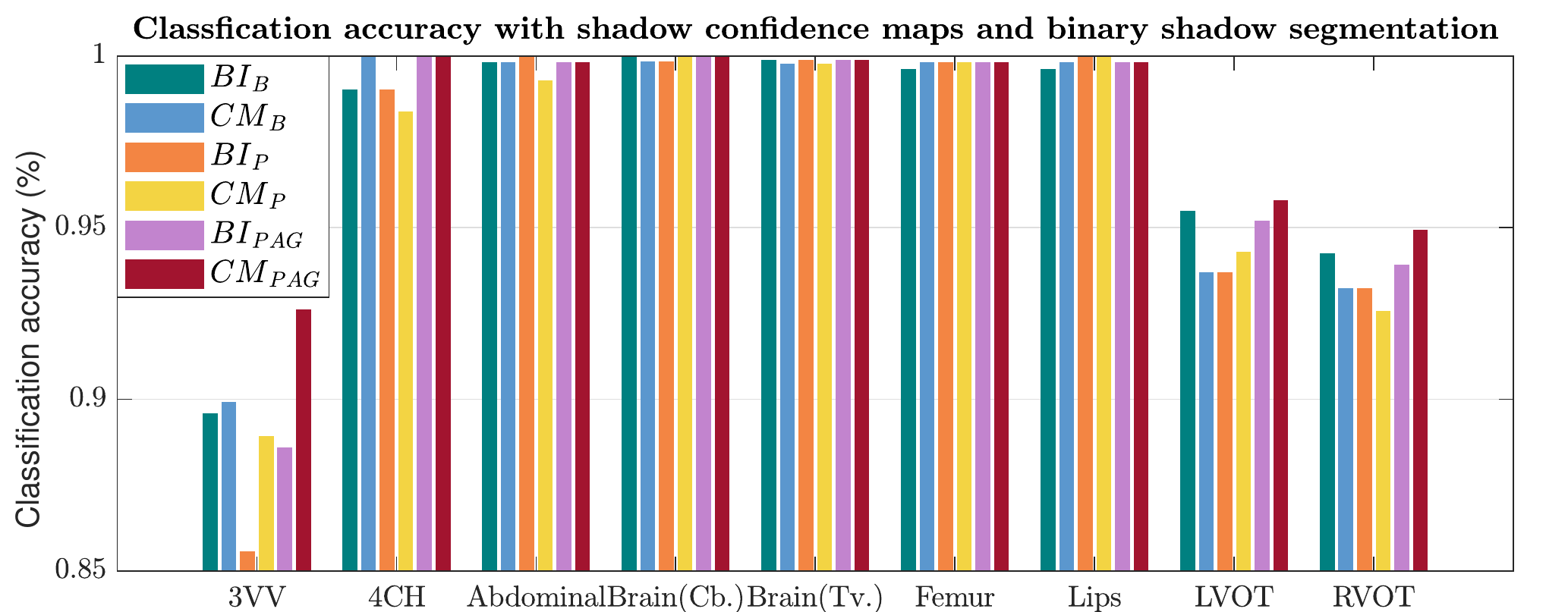}
 \caption{Comparison of classification accuracy between using shadow confidence maps and using shadow segmentation. ${BI}_B$, ${BI}_P$ and ${BI}_{PAG}$ are networks with binary shadow segmentation from the baseline, the proposed method and the proposed$+$AG method. ${CM}_B$, ${CM}_P$ and ${CM}_{PAG}$ are the same networks as in Tabel~\ref{classification}.}
 \label{BiVSConf}
\end{figure}

We additionally explore the importance of estimating confidence maps over binary segmentation of shadow regions. We compare the classification accuracy between using shadow confidence maps and directly using binary shadow segmentations generated from different methods. Fig.\ref{BiVSConf} shows that for classes with high classification accuracy such as Abdominal, Brain(Cb.), Brain(Tv.), Femur and Lips, integrating shadow confidence maps into the classification task yields minor improvement. However, for classes with relatively low classification accuracy such as 3VV and LVOT, classification with shadow confidence maps achieves higher accuracy than classification with only binary shadow segmentations. 

\subsection{Multi-view Image Fusion}
%\subsubsection*{\textbf{Multi-view Image Fusion}}
Routine US screening is usually performed using a single 2D probe. 
%US imaging normally utilize a single probe to visualize anatomical structures for abnormalities detection during routine diagnose. 
However, the position of the probe and resulting tomographic view through the anatomy has great impact on diagnosis. Zimmer et al.~\cite{veronika2018} proposed a multi-view image reconstruction method, which compounds different images of the same anatomical structure acquired from different view directions. They use a Gaussian weighting strategy to blend intensity information from different views.
Here, we combine predicted shadow confidence maps from these multi-view images as additional image fusion weights to investigate if these confidence maps can further improve image quality. 

The proposed method generally outperforms the baseline and the proposed$+$AG method, thus we only integrate the shadow confidence maps generated by the proposed method (${CM}_P$) into the weighting strategy in~\cite{veronika2018}. 
% Specifically, the proposed method is utilized to generate shadow confidence maps for multi-view images.
In detail, the probability value of each pixel in a shadow confidence map is multiplied to the original weight of the same pixel computed in~\cite{veronika2018}.
% , which are built based on the position in the frustum and the signal intensities. 
The generated new weights are normalized as described in~\cite{veronika2018} and then are used for image fusion. The data set in this experiment is same as used for~\cite{veronika2018}.
% Based on the probability value of each pixel, each shadow confidence map is encoded by the Gaussian kernel as a weighting factor. In the weighting factor, the higher the shadow probability is at one pixel, the lower the weight of that point. This weighting factor is then multiplied to the original weights in~\cite{veronika2018} and the generated new weights are used for image fusion. The data set in this experiment is same as used for~\cite{veronika2018}.

\begin{figure*}[tb]
\centering
%   \begin{minipage}{0.05\columnwidth}
%   \centering
%     \subfloat{\rotatebox{90}{~ no ${CM}_P$}} \\ 
%     \subfloat{\rotatebox{90}{~~~ + ${CM}_P$}}\\
%   \end{minipage}
  \begin{tikzpicture}
  \begin{axis}[
     hide axis,
     scale only axis,
    height=0cm,
    width=17.5cm,
    colormap/viridis,
      colorbar horizontal,
    point meta min=0,
    point meta max=1,
    colorbar style={
        height=5,                 % Höhe der Colorbar
      xtick={0.13, 1},
      xticklabels={$\text{low difference}$, $\text{high difference}$},
      tick label style={font=\small},
      xticklabel pos=upper,
      xticklabel style = {xshift=-1.2cm},
    }]
    % \addplot [] {};
  \end{axis}
  \end{tikzpicture}
 \\
  \vspace{-5pt}

 \setcounter{subfigure}{0}
 \subfloat[Gaussian, MSE]{
 \begin{tabular}{@{\hspace{0\tabcolsep}}c@{\hspace{0.3\tabcolsep}}c@{\hspace{0.3\tabcolsep}}c@{\hspace{0\tabcolsep}}}
   \includegraphics[height=1.25cm, trim=2.9cm 3.2cm 2cm 3.2cm, clip]{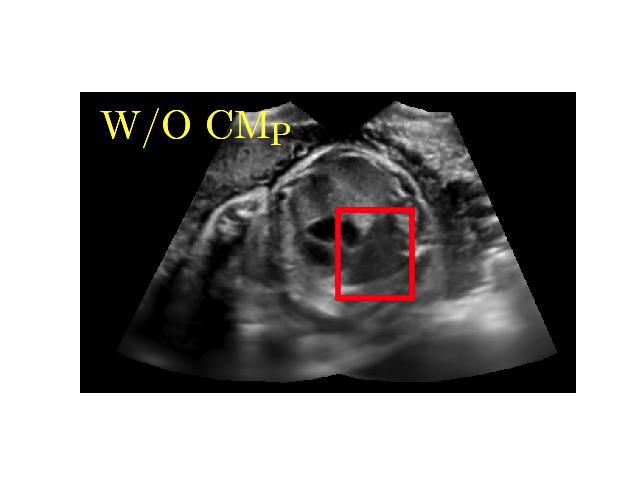} & \includegraphics[height=1.25cm]{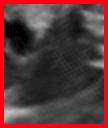} & \multirow{2}{*}[7pt]{\includegraphics[height=1.25cm, trim=4.5cm 1.5cm 4.5cm 1.5cm, clip]{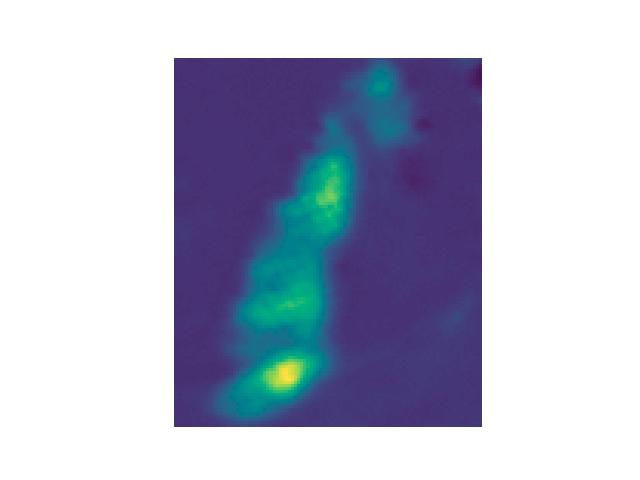}}\\
     \includegraphics[height=1.25cm, trim=2.9cm 3.2cm 2cm 3.2cm, clip]{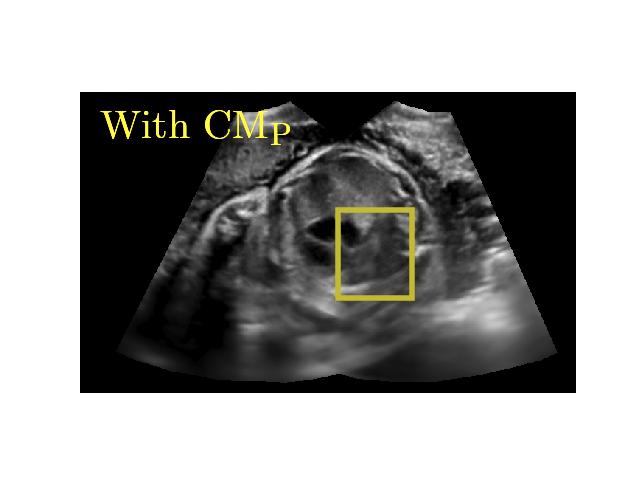}&\includegraphics[height=1.25cm]{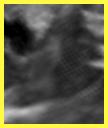} 
 \end{tabular}
 }
\hfill
\hspace{-3.5ex}
 \setcounter{subfigure}{1}
 \subfloat[Int. \& Gaussian, MSE]{
 \begin{tabular}{@{\hspace{0\tabcolsep}}c@{\hspace{0.3\tabcolsep}}c@{\hspace{0.3\tabcolsep}}c@{\hspace{0\tabcolsep}}}
 \includegraphics[height=1.25cm, trim=2.9cm 3.2cm 2cm 3.2cm, clip]{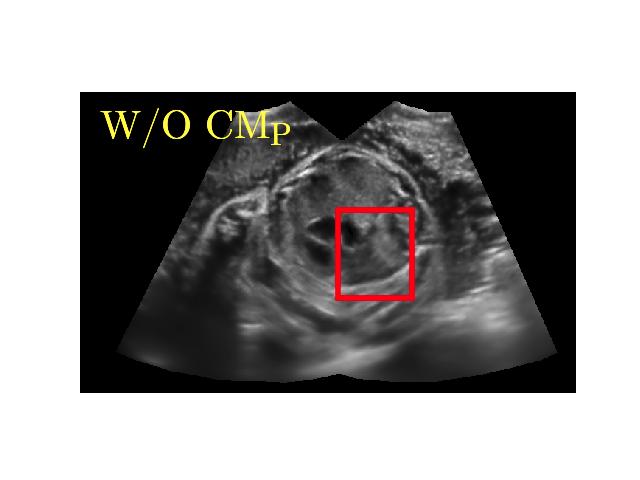} &
 \includegraphics[height=1.25cm]{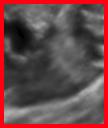} & \multirow{2}{*}[7pt]{\includegraphics[height=1.25cm, trim=4.5cm 1.5cm 4.5cm 1.5cm, clip]{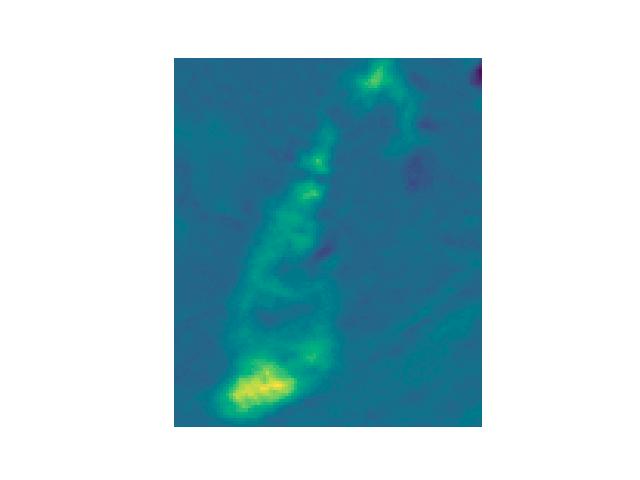}}\\
 \includegraphics[height=1.25cm, trim=2.9cm 3.2cm 2cm 3.2cm, clip]{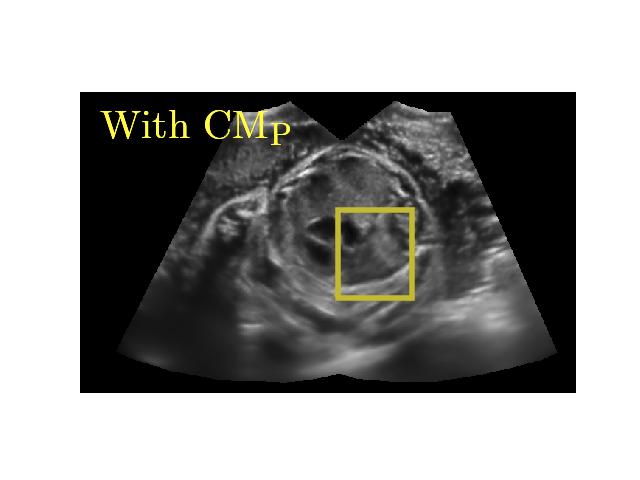} &
  \includegraphics[height=1.25cm]{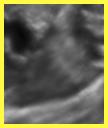}
  \end{tabular}
  }
  \hfill
\hspace{-3.5ex}
\setcounter{subfigure}{2}
\subfloat[Gaussian, Sigmoid]{
 \begin{tabular}{@{\hspace{0\tabcolsep}}c@{\hspace{0.3\tabcolsep}}c@{\hspace{0.3\tabcolsep}}c@{\hspace{0\tabcolsep}}}
 \includegraphics[height=1.25cm, trim=2.9cm 3.2cm 2cm 3.2cm, clip]{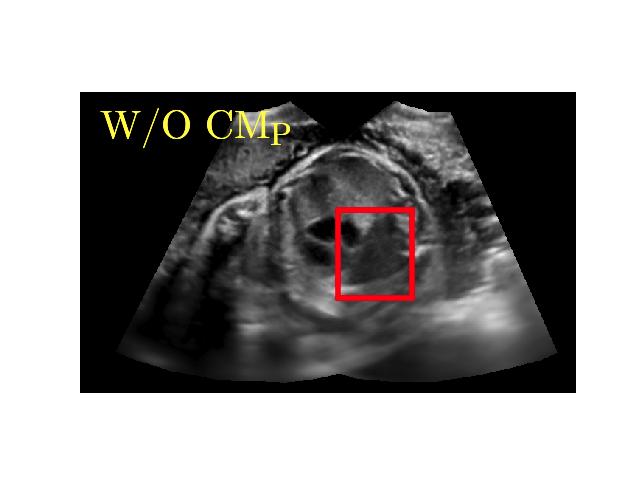} &
 \includegraphics[height=1.25cm]{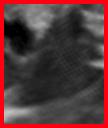} & \multirow{2}*[7pt]{\includegraphics[height=1.25cm, trim=4.5cm 1.5cm 4.5cm 1.5cm, clip]{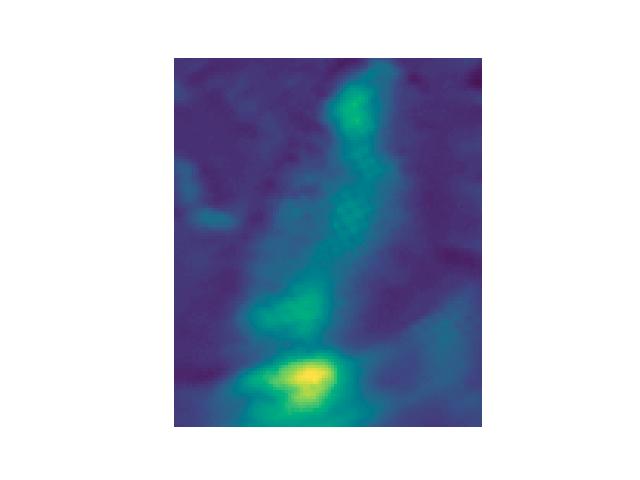}}\\
\includegraphics[height=1.25cm, trim=2.9cm 3.2cm 2cm 3.2cm, clip]{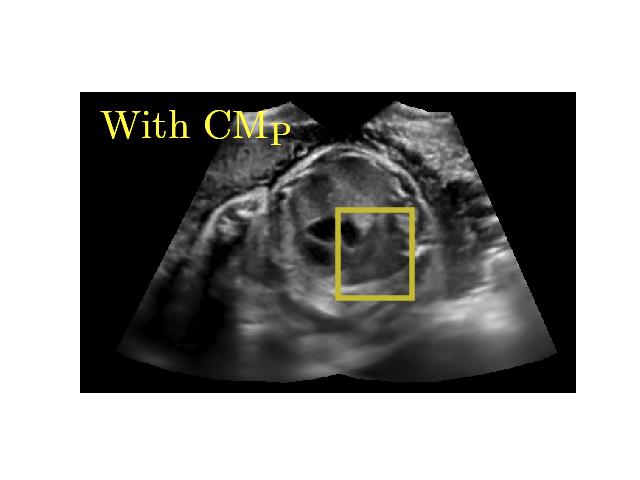} & \includegraphics[height=1.25cm]{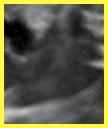}
  \end{tabular}
  } 
  \hfill
 \hspace{-3.5ex}
 \setcounter{subfigure}{3}
 \subfloat[Int. \& Gaussian, Sigmoid]{
 \begin{tabular}{@{\hspace{0.1\tabcolsep}}c@{\hspace{0.3\tabcolsep}}c@{\hspace{0.3\tabcolsep}}c@{\hspace{0.1\tabcolsep}}}
 \includegraphics[height=1.25cm, trim=2.9cm 3.2cm 2cm 3.2cm, clip]{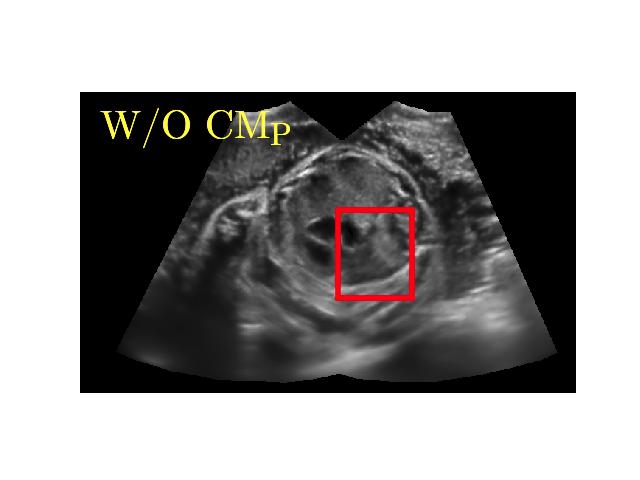} &
 \includegraphics[height=1.25cm]{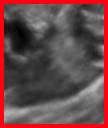}
 & \multirow{2}*[7pt]{\includegraphics[height=1.25cm, trim=4.5cm 1.5cm 4.5cm 1.5cm, clip]{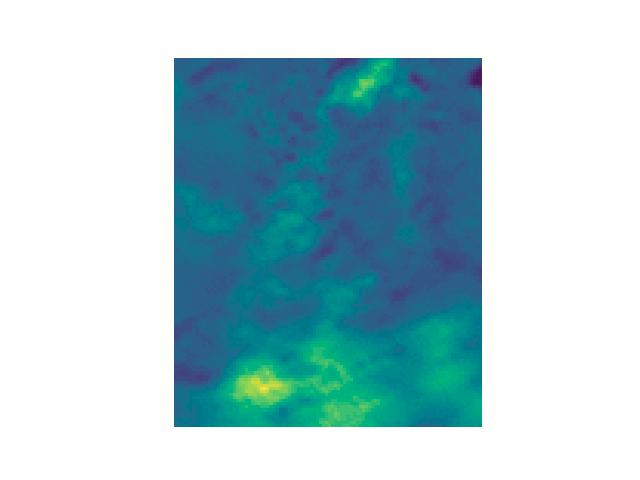}}\\
\includegraphics[height=1.25cm, trim=2.9cm 3.2cm 2cm 3.2cm, clip]{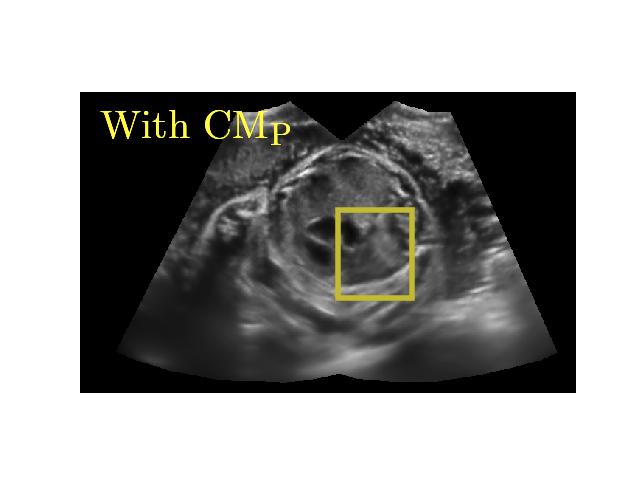} &
  \includegraphics[height=1.25cm]{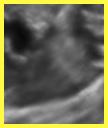}
  \end{tabular}
  } \\
   \setcounter{subfigure}{4}
 \subfloat[Gaussian, MSE, $+$]{
 \begin{tabular}{@{\hspace{0\tabcolsep}}c@{\hspace{0.3\tabcolsep}}c@{\hspace{0.3\tabcolsep}}c@{\hspace{0\tabcolsep}}}
   \includegraphics[height=1.25cm, trim=2.9cm 3.2cm 2cm 3.2cm, clip]{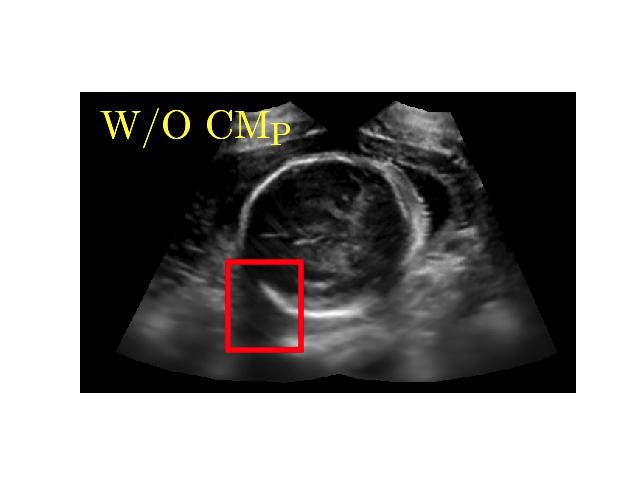} & \includegraphics[height=1.25cm]{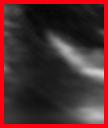} & \multirow{2}{*}[7pt]{\includegraphics[height=1.25cm, trim=4.5cm 1.5cm 4.5cm 1.5cm, clip]{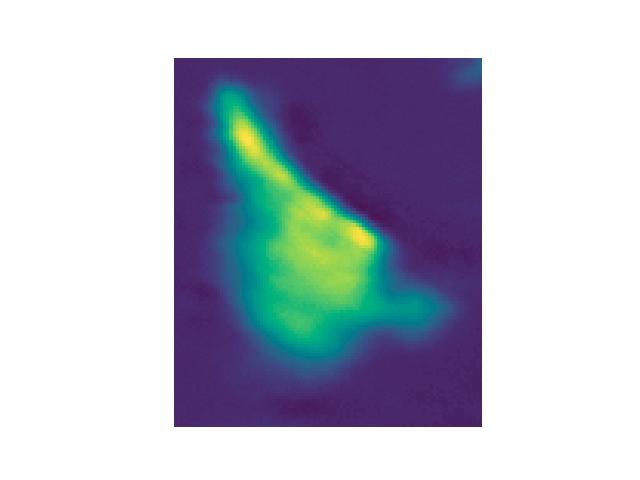}}\\
     \includegraphics[height=1.25cm, trim=2.9cm 3.2cm 2cm 3.2cm, clip]{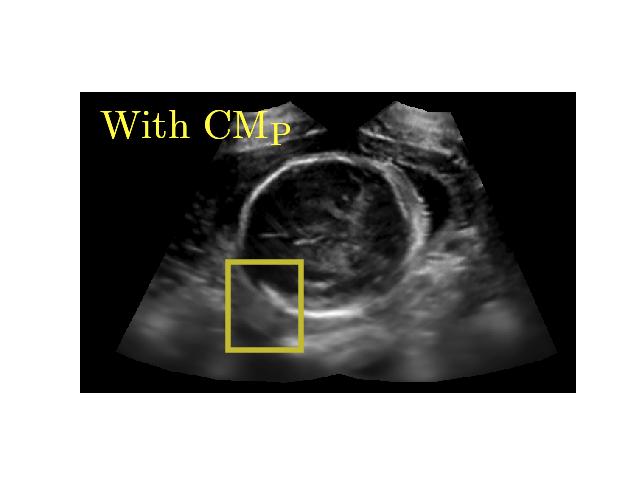}&\includegraphics[height=1.25cm]{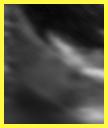} 
 \end{tabular}
 }
\hfill
\hspace{-3.5ex}
 \setcounter{subfigure}{5}
 \subfloat[Int. \& Gaussian, Sigmoid, $+$]{
 \begin{tabular}{@{\hspace{0.1\tabcolsep}}c@{\hspace{0.3\tabcolsep}}c@{\hspace{0.3\tabcolsep}}c@{\hspace{0.1\tabcolsep}}}
 \includegraphics[height=1.25cm, trim=2.9cm 3.2cm 2cm 3.2cm, clip]{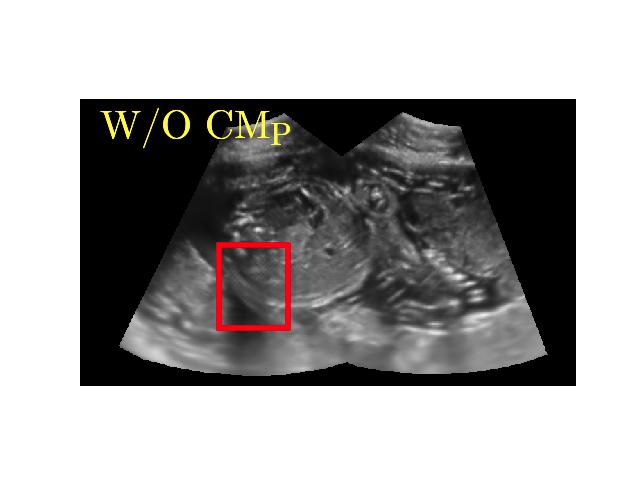} &
 \includegraphics[height=1.25cm]{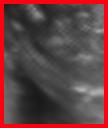}
 & \multirow{2}*[7pt]{\includegraphics[height=1.25cm, trim=4.5cm 1.5cm 4.5cm 1.5cm, clip]{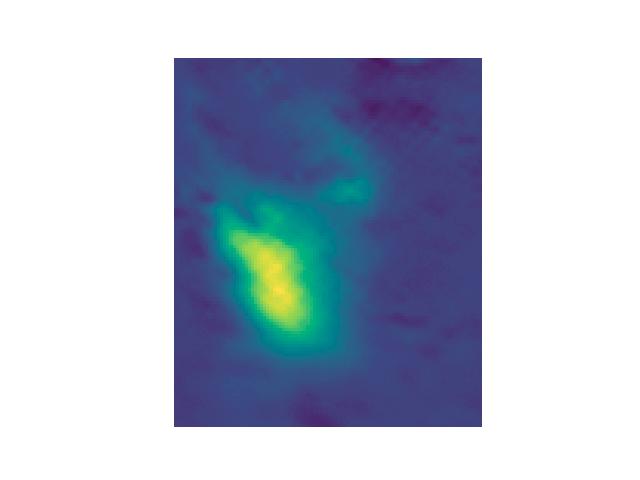}}\\
\includegraphics[height=1.25cm, trim=2.9cm 3.2cm 2cm 3.2cm, clip]{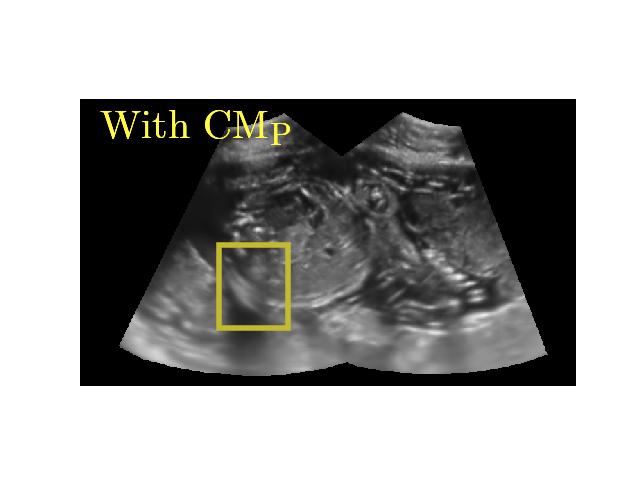} &
  \includegraphics[height=1.25cm]{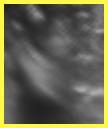}
  \end{tabular}
  }
  \hfill
\hspace{-3.5ex}
\setcounter{subfigure}{6}
\subfloat[Gaussian, Sigmoid, $-$]{
 \begin{tabular}{@{\hspace{0\tabcolsep}}c@{\hspace{0.3\tabcolsep}}c@{\hspace{0.3\tabcolsep}}c@{\hspace{0\tabcolsep}}}
 \includegraphics[height=1.25cm, trim=2.9cm 3.2cm 2cm 3.2cm, clip]{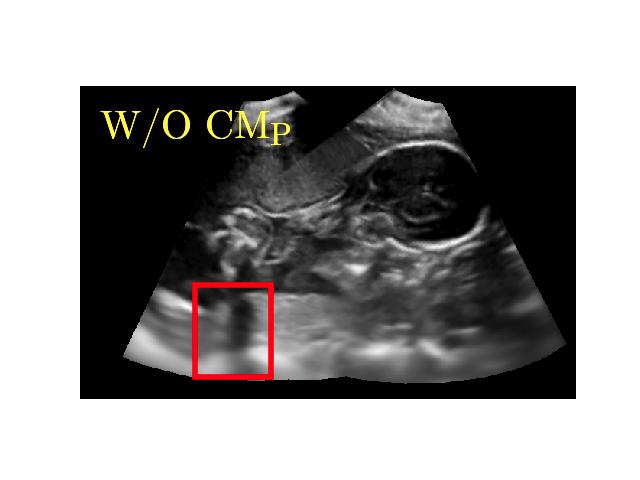} &
 \includegraphics[height=1.25cm]{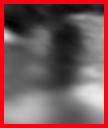} & \multirow{2}*[7pt]{\includegraphics[height=1.25cm, trim=4.5cm 1.5cm 4.5cm 1.5cm, clip]{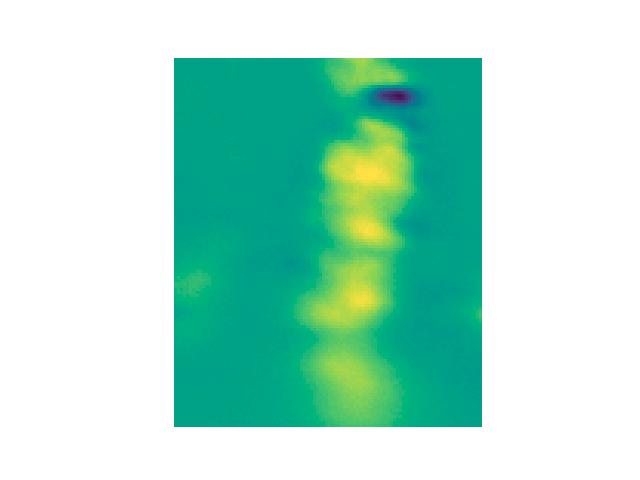}}\\
\includegraphics[height=1.25cm, trim=2.9cm 3.2cm 2cm 3.2cm, clip]{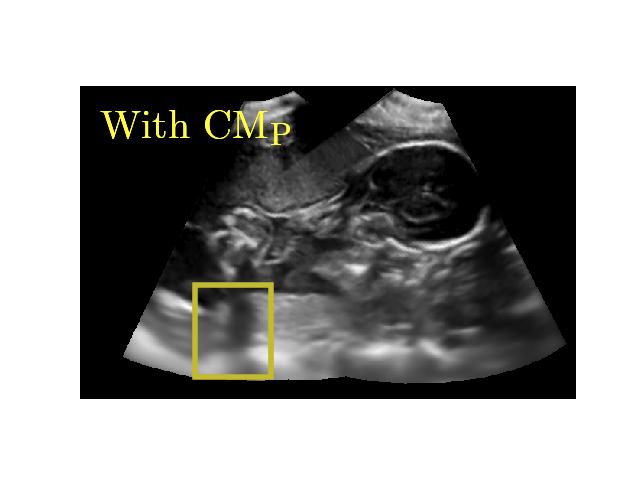} & \includegraphics[height=1.25cm]{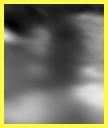}
  \end{tabular}
  } 
  \hfill
 \hspace{-3.5ex}
 \setcounter{subfigure}{7}
 \subfloat[Int. \& Gaussian, Sigmoid, $-$]{
 \begin{tabular}{@{\hspace{0.1\tabcolsep}}c@{\hspace{0.3\tabcolsep}}c@{\hspace{0.3\tabcolsep}}c@{\hspace{0.1\tabcolsep}}}
 \includegraphics[height=1.25cm, trim=2.9cm 3.2cm 2cm 3.2cm, clip]{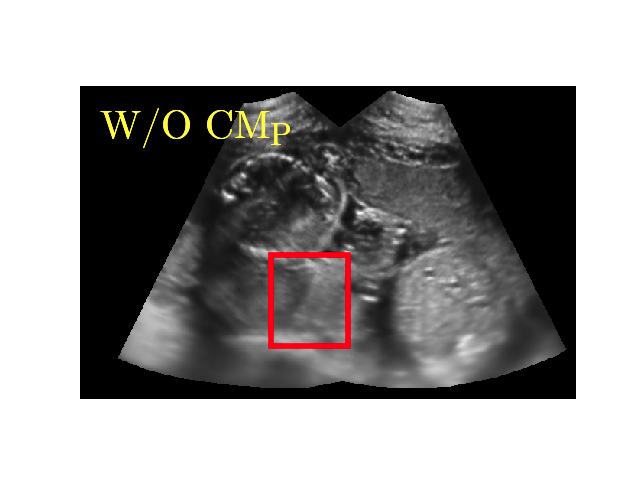} &
 \includegraphics[height=1.25cm]{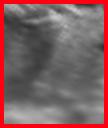}
 & \multirow{2}*[7pt]{\includegraphics[height=1.25cm, trim=4.5cm 1.5cm 4.5cm 1.5cm, clip]{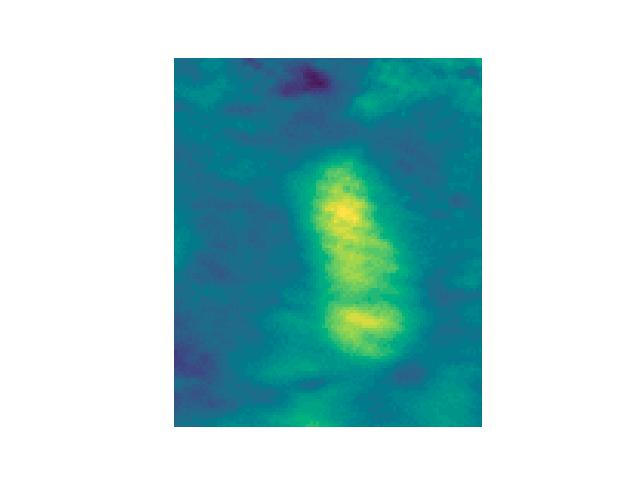}}\\
\includegraphics[height=1.25cm, trim=2.9cm 3.2cm 2cm 3.2cm, clip]{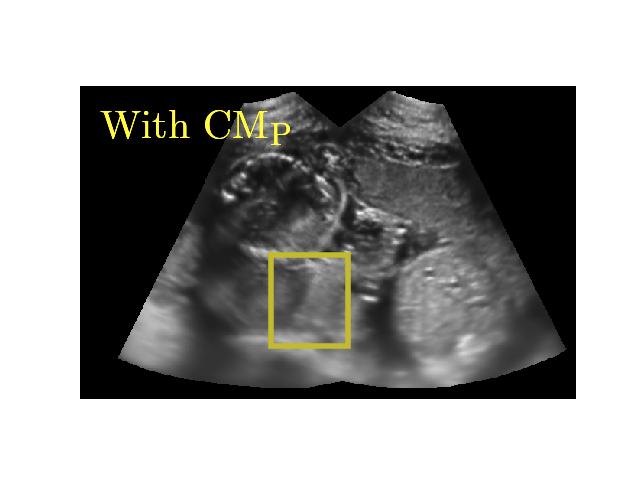} &
  \includegraphics[height=1.25cm]{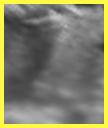}
  \end{tabular}
  } 
  \caption{Results of image fusion based on different weighting strategies and loss functions (Gaussian weighting vs. Intensity-and-Gaussian weighting (Int. \& Gaussian), MSE loss vs. Sigmoid loss). Note that the MSE loss and the Sigmoid loss are used for training of the confidence estimation network, which generates the shadow confidence maps. (a-d) are the image fusion results of the same case. (a,c) are the image fusion of Gaussian weighting with MSE loss and Sigmoid loss respectively and (b,d) are the results of Intensity-and-Gaussian weighting with MSE loss and Sigmoid loss respectively. (e-h) show the image fusion results on four different cases. (e,f) are examples for visually improved cases ($+$) showing notable positive differences of image fusion before and after adding ${CM}_P$ confirmed by our sonographers while (g,h) are cases with less change ($-$). For each sub-figure (e.g. (a)), in the first column, the top row is the result without integrating a shadow confidence map ${CM}_P$ and the bottom row is the result with integrated ${CM}_P$. The second column shows the corresponding enlarged framed areas of the images. The third column is the difference map of corresponding framed areas. The color bar on the top shows that the more yellow/brighter, the higher the difference between the two framed areas. }
  \label{imgFusion}
\end{figure*}

Fig.~\ref{imgFusion} qualitatively shows that shadow confidence maps are able to improve the performance of US image fusion algorithms with different weighting strategies. Fig.~\ref{imgFusion} also shows the difference between adding two different types of confidence maps. These two types of confidence maps are generated by the confidence estimation network which are separately trained by either MSE or Sigmoid loss. %Here, we use Fig.~\ref{imgFusion} to demonstrate the effectiveness of shadow confidence maps under various scenarios, instead of comparing the image fusion results under different weighting strategies or loss functions. 
Fig.~\ref{imgFusion} (a) to (d) illustrate image fusion results for the same case using different combinations of weighting strategies and loss functions. The difference maps indicate that shadow confidence maps are capable of improving image fusion performance. Fig.~\ref{imgFusion} (e) to (h) show image fusion results on four different cases. We randomly select two positively affected cases (Fig.~\ref{imgFusion} (e) and (f)) to show visual  improvement. We additionally show two randomly selected examples (Fig.~\ref{imgFusion} (g) and (h)) that don't show perceptually significant improvements after adding shadow confidence maps.
% , which has no significant effect on quantitative results in Sections A and B.  
Quantitative evaluation for image fusion is not possible because of lacking a ground truth for US compounding tasks. %Here, ${CM}_P$ has the same meaning as in Table.\ref{classification}. (\cref{appdImgF} shows detailed results)

\subsection{Automated Biometric Measurements}
%\subsubsection*{\textbf{Automated Biometric Measurements}}
We integrate our shadow confidence maps into an automatic biometric measurement approach~\cite{Sinclair2018}, and show the biometric measurement performance (measured by DICE) before and after adding shadow confidence maps. 

Similar to the ultrasound standard plane classification, shadow confidence maps are integrated into a biometric estimation model described in~\cite{Sinclair2018} as an extral channel. Specifically, we train and test four fully convolutional networks with the same hyper-parameters as detailed in ~\cite{Sinclair2018}, and use the same ellipse fitting algorithm described therein. The first network is trained only on the image data used in~\cite{Sinclair2018}. The other three networks are trained with an additional input channel for shadow confidence maps that are separately generated by the baseline, the proposed, and the proposed$+$AG method.

We show three examples that are affected by shadows, and show their biometric measurement results in Table~\ref{bioMet}. %Here, \emph{w/o $CM$}, ${CM}_B$, ${CM}_P$, ${CM}_{PAG}$ have the same meaning to those in Table.~\ref{classification}. 
From this experiment, we find that biometric measurement performance is boosted by up to $7\%$ for problematic failure cases after adding shadow confidence maps. The average performance on the entire test data set stays almost the same since only a small proportion of the test images are affected by strong shadows, mainly because of the image acquisition by highly skilled sonographers. 

%shadow confidence maps may introduce noise for other images. As a result, the average segmentation performance stays almost the same after adding shadow confidence maps. See \cref{appdBioM} for visual results.   

\begin{table}[t]
\centering
\caption{Biometric measurement performance (DICE) with vs. without shadow confidence maps.}
\label{bioMet}
\begin{threeparttable}
\begin{tabular}{*5c}
\toprule
~~~~~                                               & 
w/o $CM$                                              &
${CM}_B$                                            & 
${CM}_P$                                            & 
${CM}_{PAG}$                                          \\ 
\midrule                                            
$\# 1$                                                 &
0.947                                     &  
0.940                                               &
\textbf{0.988}                                    &
0.969                                             \\
$\# 2$                                                 &
0.956                                               &     
0.958                                               &
\textbf{0.974}                                               &
0.968                                              \\ 
$\# 3$                                                 &
0.880                                                &
0.915                                               &
0.923                                              &
\textbf{0.955}                                      \\
\midrule 
\textbf{Avg.}                                    &
0.966                                  &
0.964                                  &
0.965                                  &
0.964                                  \\
\bottomrule
\end{tabular}
\begin{tablenotes}
\item The symbols of the methods are the same to Table~\ref{classification}.
\end{tablenotes}
\end{threeparttable}
\end{table}

\section{Discussion}
In this paper, we propose a weakly supervised method to tackle the ill-defined problem of shadow detection in US. A na\"ive alternative to our method would be to train a fully supervised shadow segmentation network using pixel-wise annotation of shadow regions. However, pixel-wise annotation is infeasible because (a) accurately annotating a large number of images requires a vast amount of labour and time and has scanner dependencies (b) binary annotations of shadow regions would lead to high inter-observer variability as shadow features are poorly defined, and (c) real-valued annotations of shadow regions are affected by subjectivity of annotators.
% , and would be even time consuming.
%Moreover, binary shadow segmentation is difficult to describe the uncertain edges of shadow regions.
% This is a similar problem as trying to record observer annotation uncertainty as ground truth for  existing uncertainty estimation approaches.~\cite{eaton2018towards}.

%In this paper, we propose a weakly supervised CNN-based method to estimate confidence for shadow regions and produce dense shadow confidence maps. This method is able to generalize the shadow features extracted from a binary classification to be suitable for confidence estimation of shadow regions. 
% Although ground truth for confidence estimation is unavailable, qualitative and quantitative results indicate the effectiveness of our method when compared to the state-of-the-art and human performance. 

The performance of shadow region confidence estimation on different anatomical structures can be improved after integrating attention mechanisms. For example, the soft DICE is increased on $S_{test}$. This also results in improved ultrasound classification (Table~\ref{classification}). However, the quantitative results %of shadow segmentation on both test data sets 
show that attention mechanisms are not essential. Networks with attention mechanisms are sometimes outperformed by networks without attention mechanisms. This may be caused by the way we integrate the attention mechanism. Since we add attention gates to encoders of all networks, the shadow features are emphasized for the shadow/shadow-free classification, which increases the difficulty of generalizing shadow features from classification to shadow segmentation. 
%However, further experiments that only add attention gates to segmentation network and confidence estimation network are necessary to verify the usefulness of attention mechanism.

% In the proposed method, the loss function of the confidence estimation network is set as the mean squared error, but this loss can be measured by any other functions such as sigmoid cross-entropy. In the multi-view image fusion task, we additionally provide the image fusion results with confidence maps generated according to sigmoid cross-entropy loss. The experimentation indicates that the usefulness of shadow confidence maps are not affected by the loss function although small difference exist in individual examples.

% In the standard plane classification task, we use only a subset of data set in \cite{christian2017} instead of the whole data set because (1) we aim at verifying the usefulness of our method rather than improving performance of \cite{christian2017}, and (2) it is convenient to keep inter-class balance and to implement the experiments. 

We use MSE as the loss function of the confidence estimation network, but this loss can also be measured by other functions. Practically this choice has no effect on our quantitative results. However, in the image fusion task, we observe qualitative differences, which we show in Fig.~\ref{imgFusion} for Sigmoid cross-entropy loss.
%we additionally provide the image fusion results with confidence maps generated with Sigmoid cross entropy loss. 

In the standard plane classification task, we use only a subset of target standard planes compared to~\cite{christian2017} because (1) we aim at verifying the usefulness of our method rather than improving performance of~\cite{christian2017}, (2) it is desirable to keep inter-class balance to avoid side-effects from under-represented classes, and (3) we chose standard planes for which~\cite{christian2017} did not show optimal classification performance. 

% We train a transfer network to learn the intensity distribution modeled by a matrix $T(\cdot)$ for generating the reference confidence maps. An alternative way is only using $T(\cdot)$ to generate reference confidence maps, which shows no obvious difference on all the experiments to our current pipeline. However, since the piecewise functions in $T(\cdot)$ is non-differential, the transfer network enables our pipeline end-to-end trainable.

% We train a transfer network to learn the matrix $T(\cdot)$. An easy way is only using $T(\cdot)$ to generate reference confidence maps, which shows no obvious difference on all experiment results to our current pipeline. However, since $T(\cdot)$ is non-differential, the transfer network enables our pipeline end-to-end trainable.
%We use $T(\cdot)$ to generate reference confidence maps in the evaluations. As aforementioned, $T(\cdot)$ can also be approximated by continuous functions or CNNs so that the segmentation network and the confidence estimation network can be jointly fine-tuned. However, with coarse pixel-wise shadow annotations as weak segmentation ground truth and without ground truth for shadow confidence maps, evaluating the performance of this jointly fine-tune is infeasible.  

$T(\cdot)$, as defined in Eq.~\ref{T_1} or Eq.~\ref{T_2} is one example how prior knowledge can be integrated into the training process. If $T(\cdot)$ is chosen to be a continuous non-trainable function, e.g. quadratic or Gaussian, further weight relaxation can be introduced for joint refinement of both, the shadow-seg module in Fig.~\ref{overview}a and the confidence estimation in Fig.~\ref{overview}c. However, since probabilistic ground truth does not exist for our applications, evaluation would become purely subjective, thus we decide to use direct but discontinuous integration of shadow-intensity assumptions for $T(\cdot)$.

Task-specific deep networks, e.g. for classification, may inadvertently learn to ignore weak shadows in some cases, but the learning capacity of shadow properties is unknown. By estimating confidence of shadow regions independently, our method guarantees that shadow property information is separately extracted and can be seamlessly integrated into other image analysis algorithms.
With additional shadow property information, our method can improve steerability and interpretability for deep neural networks, and also enables extensions for non-deep learning algorithms. As shown in the experiments, prior knowledge provided by shadow confidence maps can improve the performance of various applications. 

Binary shadow segmentation generated by the shadow-seg module (Fig.1a) may provide shadow information to some extent. The easiest way to utilize shadow information is integrating this binary shadow segmentation into other applications. However, a binary segmentation of shadow regions is improper to describe inherent ambiguity of acoustic shadows caused by various attenuation of sound waves. Compared with binary shadow segmentation, a real-valued shadow confidence map is more reasonable to represent shadows, especially uncertain boundaries. With this more accurate representation, shadow confidence maps are able to improve the performance of other applications compared to using simple binary segmentation.%, such as ultrasound standard plane classification.  

\begin{figure}[t]
 \centering
  \subfloat{\includegraphics[height=1.7cm]{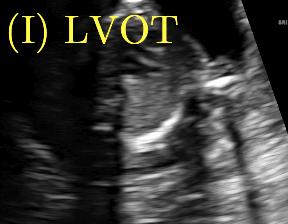}} \hfill 
  \subfloat{\includegraphics[height=1.7cm,  trim=2.29cm 1.29cm 1.88cm 1.48cm, clip]{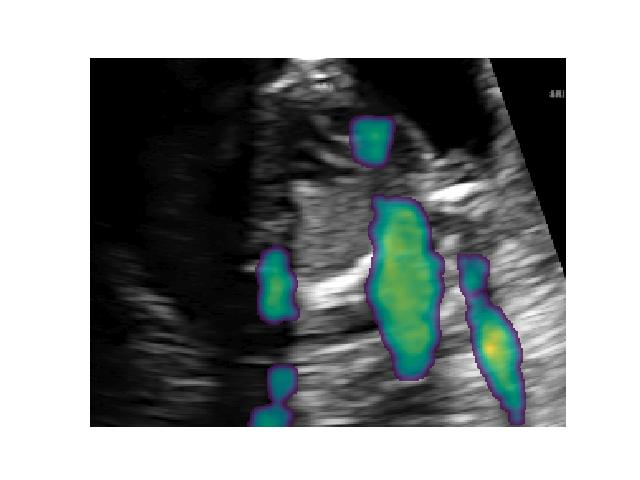}} \hfill 
  \subfloat{\includegraphics[height=1.7cm,  trim=2.29cm 1.29cm 1.88cm 1.48cm, clip]{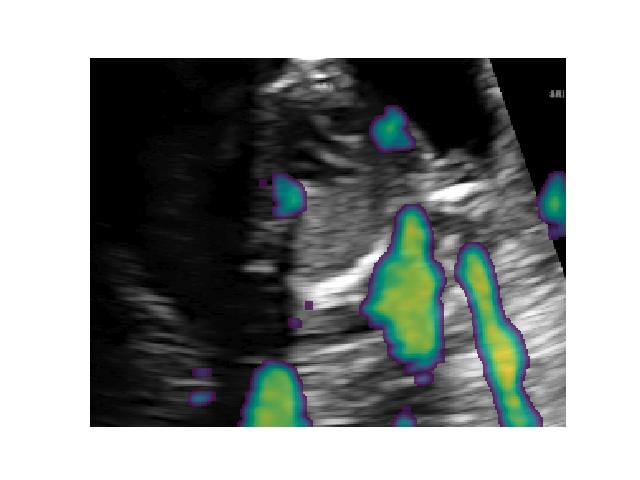}} \hfill 
  \subfloat{\includegraphics[height=1.7cm,  trim=2.29cm 1.29cm 1.88cm 1.48cm, clip]{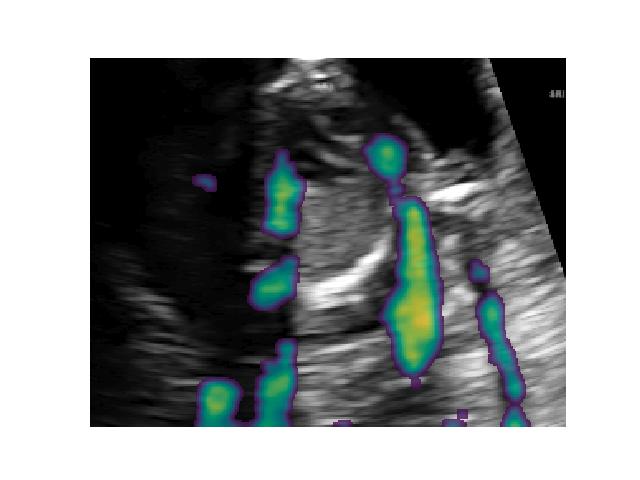}} \hfill 
  \\
 \setcounter{subfigure}{0}
   \subfloat[Image]{\includegraphics[height=1.7cm]{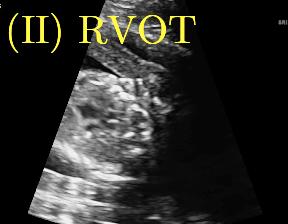}} \hfill 
  \subfloat[Baseline]{\includegraphics[height=1.7cm,  trim=2.29cm 1.29cm 1.88cm 1.48cm, clip]{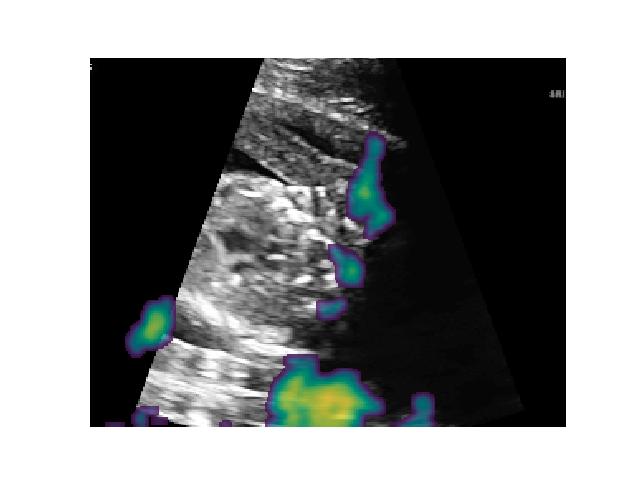}} \hfill
  \subfloat[Proposed]{\includegraphics[height=1.7cm,  trim=2.29cm 1.29cm 1.88cm 1.48cm, clip]{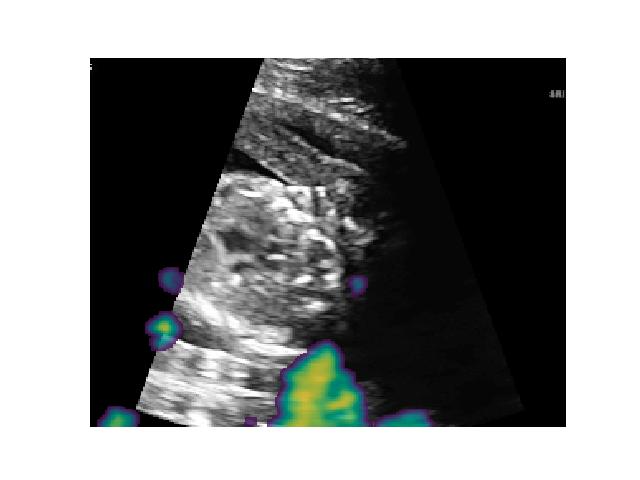}} \hfill 
  \subfloat[Proposed$+$AG]{\includegraphics[height=1.7cm,  trim=2.29cm 1.29cm 1.88cm 1.48cm, clip]{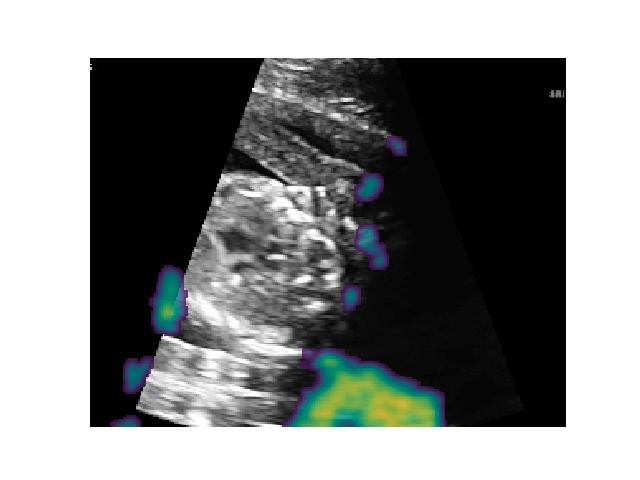}} \hfill 
  \\
  \caption{Qualitative performance of our methods for detecting signal lacking regions caused by insufficient gel.}
  \label{InsufficientShadow}
\end{figure}

Corrupted images such as images with shadows caused by insufficient acoustic impedance gel are excluded in the training. This type of shadows can be regarded as background since signals can hardly reach the tissues, and corrupted images with these shadows contain incomplete anatomical information. Additionally, during scanning, regions of missing signals caused by insufficient gel can be discovered and avoided in contrast to shadows generated by the interaction between signals and tissues. Therefore, our work excluded the corrupt images and focus on shadows within valid anatomy. Nevertheless, Fig.~\ref{InsufficientShadow} further shows that our proposed method is capable of indicating regions suffering from signal decay, especially on the boundaries.

% For the ultrasound standard plane classification, we only summarized the improvements of classification accuracy instead of statistically evaluating the significance of the improvements. Statistical methods such as the Paired Sample T-Test and the McNemar's test are not suitable in this case. In detail, the Paired Sample T-Test requires normal distribution variables which can hardly be the satisfied by the 9 classification accuracy. The McNemar's test only evaluates the difference of correct/incorrect predictions of each category rather than the difference of classification accuracy between two methods.

We use the coarse pixel-wise binary manual segmentation as ground truth for the shadow segmentation network and the transfer function since accurate manual annotation for shadow regions is unavailable as we discussed before. However, the inaccuracy of the coarse ground truth can hardly affect the quantitative assessments and the generation of reference confidence maps, because (1) DICE, recall, precision and MSE are still positively related to the effectiveness of the methods, (2) soft DICE and ICC are not related to the coarse ground truth, and (3) reference confidence maps are generated based on $I_{mean}$ (Eq.~\ref{Imean}), which smooth the influence of coarse ground truth by using mean intensity of TP regions. Additionally, we use human inter-observer variability which is computed by two coarse  binary manual annotations to further fairly assess the effectiveness of the methods.

%As we introduced in Section I, 
Acoustic shadows are caused by absorption, refraction or reflection of sound waves, which each leads to a different degree of signal attenuation. % and thus acoustic shadows behind these anatomies. 
% The proposed method is designed to estimate shadow confidence maps based on common shadow features appeared in the images, including intensity drop and uncertain boundaries, instead of aiming for acquisition-related causes of acoustic shadows.
Our method is predominately trained on fetal US images containing shadow regions with an elongated shape and a relatively strong drop of intensity. These are the shadow features that we have observed in a majority of images in our data sets. However, our method might be limited to perform effectively for shadows caused by different acquisition-related causes which are less well represented in our current training data.

% Although in some cases, However, we hypothesize that task-specific deep networks inadvertently learn to ignore weak shadows as well. The capacity to which degree shadows are learned is unknown, thus our approach allows to guarantee this information channel, seamlessly integrates into other deep networks, fosters interpretability and enables extensions for non-deep learning algorithms. 

%By providing additional information, shadow confidence maps can help to improve the performance of other applications. However, this additional information can also be redundant and misleading. As shown in Table.~\ref{classification}, the classification accuracy of abdominal, LVOT and RVOT decrease after adding shadow confidence maps.
%In particular, as shown in \cref{appdccm}, more RVOT is mistakenly classified as 3VV and LVOT is classified as 4CH after adding shadow confidence maps. This fact indicates that in classification task, shadow confidence maps may introduce misleading information, especially among similar anatomical structures. 

\section{Conclusion}
We propose a CNN-based, weakly supervised method for automatic confidence estimation of shadow regions in 2D US images. By learning and transferring shadow features from weakly-labelled images, our method can predict dense, continuous shadow confidence maps directly from input images.

We evaluate the performance of our method by comparing it to the state-of-the-art and human performance. Our experiments show that our method is quantitatively better than the state-of-the-art and human annotation for shadow segmentation. For confidence estimation of shadow regions, our method is also qualitatively better than the state-of-the-art and is more consistent than human annotation. More importantly, our method is capable of detecting disjoint multiple shadow regions without being limited by the correlation between adjacent pixels as in \cite{Karamalis2012}, and the heuristically selected hyperparameters in \cite{meng2018}. 

We further demonstrate that our method improves the performance of other automatic image analysis algorithms when integrating the obtained shadow confidence maps into other US applications such as standard plane classification, image fusion and automated biometric measurements.

Our method has significantly short inference time, which enables 
effective real-time feedback of local image properties. This feedback can 
guiding inexperienced sonographers to find diagnostically valuable viewing directions and pave the way for standardized image acquisition training.

% Our method has significantly shorter inference time compared with the state-of-the-art. Enabling real-time feedback of local image properties allows low overhead augmentation (this means to extend an existing algorithm with minimal additional computational costs.)of existing image analysis methods, guidance of inexperienced sonographers to find diagnostically valuable viewing directions and paves the way for standardized training.

%our method is potentially capable of helping train sonographers, as well as guiding non-experienced sonographers to easily find satisfied viewing direction during image acquisition process. 

\iftrue
\section*{Acknowledgment}
We thank the volunteers, sonographers and experts for providing manually annotated datasets and NVIDIA for their GPU donations. This work was supported by the Wellcome Trust IEH Award [102431], EPSRC grants (EP/L016796/1, EP/P001009/1), ERC 319456, and the Wellcome/EPSRC Center for Medical Engineering [WT 203148/Z/16/Z]. The research was funded/supported by the National Institute for Health Research (NIHR) Biomedical Research Center based at Guy's and St Thomas' NHS Foundation Trust, King's College London and the NIHR Clinical Research Facility (CRF) at Guy's and St Thomas'. Q. Meng is funded by the CSC-Imperial Scholarship. The views expressed are those of the author(s) and not necessarily those of the NHS, the NIHR or the Department of Health.
\fi

% \bibliographystyle{IEEEtran}
% \balance
% \bibliography{IEEEabrv,reference}

\begin{thebibliography}{10}

\bibitem{abbott1979}
J.~Abbott and F.~Thurstone.
\newblock Acoustic speckle: Theory and experimental analysis.
\newblock {\em Ultrasonic Imaging}, 1(4):303--324, 1979.

\bibitem{anbeek2005}
P.~Anbeek, K.~L. Vincken, G.~S. van Bochove, M.~J. van Osch, and J.~van~der
  Grond.
\newblock Probabilistic segmentation of brain tissue in mr imaging.
\newblock {\em NeuroImage}, 27(4):795 -- 804, 2005.

\bibitem{christian2017}
C.~Baumgartner, K.~Kamnitsas, J.~Matthew, T.~P. Fletcher, S.~Smith, L.~M. Koch,
  B.~Kainz, and D.~Rueckert.
\newblock Sononet: Real-time detection and localisation of fetal standard scan
  planes in freehand ultrasound.
\newblock {\em IEEE Trans. Med. Imag.}, 36:2204--2215, 2017.

\bibitem{Baumgartner2017}
C.~Baumgartner, L.~Koch, K.~Tezcan, J.~Ang, and E.~Konukoglu.
\newblock Visual feature attribution using wasserstein gans.
\newblock {\em CoRR}, abs/1711.08998, 2017.

\bibitem{Berton2016}
F.~Berton, F.~Cheriet, M.~M., and C.~Laporte.
\newblock Segmentation of the spinous process and its acoustic shadow in
  vertebral ultrasound images.
\newblock {\em Computers in Biology and Medicine}, 72:201--211, 2016.

\bibitem{Bouhemad2007}
B.~Bouhemad, M.~Zhang, Q.~Lu, and J.~Rouby.
\newblock Clinical review: bedside lung ultrasound in critical care practice.
\newblock {\em Critical Care}, 11(1):205, 2007.

\bibitem{Broersen2015}
A.~Broersen, M.~Graaf, J.~Eggermont, R.~Wolterbeek, P.~Kitslaar, J.~Dijkstra,
  J.~Bax, J.~Reiber, and A.~Scholte.
\newblock Enhanced characterization of calcified areas in intravascular
  ultrasound virtual histology images by quantification of the acoustic shadow:
  validation against computed tomography coronary angiography.
\newblock {\em Int J Cardiovasc Imaging}, 32:543--552, 2015.

\bibitem{swor2017}
{\relax Centre for Workforce Intelligence}.
\newblock Securing the future workforce supply – sonography workforce review.
\newblock 2017.

\bibitem{choi2015}
H.~Choi, J.~Lee, S.~Kim, and S.~Park.
\newblock Speckle noise reduction in ultrasound images using a discrete wavelet
  transform-based image fusion technique.
\newblock {\em Bio-Medical Materials and Engineering}, 26(1):1587--1597, 2015.

\bibitem{Coupe2009}
P.~Coup{\'e}, P.~Hellier, C.~Kervrann, and C.~Barillot.
\newblock Nonlocal means-based speckle filtering for ultrasound images.
\newblock {\em IEEE Trans. Image Process.}, 18(10):2221--2229, 2009.

\bibitem{DiceLR1945}
L.~R. Dice.
\newblock Measures of the amount of ecologic association between species.
\newblock {\em Ecology}, 26(3):297--302, 1945.

\bibitem{feldman2005}
M.~K. Feldman, S.~Katyal, and M.~S. Blackwood.
\newblock Us artifacts.
\newblock {\em Radio Graphics}, 29:1179–1189, 2009.

\bibitem{he2016}
K.~He, X.~Zhang, S.~Ren, and J.~Sun.
\newblock Deep residual learning for image recognition.
\newblock In {\em CVPR'16}, 2016.

\bibitem{he2016identity}
K.~He, X.~Zhang, S.~Ren, and J.~Sun.
\newblock Identity mappings in deep residual networks.
\newblock In {\em ECCV}, pages 630--645. Springer, 2016.

\bibitem{Hellier2010}
P.~Hellier, P.~Coup{\'e}, X.~Morandi, and D.~Collins.
\newblock An automatic geometrical and statistical method to detect acoustic
  shadows in intraoperative ultrasound brain images.
\newblock {\em Med Image Anal}, 14(2):195--204, 2010.

\bibitem{Karamalis2012}
A.~Karamalis, W.~Wein, T.~Klein, and N.~Navab.
\newblock Ultrasound confidence maps using random walks.
\newblock {\em Med Image Anal}, 16(6):1101--1112, 2012.

\bibitem{Kim2008}
H.~Kim and T.~Varghese.
\newblock Hybrid spectral domain method for attenuation slope estimation.
\newblock {\em Ultrasound Med Biol}, 34:1808--1819, 2008.

\bibitem{Klein2015}
T.~Klein and W.~Wells.
\newblock Rf ultrasound distribution-based confidence maps.
\newblock In {\em MICCAI'15}, pages 595--602. Springer, 2015.

\bibitem{Kremkau1986}
F.~W. Kremkau and K.~Taylor.
\newblock Artifacts in ultrasound imaging.
\newblock {\em J Ultrasound Med}, 5(4):227--237, 1986.

\bibitem{Krizhevsky2012}
A.~Krizhevsky, I.~Sutskever, and G.~Hinton.
\newblock Imagenet classification with deep convolutional neural networks.
\newblock In {\em NIPS'12}, pages 1097--1105, 2012.

\bibitem{Lange2009}
T.~Lange, N.~Papenberg, S.~Heldmann, J.~Modersitzki, B.~Fischer, H.~Lamecker,
  and P.~Schlag.
\newblock {3D} ultrasound-{CT} registration of the liver using combined
  landmark-intensity information.
\newblock {\em Int J Comput Assist Radiol Surg}, 4(1):79--88, 2009.

\bibitem{meng2018}
Q.~Meng, C.~Baumgartner, M.~Sinclair, J.~Housden, M.~Rajchl, A.~Gomez, B.~Hou,
  N.~Toussaint, V.~Zimmer, J.~Tan, et~al.
\newblock Automatic shadow detection in 2d ultrasound images.
\newblock In {\em MICCAI Workshop on PIPPI}, 2018.

\bibitem{screening2015}
{NHS}.
\newblock {\em Fetal anomaly screening programme: programme handbook June
  2015}.
\newblock Public Health England, 2015.

\bibitem{Noble2010}
J.~A. Noble.
\newblock Ultrasound image segmentation and tissue characterization.
\newblock {\em Proc Inst Mech Eng H.}, 224(2):307--316, 2010.

\bibitem{ozan2018}
O.~Oktay, J.~Schlemper, L.~L. Folgoc, M.~Lee, M.~P. Heinrich, K.~Misawa,
  K.~Mori, S.~G. McDonagh, N.~Y. Hammerla, B.~Kainz, et~al.
\newblock Attention u-net: Learning where to look for the pancreas.
\newblock {\em CoRR}, abs/1804.03999, 2018.

\bibitem{pawlowski2017state}
N.~Pawlowski, S.~I. Ktena, M.~Lee, B.~Kainz, D.~Rueckert, B.~Glocker, and
  M.~Rajchl.
\newblock Dltk: State of the art reference implementations for deep learning on
  medical images.
\newblock {\em arXiv preprint arXiv:1711.06853}, 2017.

\bibitem{penney2004}
G.~P. Penney, J.~M. Blackall, M.~S. Hamady, T.~Sabharwal, A.~Adam, and D.~J.
  Hawkes.
\newblock Registration of freehand 3d ultrasound and magnetic resonance liver
  images.
\newblock {\em Med Image Anal}, 8:81--91, 2004.

\bibitem{Rajchl2017}
M.~Rajchl, M.~Lee, O.~Oktay, K.~Kamnitsas, J.~Passerat-Palmbach, W.~Bai,
  M.~Damodaram, M.~Rutherford, J.~Hajnal, B.~Kainz, et~al.
\newblock Deepcut: Object segmentation from bounding box annotations using
  convolutional neural networks.
\newblock {\em IEEE Trans. Med. Imag.}, 36(2):674--683, 2017.

\bibitem{salomon2011}
L.~J. Salomon, Z.~Alfirevic, V.~Berghella, C.~Bilardo, E.~Hernandez-Andrade,
  S.~L. Johnsen, K.~Kalache, K.~Leung, G.~Malinger, H.~Munoz, et~al.
\newblock Practice guidelines for performance of the routine mid‐trimester
  fetal ultrasound scan.
\newblock {\em Ultrasound Obst Gyn}, 37:116--126, 2011.

\bibitem{shen2018}
T.~Shen, T.~Zhou, G.~Long, J.~Jiang, S.~Pan, and C.~Zhang.
\newblock Disan: Directional self-attention network for rnn/cnn-free language
  understanding.
\newblock In {\em AAAI}, 2018.

\bibitem{shrout1979}
P.~E. Shrout and J.~L. Fleiss.
\newblock Intraclass correlations: Uses in assessing rater reliability.
\newblock {\em Psychol Bull.}, 86(2):420--428, 1979.

\bibitem{Sinclair2018}
M.~Sinclair, C.~Baumgartner, J.~Matthew, W.~Bai, J.~Cerrolaza, Y.~Li, S.~Smith,
  C.~Knight, B.~Kainz, J.~Hajnal, et~al.
\newblock Human-level performance on automatic head biometrics in fetal
  ultrasound using fully convolutional neural networks.
\newblock In {\em EMBC'18}, 2018.

\bibitem{Springeberg2014}
J.~Springenberg, A.~Dosovitskiy, T.~Brox, and M.~Riedmiller.
\newblock Striving for simplicity: The all convolutional net.
\newblock {\em CoRR}, abs/1412.6806, 2014.

\bibitem{steel2005}
R.~Steel, T.~L. Poepping, R.~S.Thompson, and C.~Macaskill.
\newblock Origins of the edge shadowing artefact in medical ultrasound imaging.
\newblock {\em Ultrasound Med Biol}, 39:1153--1162, 2005.

\bibitem{wilson2018}
A.~C. Wilson, R.~Roelofs, M.~Stern, N.~Srebro, and B.~Recht.
\newblock The marginal value of adaptive gradient methods in machine learning.
\newblock {\em CoRR}, abs/1705.08292, 2018.

\bibitem{zhang2018}
Y.~Zhang, Y.~Tian, Y.~Kong, B.~Zhong, and Y.~Fu.
\newblock Residual dense network for image super-resolution.
\newblock In {\em CVPR'18}, 2018.

\bibitem{Zhou2016}
B.~Zhou, A.~Khosla, A.~Lapedriza, A.~Oliva, and A.~Torralba.
\newblock Learning deep features for discriminative localization.
\newblock In {\em CVPR'16}, pages 2921--2929. IEEE, 2016.

\bibitem{Zhu2017}
J.~Zhu, T.~Park, P.~Isola, and A.~A. Efros.
\newblock Unpaired image-to-image translation using cycle-consistent
  adversarial networks.
\newblock In {\em ICCV'17}, 2017.

\bibitem{veronika2018}
A.~V. Zimmer, A.~Gomez, Y.~Noh, N.~Toussaint, B.~Khanal, R.~Wright, L.~Peralta,
  V.~M. Poppel, E.~Skelton, J.~Matthew, et~al.
\newblock Multi-view image reconstruction: Application to fetal ultrasound
  compounding.
\newblock In {\em MICCAI Workshop on PIPPI}, 2018.

\end{thebibliography}

\bibliographystyle{abbrv}
\balance

\newpage

\appendix

\subsection{Shadow/Shadow-free Classification Network}
\label{appdclassN}
\begin{figure}[h]
 \centering
 \includegraphics[width=0.5\textwidth]{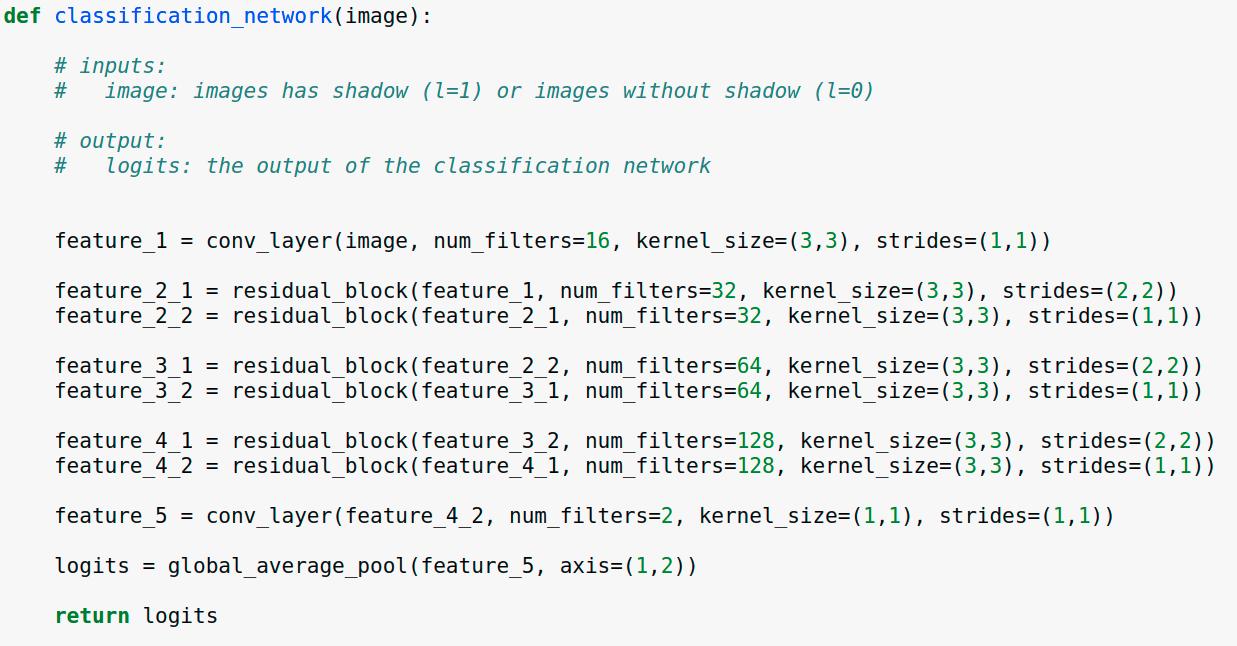}
 \caption{Shadow/shadow-free classification network architecture.}
 \label{classN}
\end{figure}

In this section, we use Python-inspired pseudo code to present the detailed network architecture of the shadow/shadow-free classification network (shown in Fig.~\ref{classN}). The {\fontfamily{qcr}\selectfont conv\_layer} function performs a standard 2D convolution without activation layer and the {\fontfamily{qcr}\selectfont global\_average\_pool} operates spatial averaging on the feature maps. The {\fontfamily{qcr}\selectfont residual\_block} is realized by DLTK~\cite{pawlowski2017state}. 

\subsection{Shadow Segmentation Network}
\label{appdsegN}
\begin{figure}[h]
 \centering
 \includegraphics[width=0.5\textwidth]{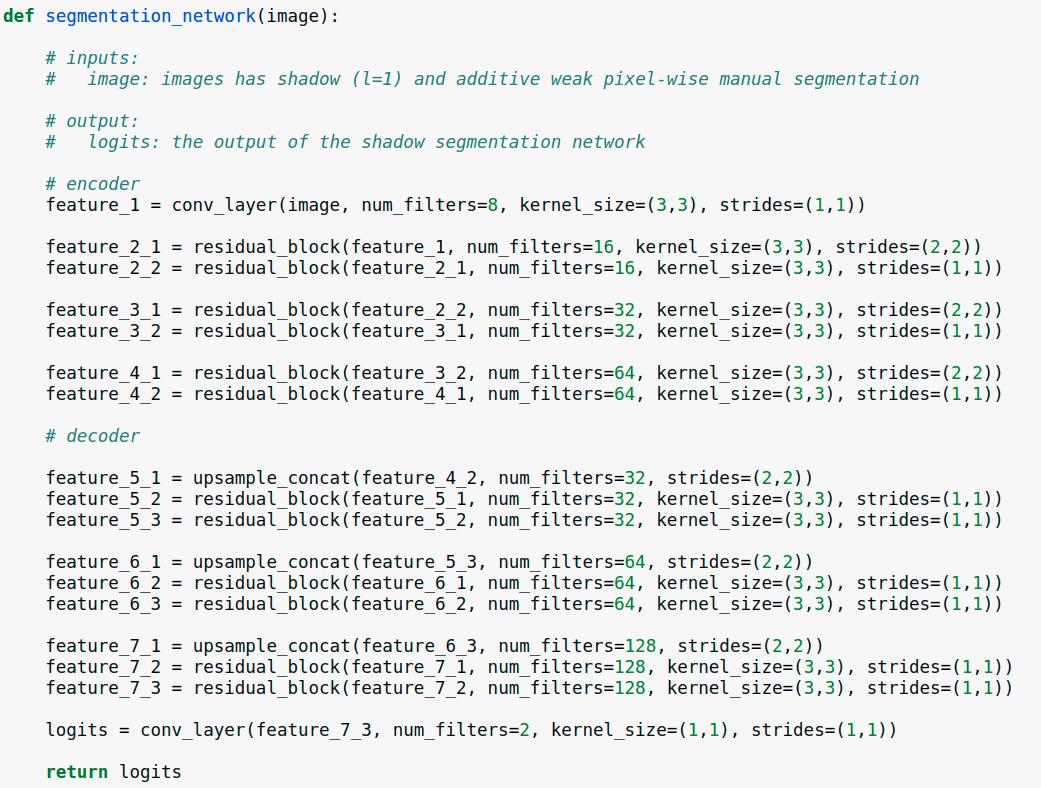}
 \caption{Shadow segmentation network architecture.}
 \label{segN}
\end{figure}

Fig.\ref{segN} shows the detailed architecture of the segmentation network. The {\fontfamily{qcr}\selectfont conv\_layer} function performs a standard 2D convolution without activation layer. The {\fontfamily{qcr}\selectfont residual\_block} and the {\fontfamily{qcr}\selectfont upsample\_concat} (the upsampling and concatenation layer) are realized by DLTK~\cite{pawlowski2017state}.

\subsection{Confidence Estimation Network}
\label{appdconfN}
\begin{figure}[h]
 \centering
 \includegraphics[width=0.5\textwidth]{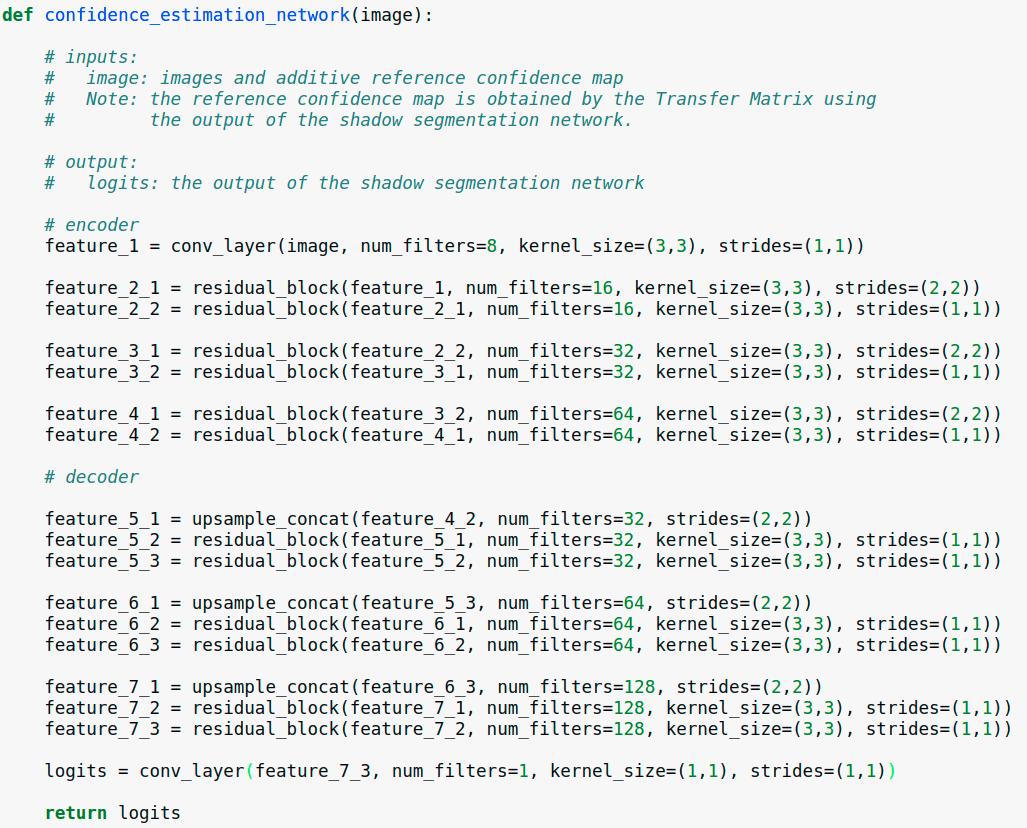}
 \caption{Shadow confidence estimation network architecture.}
 \label{confN}
\end{figure}

Fig.~\ref{confN} shows the detailed architecture of the shadow confidence estimation network. Similarly, the {\fontfamily{qcr}\selectfont conv\_layer} function performs a standard 2D convolution without activation layer. The {\fontfamily{qcr}\selectfont residual\_block} and the {\fontfamily{qcr}\selectfont upsample\_concat} (the upsampling and concatenation layer) are realized by DLTK~\cite{pawlowski2017state}.

\subsection{Alternative Examples of Shadow Confidence Estimation}
\label{appdexp}

\begin{figure*}[h]
  \centering
    
  \begin{tikzpicture}
  \begin{axis}[
     hide axis,
     scale only axis,
    height=0cm,
    width=17.5cm,
    colormap/viridis,
      colorbar horizontal,
    point meta min=0,
    point meta max=1,
    colorbar style={
        height=5,                 % Höhe der Colorbar
      xtick={0,0.2,0.4,0.6,0.8,1},
      tick label style={font=\tiny},
      xticklabel pos=upper
    }]
    % \addplot [] {};
  \end{axis}
  \end{tikzpicture}
 \\
 
  \subfloat{\includegraphics[height=2.75cm]{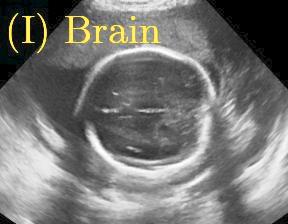}} \hfill
  \subfloat{\includegraphics[height=2.75cm,  trim=2.29cm 1.29cm 1.88cm 1.48cm, clip]{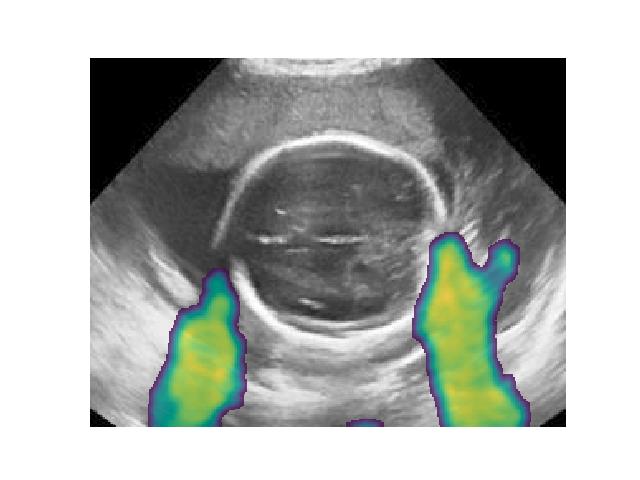}} \hfill
  \subfloat{\includegraphics[height=2.75cm,  trim=2.29cm 1.29cm 1.88cm 1.48cm, clip]{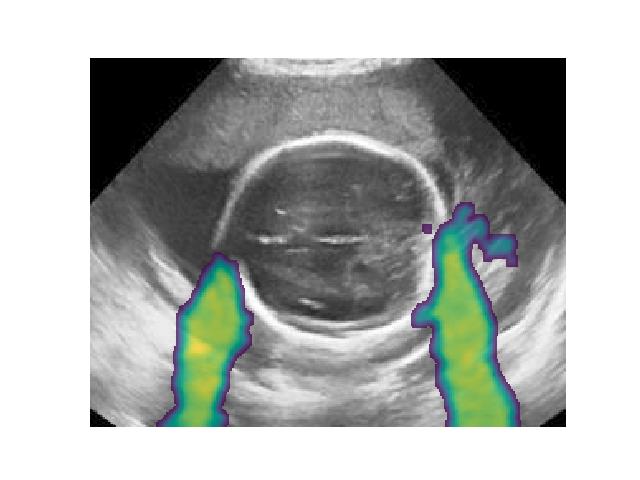}} \hfill
  \subfloat{\includegraphics[height=2.75cm,  trim=2.29cm 1.29cm 1.88cm 1.48cm, clip]{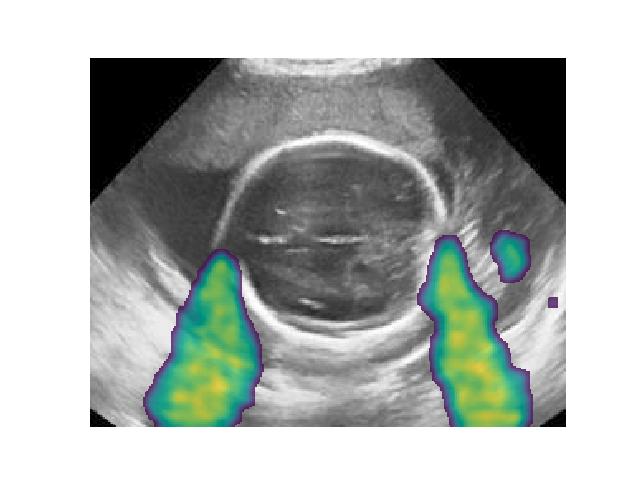}} \hfill
  \subfloat{\includegraphics[height=2.75cm,  trim=2.29cm 1.29cm 1.88cm 1.48cm, clip]{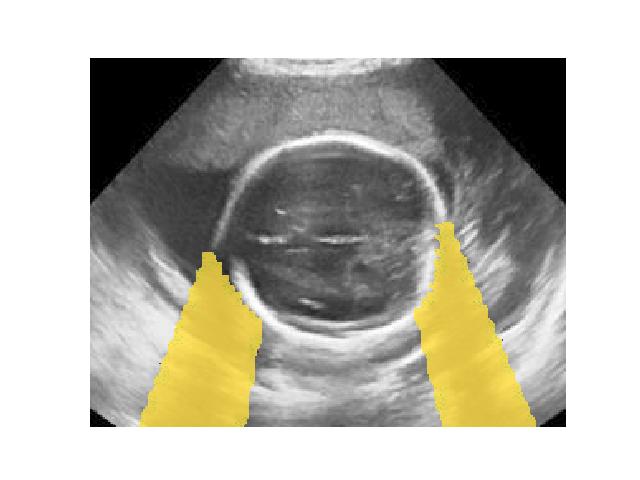}} \hfill  \\

  \subfloat{\includegraphics[height=2.75cm]{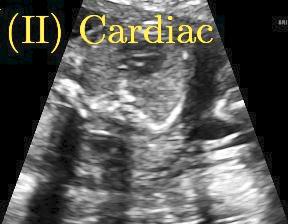}} \hfill
  \subfloat{\includegraphics[height=2.75cm, trim=2.29cm 1.29cm 1.88cm 1.48cm, clip]{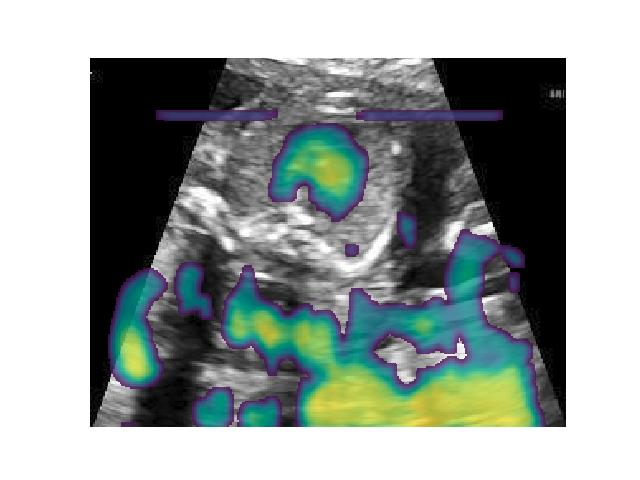}} \hfill
  \subfloat{\includegraphics[height=2.75cm, trim=2.29cm 1.29cm 1.88cm 1.48cm, clip]{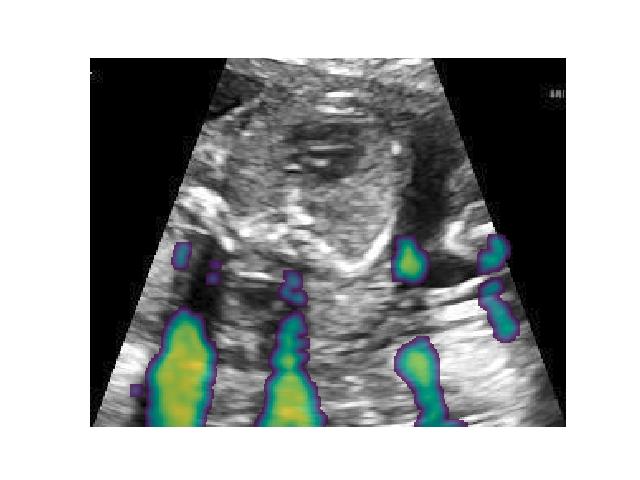}} \hfill
  \subfloat{\includegraphics[height=2.75cm, trim=2.29cm 1.29cm 1.88cm 1.48cm, clip]{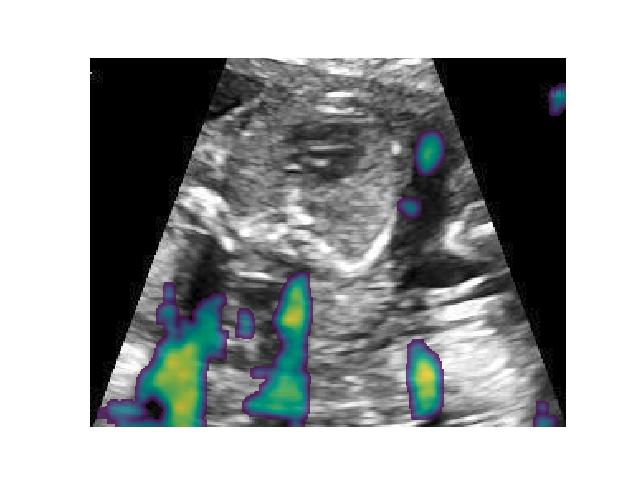}} \hfill
  \subfloat{\includegraphics[height=2.75cm, trim=2.29cm 1.29cm 1.88cm 1.48cm, clip]{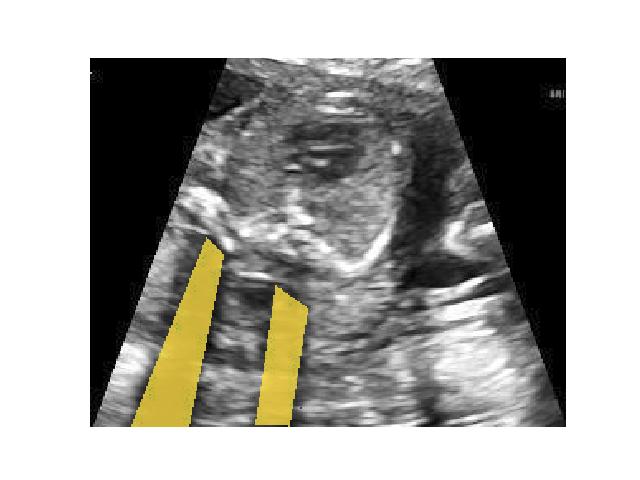}} \hfill \\
  
  \subfloat{\includegraphics[height=2.75cm]{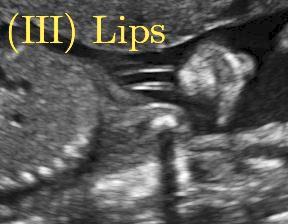}} \hfill
  \subfloat{\includegraphics[height=2.75cm, trim=2.29cm 1.29cm 1.88cm 1.48cm, clip]{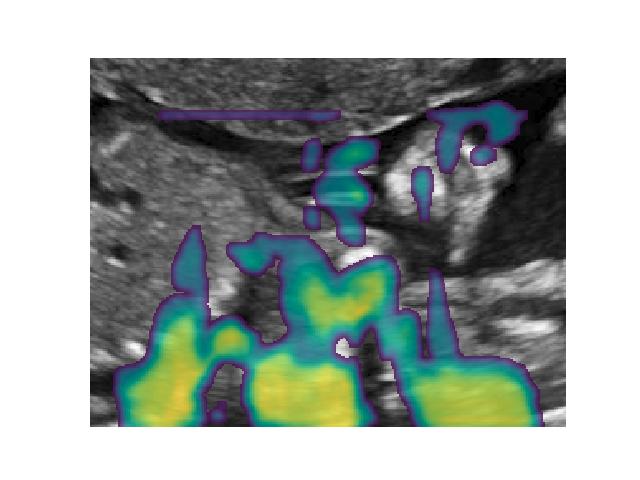}} \hfill
  \subfloat{\includegraphics[height=2.75cm, trim=2.29cm 1.29cm 1.88cm 1.48cm, clip]{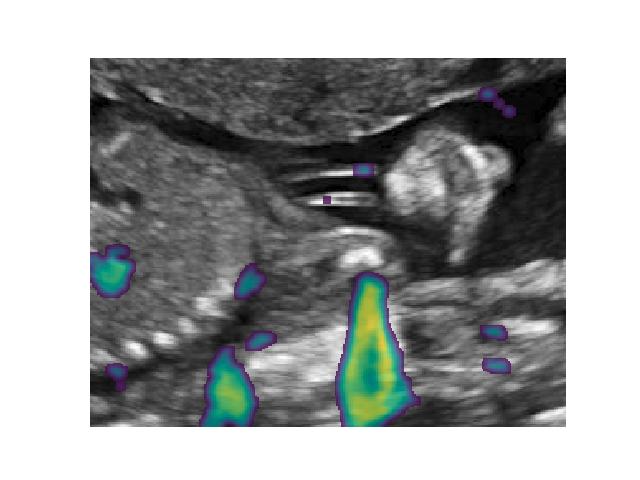}} \hfill
  \subfloat{\includegraphics[height=2.75cm, trim=2.29cm 1.29cm 1.88cm 1.48cm, clip]{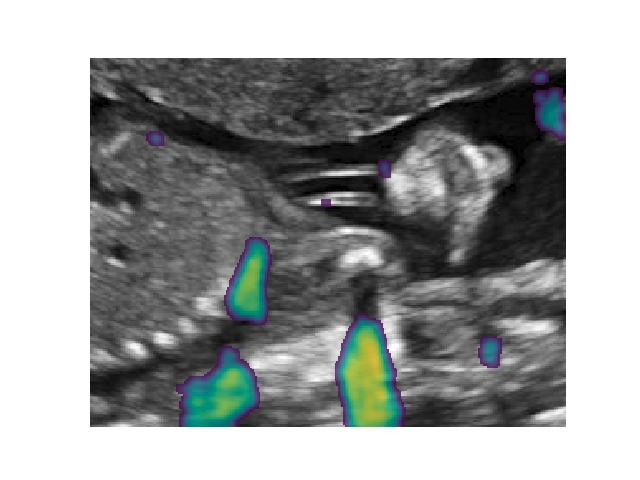}} \hfill
  \subfloat{\includegraphics[height=2.75cm, trim=2.29cm 1.29cm 1.88cm 1.48cm, clip]{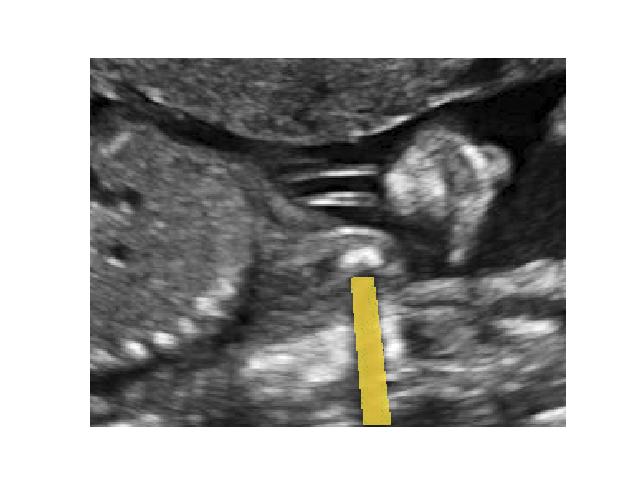}} \hfill \\
  
 \setcounter{subfigure}{0}
  \subfloat[Image]{\includegraphics[height=2.75cm]{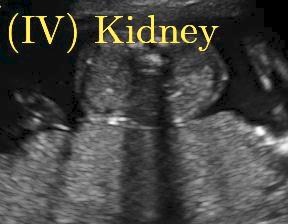}} \hfill
  \subfloat[Baseline]{\includegraphics[height=2.75cm, trim=2.29cm 1.29cm 1.88cm 1.48cm, clip]{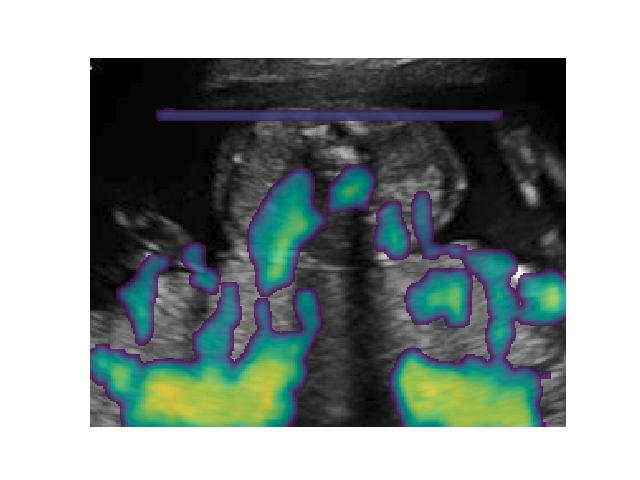}} \hfill
  \subfloat[Proposed]{\includegraphics[height=2.75cm, trim=2.29cm 1.29cm 1.88cm 1.48cm, clip]{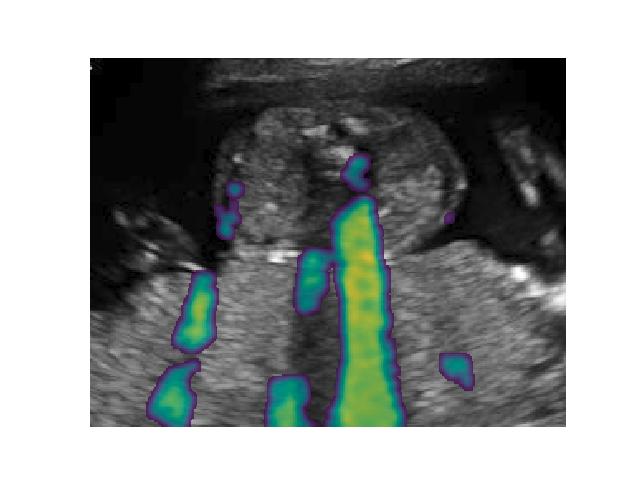}} \hfill
  \subfloat[Proposed$+$AG]{\includegraphics[height=2.75cm, trim=2.29cm 1.29cm 1.88cm 1.48cm, clip]{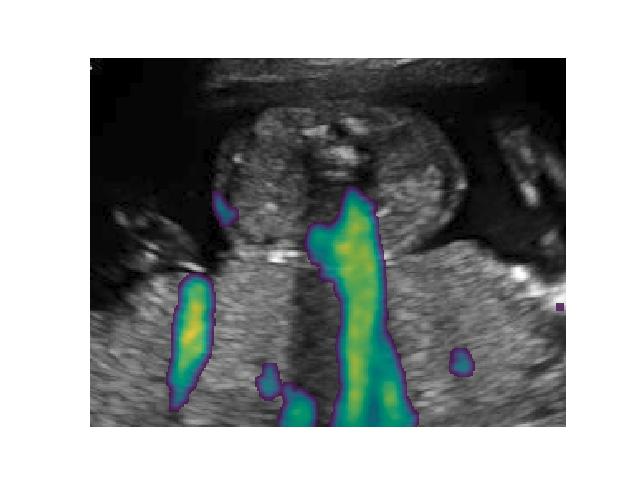}} \hfill
  \subfloat[Weak GT]{\includegraphics[height=2.75cm, trim=2.29cm 1.29cm 1.88cm 1.48cm, clip]{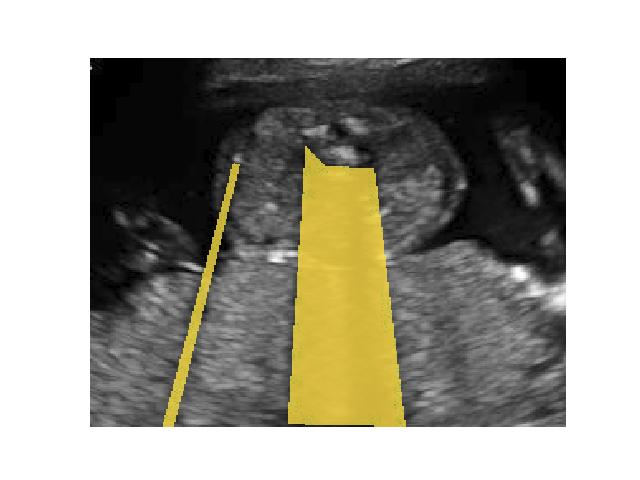}} \hfill \\
  
  \caption{Shadow confidence maps of different methods on various anatomical US images. Rows I-IV show four examples of shadow confidence estimation; Brain (top), Cardiac (middle), Lips (third) and Kidney (bottom). Columns (b-d) are shadow confidence maps from the baseline, the proposed method and the proposed method with attention gate (Proposed$+$AG). (f) is the binary map of manual segmentation.}
  \label{confEst}
\end{figure*}

We show an alternative group of examples for the confidence estimation of shadow regions (shown in Fig.~\ref{confEst}). These examples include fetal brain from $M_{test}$, and cardiac, lips, kidney from $S_{test}$. Similar to the Fig.~\ref{compareConf} in the main paper, Fig.~\ref{confEst} shows that the baseline fails to handle unseen data while the proposed method and the proposed$+$AG method are able to predict pixel-wise confidence of multiple shadow regions. These examples demonstrate that the shadow-seg module is able to generalize the shadow representation and transfer shadow representation from the shadow/shadow-free classification task to a confidence estimation task.

\subsection{Data in Ultrasound Classification}
\label{appddata}
Table~\ref{datasetsplit} shows the exact number of data used in the application of 2D US standard plane classification (Section V. Part $A$). The training data of each class is almost the same so that we can keep class balance between different classes during training.

\begin{table}[h]
\centering
\caption{Summary of the Data Set used in Ultrasound Standard Plane Classification Task.}
\label{datasetsplit}
\begin{tabular}{*4c}
\toprule
\textbf{Class}                                  & 
\textbf{Training}                                   & 
\textbf{Validation}                                 & 
\textbf{Testing}                                          \\ 
3VV                                                 &
1480                                               &
50                                               &     
298                                               \\
4CH                                                 &
1544                                               &
50                                               &
309                                               \\ 
Abdominal                                           &
2000                                              & 
50                                               & 
553                                               \\
Brain(Cb.)                                          &
2000                                              &
50                                               & 
634                                              \\
Brain(Tv.)                                          &
2000                                               & 
50                                               & 
899                                               \\
Femur                                               &
2000                                               & 
50                                               & 
520                                               \\
Lips                                                &
2000                                               & 
50                                              & 
526                                              \\
LVOT                                                &
1633                                               &
50                                               & 
333                                               \\
RVOT                                                &
1432                                               &
50                                               & 
296                                               \\
\textbf{Sum}                                    &
16089                                               &
450                                               & 
4368                                               \\ 
\bottomrule
\end{tabular}
\end{table}

\subsection{Class Confusion Matrix}
\label{appdccm}
Fig.~\ref{cm} additionally shows the class confusion matrix of 2D US standard plane classification in Section V. Part A. This class confusion matrix demonstrates that less 3VV images are mis-classified as RVOT images and less 4CH images are mis-classified as LVOT images after adding shadow confidence maps. However, as we discussed in the above Discuss Section, the shadow confidence maps can also introduce redundant information for similar anatomical structures in this classification task. For example, more LVOT images are wrongly classied as RVOT and more RVOT images are classified as 3VV images.

\subsection{Examples for Image Fusion}
\label{appdImgF}

\begin{figure*}[htb]
  \centering
    \begin{tikzpicture}
  \begin{axis}[
     hide axis,
     scale only axis,
    height=0cm,
    width=17.5cm,
    colormap/viridis,
      colorbar horizontal,
    point meta min=0,
    point meta max=1,
    colorbar style={
        height=5,                 % Höhe der Colorbar
      xtick={0.13, 1},
      xticklabels={$\text{low difference}$, $\text{high difference}$},
      tick label style={font=\small},
      xticklabel pos=upper,
      xticklabel style = {xshift=-1.2cm},
    }]
    % \addplot [] {};
  \end{axis}
  \end{tikzpicture}
 \\
  \vspace{-5pt}
  
  \subfloat{\includegraphics[height=1.75cm]{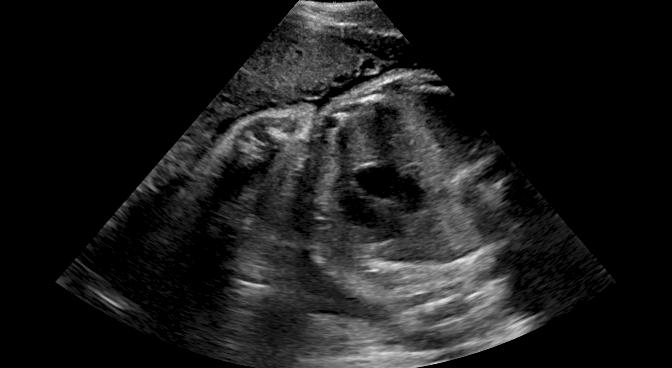}} \hfill
  \subfloat{\includegraphics[height=1.75cm]{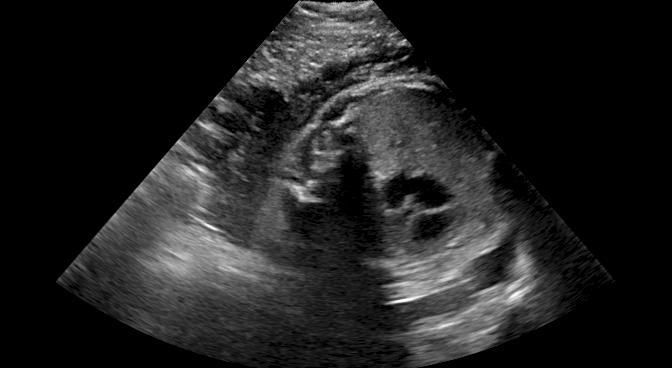}} \hfill
  \subfloat{\includegraphics[height=1.75cm, trim=3cm 3.2cm 3cm 3.2cm, clip]{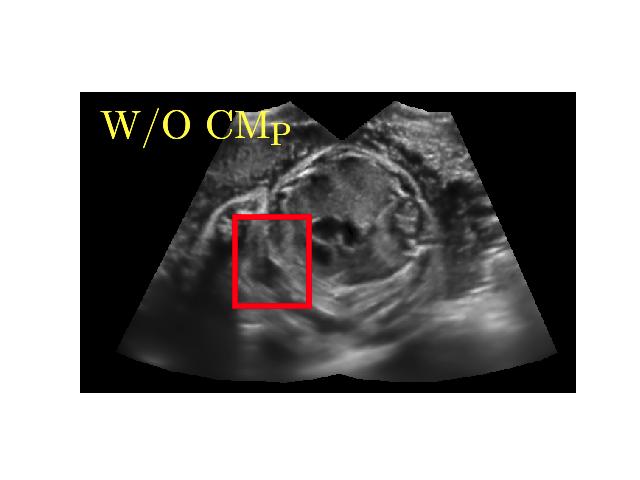}} \hfill 
  \subfloat{\includegraphics[height=1.75cm, trim=3cm 3.2cm 3cm 3.2cm, clip]{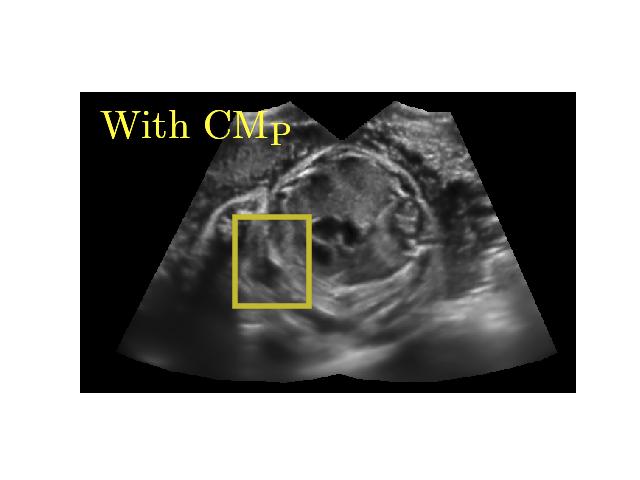}} \hfill
  \subfloat{\includegraphics[height=1.75cm]{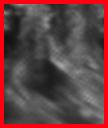}} \hfill
  \subfloat{\includegraphics[height=1.75cm]{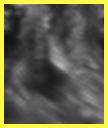}} \hfill
  \subfloat{\includegraphics[height=1.75cm, trim=4.5cm 1.5cm 4.5cm 1.5cm, clip]{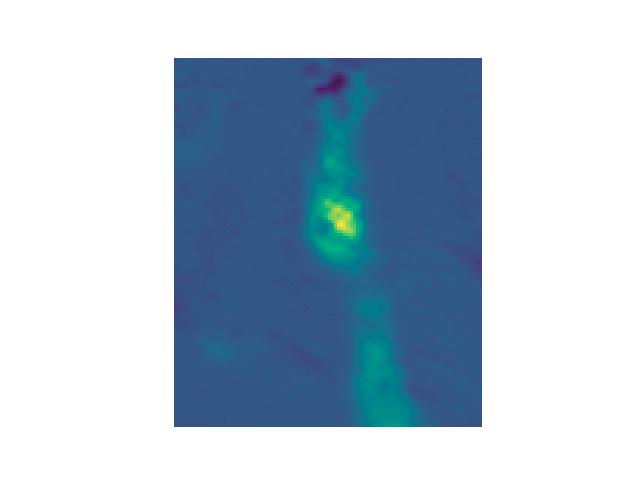}} \\
  
  \subfloat{\includegraphics[height=1.75cm]{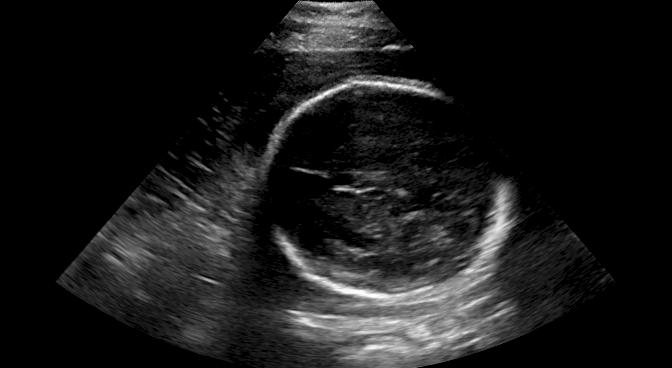}} \hfill
  \subfloat{\includegraphics[height=1.75cm]{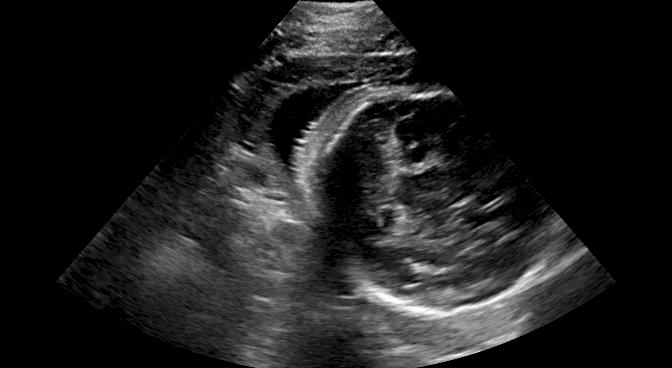}} \hfill
  \subfloat{\includegraphics[height=1.75cm, trim=3cm 3.2cm 3cm 3.2cm, clip]{iFIND00393_Gauss_bbox_text_mse_appendix.jpg}} \hfill
  \subfloat{\includegraphics[height=1.75cm, trim=3cm 3.2cm 3cm 3.2cm, clip]{iFIND00393_Shadow+Gauss_bbox_text_mse_appendix.jpg}} \hfill
  \subfloat{\includegraphics[height=1.75cm]{iFIND00393_Gauss_enlarge_mse_appendix.jpg}} \hfill
  \subfloat{\includegraphics[height=1.75cm]{iFIND00393_Shadow+Gauss_enlarge_mse_appendix.jpg}} \hfill
  \subfloat{\includegraphics[height=1.75cm, trim=4.5cm 1.5cm 4.5cm 1.5cm, clip]{iFIND00393_Gaussian_diff_mse.jpg}} \\
  
  \subfloat{\includegraphics[height=1.75cm]{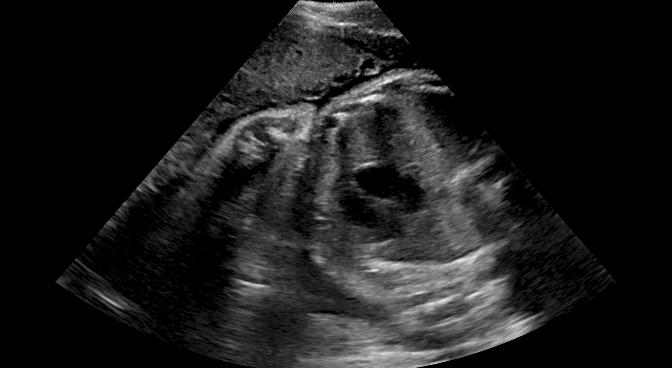}} \hfill
  \subfloat{\includegraphics[height=1.75cm]{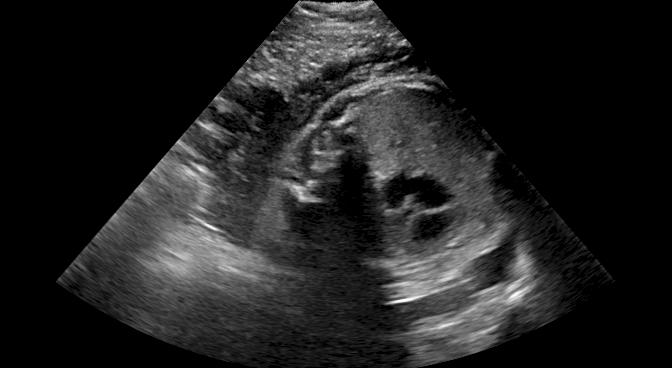}} \hfill
  \subfloat{\includegraphics[height=1.75cm, trim=3cm 3.2cm 3cm 3.2cm, clip]{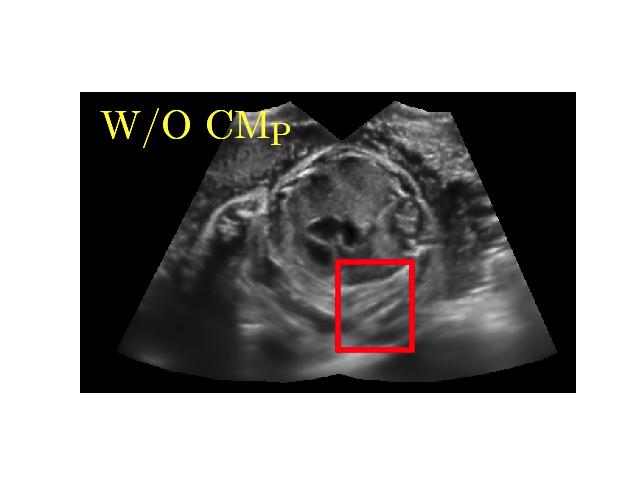}} \hfill
  \subfloat{\includegraphics[height=1.75cm, trim=3cm 3.2cm 3cm 3.2cm, clip]{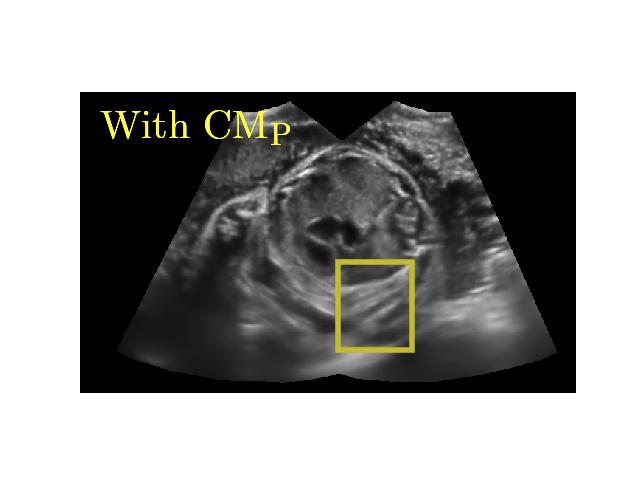}} \hfill
  \subfloat{\includegraphics[height=1.75cm]{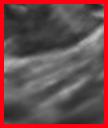}} \hfill
  \subfloat{\includegraphics[height=1.75cm]{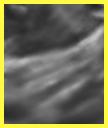}} \hfill
  \subfloat{\includegraphics[height=1.75cm, trim=4.5cm 1.5cm 4.5cm 1.5cm, clip]{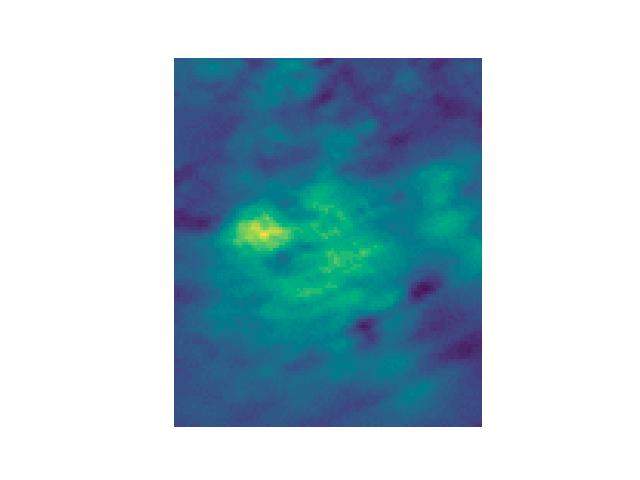}} \\

  \setcounter{subfigure}{0}
  \subfloat[Image Component 1]{\includegraphics[height=1.75cm]{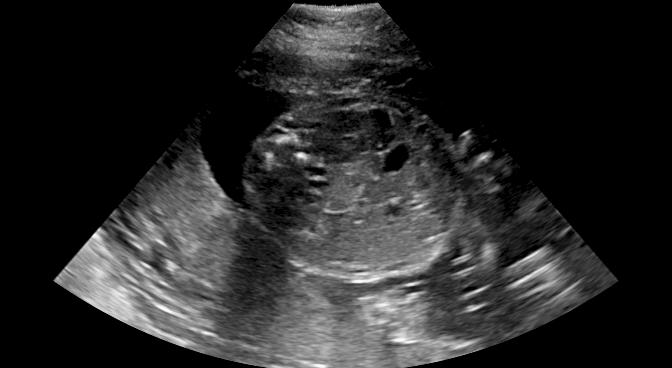}} \hfill
  \subfloat[Image Component 2]{\includegraphics[height=1.75cm]{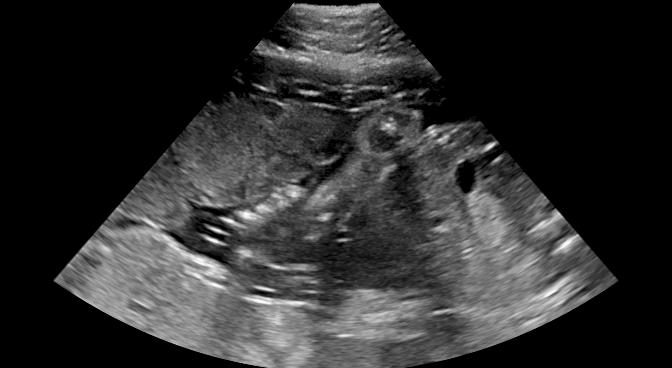}} \hfill
  \subfloat[Without ${CM}_P$]{\includegraphics[height=1.75cm, trim=3cm 3.2cm 3cm 3.2cm, clip]{iFIND00419_Signal+Gauss_bbox_text_sigmoid.jpg}} \hfill 
  \subfloat[With ${CM}_P$]{\includegraphics[height=1.75cm, trim=3cm 3.2cm 3cm 3.2cm, clip]{iFIND00419_Signal+Shadow+Gauss_bbox_text_sigmoid.jpg}} \hfill
  \subfloat[Enlarged]{\includegraphics[height=1.75cm]{iFIND00419_Signal+Gauss_enlarge_sigmoid.jpg}} \hfill
  \subfloat[Enlarged]{\includegraphics[height=1.75cm]{iFIND00419_Signal+Shadow+Gauss_enlarge_sigmoid.jpg}} \hfill
  \subfloat[D$\_$map]{\includegraphics[height=1.75cm, trim=4.5cm 1.5cm 4.5cm 1.5cm, clip]{iFIND00419_Signal+Gaussian_diff_sigmoid.jpg}}
  
  \caption{ The results of the multi-view image fusion. (a-b) The multi-view images, (c) Image fusion without shadow confidence maps (${CM}_P$), (d) Image fusion with shadow confidence maps (${CM}_P$), (e-f) Enlarged areas of (c-d) respectively, and (g) Difference maps of (e) and (f). Rows (1-2) use MSE loss to train networks for generating shadow confidence maps while Row (3-4) use Sigmoid loss. Row 2 uses the Gaussian weighting for image fusion while Rows (1, 3, 4) use the Intensity and Gaussian weighting. The color bar on the top shows that the more yellow/brighter, the higher the difference between the two framed areas.}
  \label{DetimgF}
\end{figure*}

Fig.\ref{DetimgF} shows more examples of the multi-view image fusion task which include the original multi-view images. From the column (a-b) of Fig.\ref{DetimgF}, we can see that the original images contain strong shadow artifacts that can affect the anatomical analysis. The image fusion task aims to use complementary information from images with different views for reducing artifacts and increasing anatomical information. Column (e-f) enlarge the areas within the bounding boxes in column (c-d). Column (g) shows the difference masks between column (e)and (f). The difference masks clearly indicates the improved performance of image fusion after adding shadow confidence maps for Gaussian weighting strategy as well as Intensity and Gaussian weighting strategy. 
% In contrast, since the intensity prior knowledge is powerful enough for this task to fulfill anatomical information, the performance of image fusion using Intensity and Gaussian weighting strategy has weak improvement with additional shadow confidence maps (shown in lower row of column (c) and (d)).      

\subsection{Examples for Biometric Measurement}
\label{appdBioM}

\begin{figure*}[htb]
  \centering
  \subfloat{\includegraphics[height=3.48cm,  trim=2.3cm 1.25cm 1.9cm 1.5cm, clip]{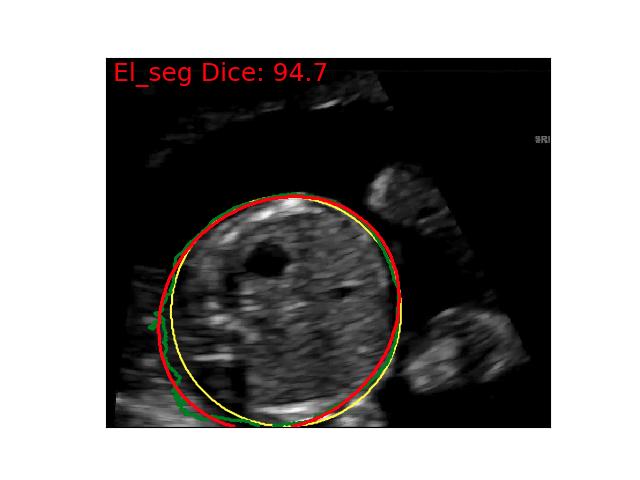}} \hfill
  \subfloat{\includegraphics[height=3.48cm,  trim=2.3cm 1.25cm 1.9cm 1.5cm, clip]{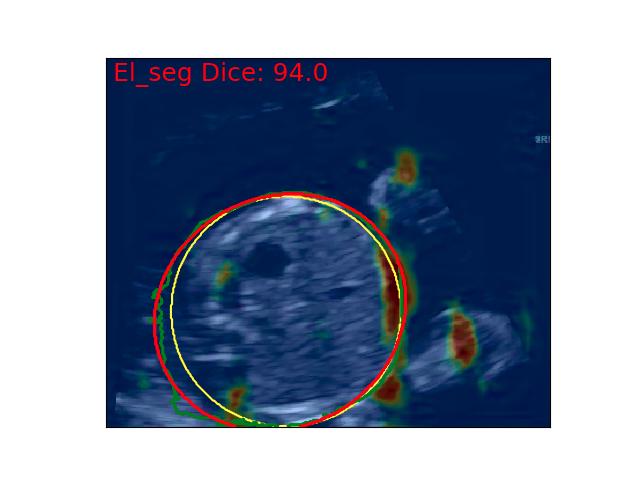}} \hfill
  \subfloat{\includegraphics[height=3.48cm,  trim=2.3cm 1.25cm 1.9cm 1.5cm, clip]{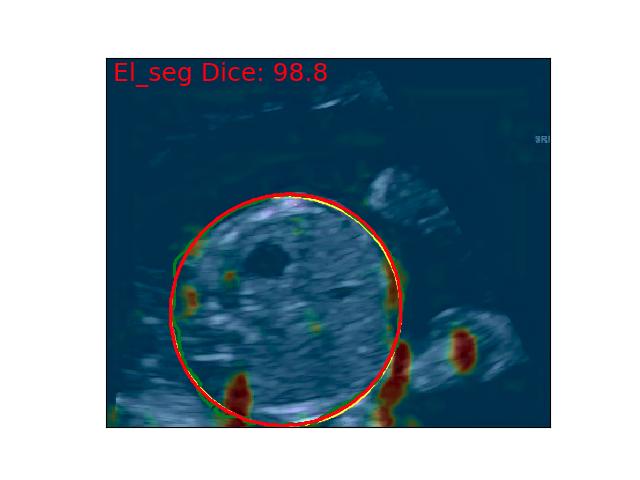}} \hfill
  \subfloat{\includegraphics[height=3.48cm,  trim=2.3cm 1.25cm 1.9cm 1.5cm, clip]{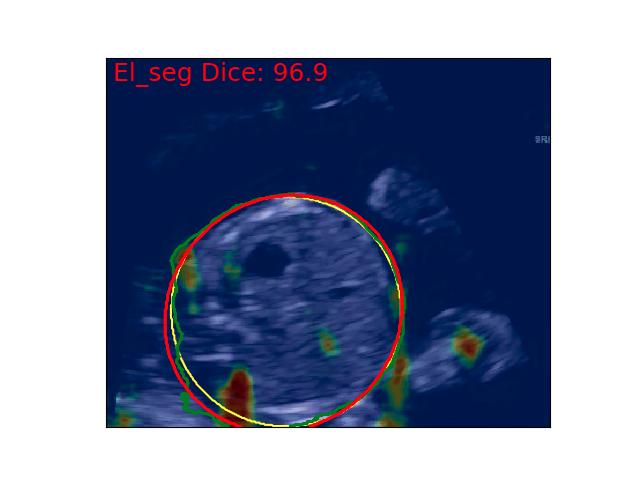}} \hfill
  \\

  \subfloat{\includegraphics[height=3.48cm,  trim=2.29cm 1.29cm 1.88cm 1.48cm, clip]{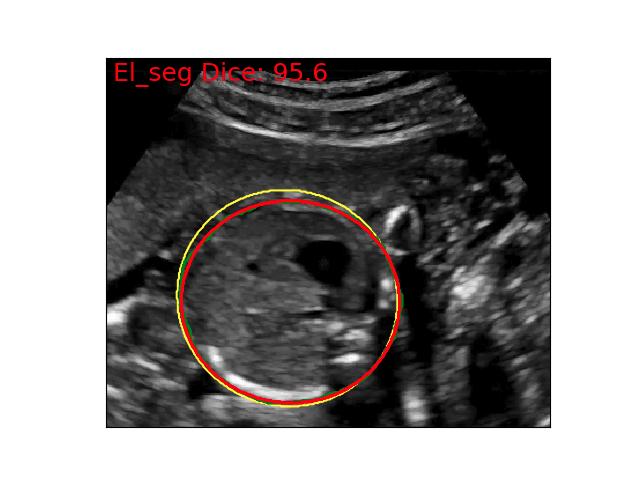}} \hfill
  \subfloat{\includegraphics[height=3.48cm,  trim=2.29cm 1.29cm 1.88cm 1.48cm, clip]{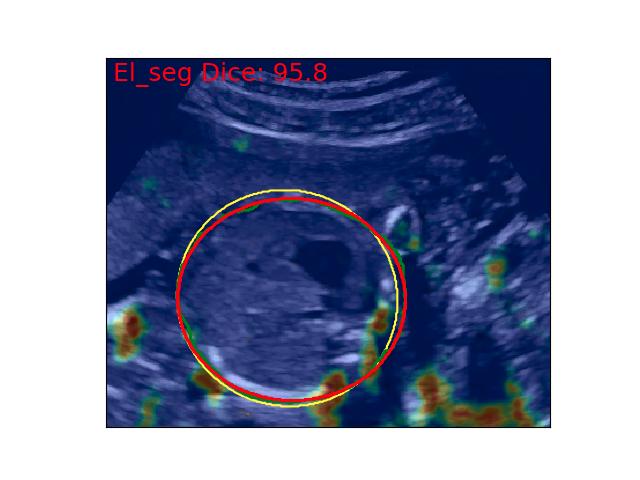}} \hfill
  \subfloat{\includegraphics[height=3.48cm,  trim=2.29cm 1.29cm 1.88cm 1.48cm, clip]{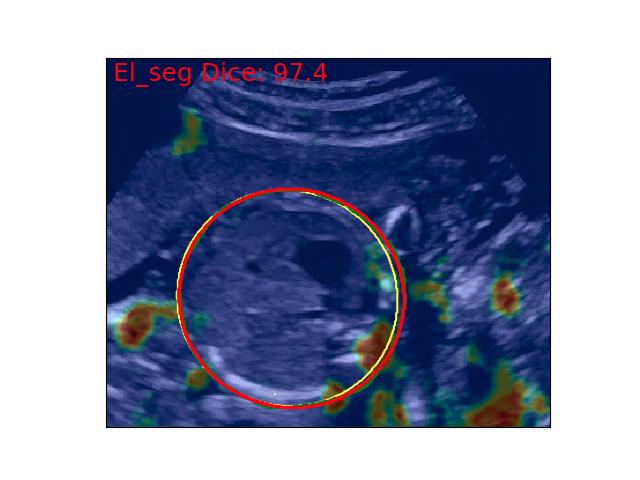}} \hfill
  \subfloat{\includegraphics[height=3.48cm,  trim=2.29cm 1.29cm 1.88cm 1.48cm, clip]{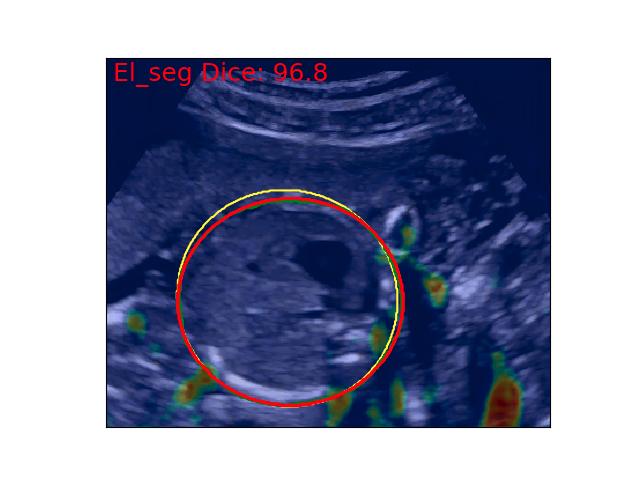}} \hfill
  \\
  
  \setcounter{subfigure}{0}
  \subfloat[w/o $CM$]{\includegraphics[height=3.48cm,  trim=2.29cm 1.29cm 1.88cm 1.48cm, clip]{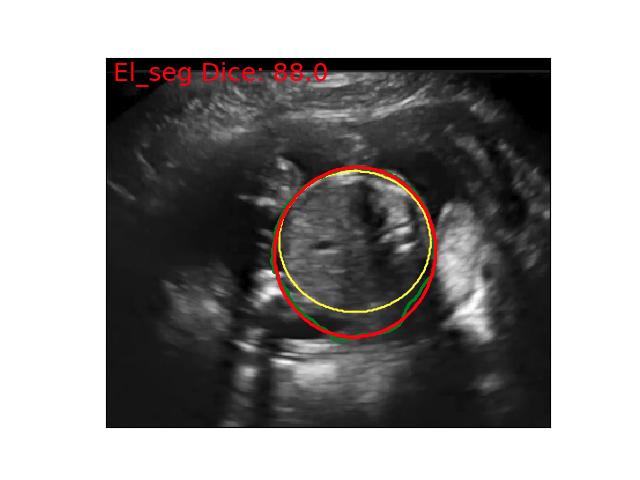}} \hfill
  \subfloat[${CM}_B$]{\includegraphics[height=3.48cm,  trim=2.29cm 1.29cm 1.88cm 1.48cm, clip]{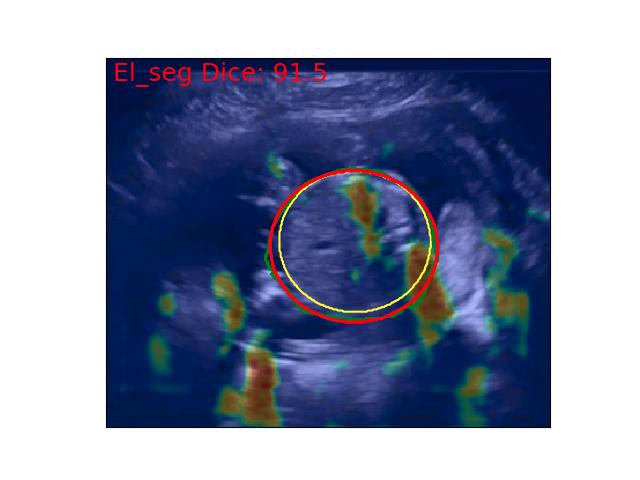}} \hfill
  \subfloat[${CM}_P$]{\includegraphics[height=3.48cm,  trim=2.29cm 1.29cm 1.88cm 1.48cm, clip]{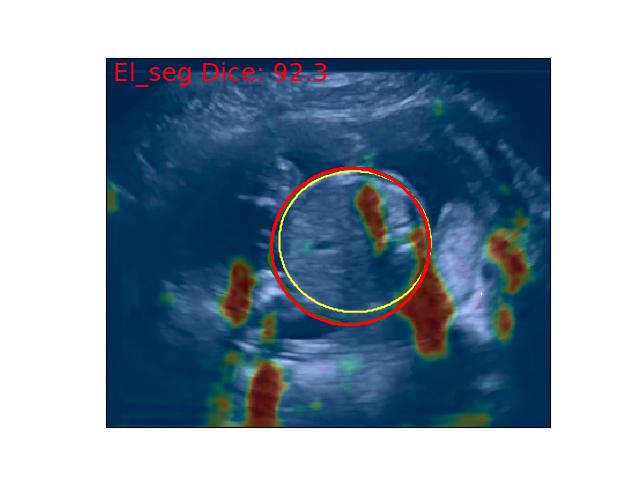}} \hfill
  \subfloat[${CM}_{PAG}$]{\includegraphics[height=3.48cm,  trim=2.29cm 1.29cm 1.88cm 1.48cm, clip]{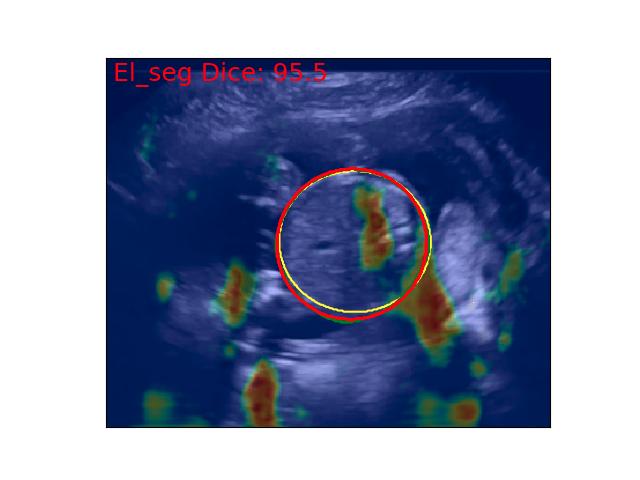}} \hfill
  \\
  
  \caption{Biometric measurement with VS. without shadow confidence maps. The yellow circles refer to the ground truth, the green curves are segmentation predictions, and the red circles are the ellipses of the segmentation prediction.}
  \label{bioM}
\end{figure*}

\begin{figure*}[htb]
 \centering
 \includegraphics[width=1.0\textwidth, trim=1.8cm 0.5cm 2.5cm 0.5cm, clip]{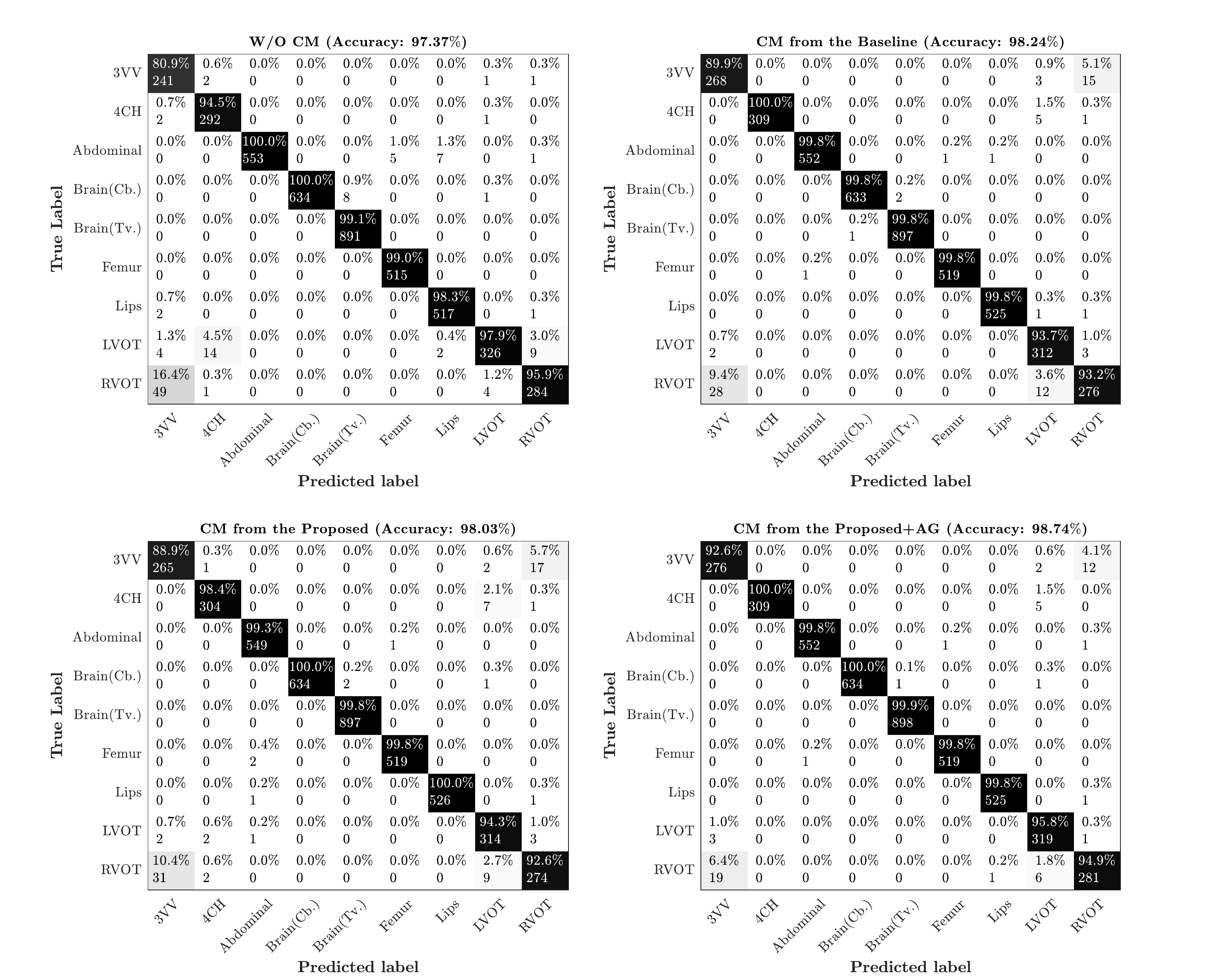}
 \caption{Class confusion metrics for 2D ultrasound standard plane classification. Upper row left: the class confusion matrix without shadow confidence maps. Upper row right: the class confusion matrix with the shadow confidence maps generated by the baseline. Lower row left: the class confusion matrix with shadow confidence maps obtained by the proposed method. Lower row right: the class confusion matrix with the shadow confidence maps produced by the proposed$+$AG method.}
 \label{cm}
\end{figure*}

We visualize the biometric measurement of the three examples shown in Table~\ref{bioMet}. Fig.~\ref{bioM} demonstrates that, for the cases affected by shadow artifacts, the segmentation performance (``EI\_seg DICE") is improved after adding shadow confidence maps as an extra channel. From the first row to the third row in Fig.~\ref{bioM}, these three samples are respectively \#1, \#2 and \#3 samples in Table~\ref{bioMet}.

\subsection{Equations for estimating floating point operations (Flops) for convolutional layers}
\label{flopsEq}

We use Eq.~\ref{flops_compute} to estimate the required Flops for convolution layers including ReLU activation. Here, $W$ and $H$ are the width and height of the input image respectively. $K$ is kernel size, $P$ is the padding, $S$ is the stride and $F$ is the number of filters. $n$ is the size of the convolution layer. ($\text{channels}*K*K$).
\begin{equation}\label{flops_compute}
%\resizebox{.9\hsize}{!}{
\begin{split}
Flops  &\approx \\ 
& \left(\frac{W-K+2*P}{S}+1\right) * \\ & \left(\frac{H-K+2*P}{S}+1\right)  *n*(n-1)*F + F*W*H.
\end{split}
%}
\end{equation}
Eq.~\ref{flops_compute} evaluates the number of Flops ($n$: multiplications and $n-1$: additions) for $W \times H$ filter convolutions adjusted for padding $P$ and stride $S$. ReLU activation is assumed to be $F*W*H$ Flops (one comparison and one multiplication).

\end{document}